%% file: 00_main.tex
\pdfoutput=1
\documentclass{article}
\usepackage{frExamplee}
\usepackage{graphicx}
\usepackage{apalike}
\usepackage{setspace}

\usepackage[binary-units]{siunitx}
\usepackage{subcaption}
\usepackage{hyperref}
\usepackage{booktabs}
\usepackage{multirow}
\usepackage{comment}
\usepackage{graphics}
\usepackage{epsfig}
\usepackage{amsmath}
\usepackage{amssymb}
\usepackage{algorithm}
\usepackage{algpseudocode}
\algnewcommand\AAND{\textbf{ and }}
\algnewcommand\Or{\textbf{ or }}
\usepackage{color}
\usepackage[table,xcdraw]{xcolor}
\usepackage{rotating}
\usepackage{cite}
\usepackage[printonlyused]{acronym}

\DeclareSIUnit[per-mode=symbol,per-symbol=p]{\MBps}{\mega\byte\per\second}
\DeclareSIUnit[per-mode=symbol,per-symbol=p]{\MB}{\mega\byte}
\DeclareSIUnit\px{P}

\newcommand*\degree{^\circ}
\newcommand*\minutes{\,\si{\minute}}

\input{acronyms}

\title{Team CERBERUS Wins the DARPA Subterranean Challenge: Technical Overview and Lessons Learned
}

\author{
Marco Tranzatto$^{1}$
\thanks{$^{1}$ Robotic Systems Lab, ETH Zurich, $^{2}$ Autonomous Robots Lab, Norwegian University of Science and Technology | University of Nevada, Reno, $^{3}$ Autonomous Systems Lab, ETH Zurich, ${4}$ Oxford Robotics Institute, University of Oxford, $^{5}$ Flyability, $^{6}$ University of California, Berkeley. Direct correspondence to Marco Tranzatto \texttt{marcot@ethz.ch} }
\And
Mihir Dharmadhikari$^{2}$\And
Lukas Bernreiter$^3$\And
Marco Camurri$^4$\And
Shehryar Khattak$^1$\And
Frank Mascarich$^2$\And
Patrick Pfreundschuh$^3$\And
David Wisth$^4$\And
Samuel Zimmermann$^1$\And
Mihir Kulkarni$^2$\And
Victor Reijgwart$^3$\And
Benoit Casseau$^4$\And
Timon Homberger$^1$\And
Paolo De Petris$^2$\And
Lionel Ott$^3$\And
Wayne Tubby$^4$\And
Gabriel Waibel$^1$\And
Huan Nguyen$^2$\And
Cesar Cadena$^3$\And
Russell Buchanan$^4$\And
Lorenz Wellhausen$^1$\And
Nikhil Khedekar$^2$\And
Olov Andersson$^3$\And
Lintong Zhang$^4$\And
Takahiro Miki$^1$\And
Tung Dang$^2$\And
Matias Mattamala$^4$\And
Markus Montenegro$^1$\And
Konrad Meyer$^1$\And
Xiangyu Wu$^5$\And
Adrien Briod$^6$\And
Mark Mueller$^5$\And
Maurice Fallon$^4$\And
Roland Siegwart$^3$\And
Marco Hutter$^1$\And
Kostas Alexis$^2$\\
}

\begin{document}

\maketitle

\begin{abstract}
This article presents the CERBERUS robotic system-of-systems, which won the DARPA Subterranean Challenge Final Event in 2021.
The Subterranean Challenge was organized by DARPA with the vision to facilitate the novel technologies necessary to reliably explore diverse underground environments despite the grueling challenges they present for robotic autonomy.
Due to their geometric complexity, degraded perceptual conditions combined with lack of GPS support, austere navigation conditions, and denied communications, subterranean settings render autonomous operations particularly demanding.
In response to this challenge, we developed the CERBERUS system which exploits the synergy of legged and flying robots, coupled with robust control especially for overcoming perilous terrain, multi-modal and multi-robot perception for localization and mapping in conditions of sensor degradation, and resilient autonomy through unified exploration path planning and local motion planning that reflects robot-specific limitations.
Based on its ability to explore diverse underground environments and its high-level command and control by a single human supervisor, CERBERUS demonstrated efficient exploration, reliable detection of objects of interest, and accurate mapping.
In this article, we report results from both the preliminary runs and the final Prize Round of the DARPA Subterranean Challenge, and discuss highlights and challenges faced, alongside lessons learned for the benefit of the community.

\end{abstract}

\section{Introduction}
\input{01_introduction}

\section{Related work}\label{sec:relatedwork}
\input{02_related_work}

\section{CERBERUS system-of-systems}\label{sec:cerberustech}
\input{03_00_cerberus_systems_of_systems}

\section{Results}\label{sec:results}
\input{04_00_results}

\section{Lessons learned}\label{sec:lessons}
\input{05_lessons_learnt}

\section{Conclusions}\label{sec:concl}
\input{06_conclusions}

\subsubsection*{Acknowledgments}
This material is based upon work supported by the Defense Advanced Research Projects Agency (DARPA) under Agreement No. HR00111820045. The presented content and ideas are solely those of the authors.

The authors would also like to thank ANYmal Bear, ANYmal Badger, ANYmal Cerberus, ANYmal Chimera, ANYmal Camel, ANYmal Caiman, ANYmal Coyote, SMB Armadillo, Aerial Scouts Alpha, Bravo, Charlie, Gagarins, RMF-Owl, Kolibri, and Kraken that were not harmed during the DARPA SubT Challenge events, even though the same might not be said for the prior field deployments. We want to thank all members of Team CERBERUS who contributed to the team’s success. We extend our gratitude to all the SubT Community and the DARPA team for the exciting challenge and collaborative community that was built.

\appendix
\section{List of open source packages and data }\label{sec:appendixA}
\input{07_open_source_list}

\bibliographystyle{apalike}

\end{document}

%% file: acronyms.tex
\acrodef{API}{Application Programming Interface}
\acrodef{BT}{Behavior Tree}
\acrodef{CI}{Continuous Integration}
\acrodef{DTW}{Dynamic Time Warping}
\acrodef{FOV}{Field of View}
\acrodefplural{FOV}{Fields of View}
\acrodef{GPU}{Graphical Process Unit}
\acrodef{GUI}{Graphical User Interface}
\acrodef{ICP}{Iterative Closest Point}
\acrodef{IP}{Ingress Protection}
\acrodef{MCI}{Mission Control Interface}
\acrodef{MARI}{Monitor and Artifacts Reporting Interface}
\acrodef{MAV}{Micro Aerial Vehicle}
\acrodefplural{MAV}{Micro Aerial Vehicles}
\acrodef{MAVs}{Micro Aerial Vehicles}
\acrodef{NTP}{Network Time Protocol}
\acrodef{PCA}{Principal Component Analysis}
\acrodef{PCI}{Planner Control Interface}
\acrodef{POE}{Power Over Ethernet}
\acrodef{PTP}{Precision Time Protocol}
\acrodef{ROS}{Robot Operating System}
\acrodef{SLAM}{Simultaneous Localization and Mapping}
\acrodef{QoS}{Quality of Service}
\acrodef{CNN}{convolutional neural network}
\acrodef{GMM}{Gaussian Mixture Models}
\acrodef{UAV}{Unmanned Aerial Vehicle}
\acrodefplural{UAV}{Unmanned Aerial Vehicles}
\acrodef{UGV}{Unmanned Ground Vehicle}
\acrodefplural{UGV}{Unmanned Ground Vehicles}

%% file: 01_introduction.tex
This paper reports the research activities and technological progress made by team ``CERBERUS''\footnote{\url{https://www.subt-cerberus.org/}} towards our winning participation in the Final Event of the DARPA Subterranean (SubT) Challenge. 
The SubT Challenge was a three year-long, \$$82$ million \cite{washingtonpost_article} robotics competition organized and coordinated by the US Defense Advanced Research Projects Agency (DARPA). This challenge aimed to accelerate the research and development of novel technologies to support operations in complex and diverse underground settings, presenting significant challenges for military and civilian first responders. Operating in these environments can often be dangerous for humans; therefore, robotic systems are a viable alternative to provide rapid situational awareness to a small team of operators before possible entry into such unknown, high-risk, and dynamic subterranean areas. Possible use-case scenarios for these new technologies include - but are not limited to - search and rescue missions in collapsed mines, post-earthquake urban settings, and cave networks.
In response to these challenges, we created CERBERUS: a system-of-systems involving walking and flying robots equipped with multi-modal perception capabilities, navigation and mapping autonomy, and self-deployed network communication-extender modules that could autonomously explore complex subterranean environments.

The SubT Challenge featured two competition tracks: Systems and Virtual. Systems track teams deployed their hardware and software solutions to compete in physical subterranean courses. In contrast, Virtual track teams developed software-only solutions evaluated in simulation. The underground scenarios for the Systems track included a) human-made tunnel networks extending for several kilometers and presenting multiple branches and vertical openings (``Tunnel Circuit'', August \num{2019}), b) multi-level urban underground structures with complex multi-storey layouts (``Urban Circuit'', February \num{2020}), and c) natural cave environments with complex, diverse morphologies and constrained passages (``Cave Circuit'', August \num{2020}, canceled event). Finally, a comprehensive benchmark mission combined all the challenges of the previous circuits (``Final Event'', September \num{2021}). The SubT Challenge brought together more than \num{300} competitors from \num{20} different teams, composed of individuals, startups, universities, and large companies that spanned \num{11} countries and \num{20} universities.

Alongside the particular operational goal of robotic subterranean deployments, DARPA initiated the SubT Challenge to stimulate development in four critical technical areas: Autonomy, Perception, Networking, and Mobility. Robotic systems were expected to show a high level of resilient autonomy, demonstrating the ability to map, navigate, and search in complex and dynamic environments with little to no human interventions (Autonomy). Additionally, they had to cope with grueling navigation challenges, operating under varying and degraded conditions including dust, fog, mist, water, smoke, darkness, obscured, and/or scattering environments (Perception). Moreover, the robots had to send information to the Human Supervisor while dealing with lack of line-of-sight, the effects of varying geological conditions, and wireless signal propagation challenges in subterranean environments (Networking). Finally, the systems were tasked to navigate mobility-stressing and dynamic terrain, including narrow passages, sharp turns, large drops/climbs, inclines, steps, falling debris, mud, sand, and water (Mobility). Competing teams were tasked with exploring and mapping unknown underground areas, with the additional requirement that only one team member, the Human Supervisor, was allowed to manage and interact with the deployed robots. Teams were scored based on their ability to localize specified objects, called ``artifacts'', representing features or items of interest applicable to subterranean settings. Artifacts for the Prize Round of the Final Event included human (mannequin) survivors, cellphones, backpacks, drills, fire extinguishers, vents, gas-filled rooms, helmets, ropes, and a SubT cube. The artifacts' surveyed positions were expressed with respect to a DARPA-defined coordinate frame. Each team had \SI{30}{\minute} before the run to set up their Base Station (a computer used by the Human Supervisor to control the robotic agents) and for four team personnel (known as Pit Crew members) to set up and prepare the robots before the beginning of the mission. After the setup time, every team had \SI{60}{\minute} to find as many of the \num{40} artifacts distributed along the course as possible, while having a limited number of available report attempts (to discourage false positives). A point was earned for each report that correctly identified an artifact's class and its position within \SI{5}{\meter} of the object's ground-truth location.

\begin{figure}[H]
    \centering
    \includegraphics[width=0.99\textwidth]{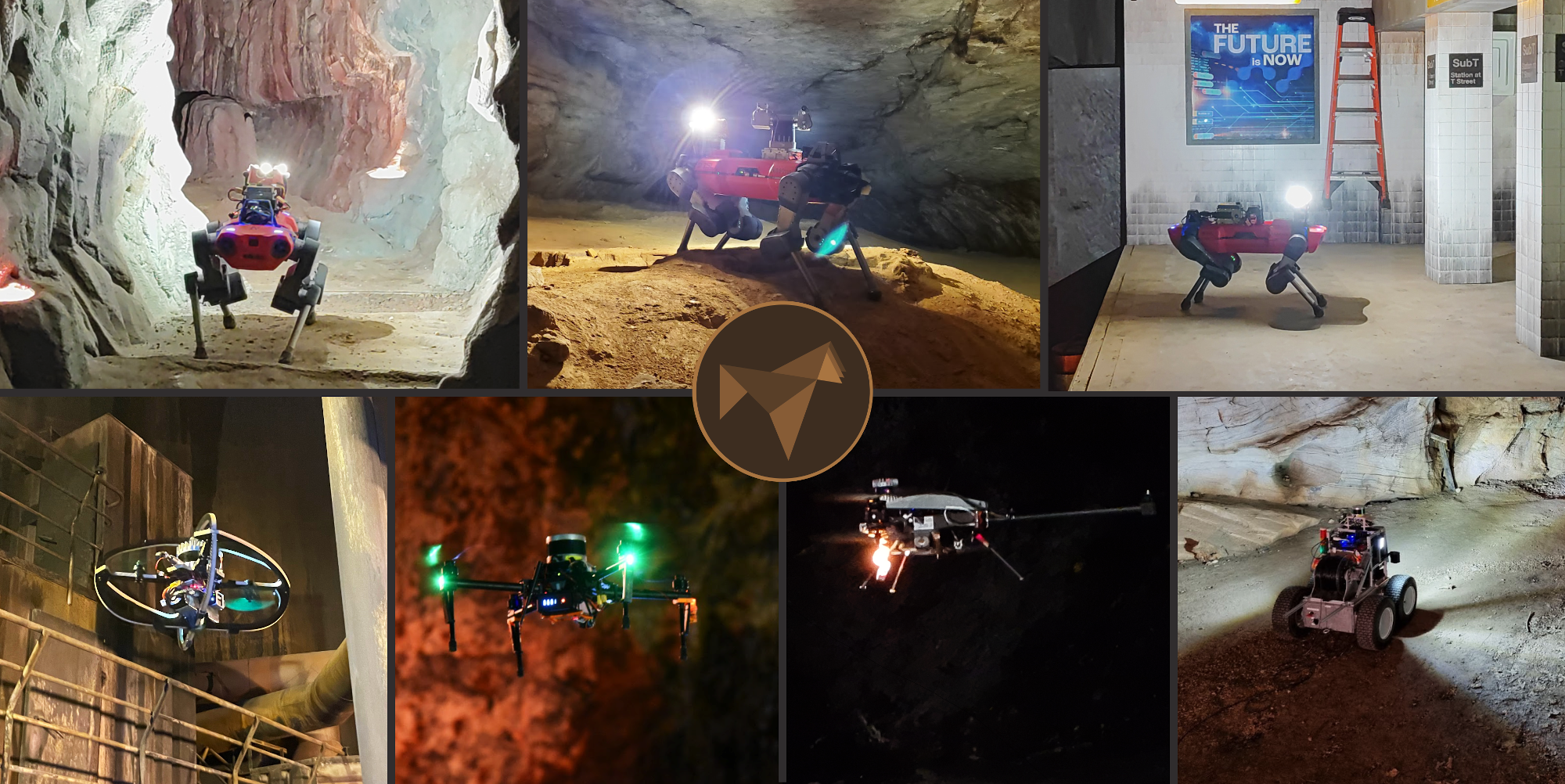}
    \caption{The types of robots developed by Team CERBERUS used for the DARPA SubT Challenge. Upper row: ANYmal C robots. Lower row (left to right): RMF-Owl, Alpha Aerial Scout, and Kolibri flying robots, alongside a tethered roving
platform acting as communication hub for the other robots. All robots are presented in diverse subterranean environments. A video showing the Team's Prize Round is available at~\url{https://youtu.be/QON8IFc8cjE}}
    \label{fig:cerberusrobots}
\end{figure}

Team CERBERUS, which stands for ``CollaborativE walking \& flying RoBots for autonomous ExploRation in Underground Settings,'' draws its name from Greek mythology where Cerberus was the three-headed dog guarding the gates of the underworld. The CERBERUS system-of-systems comprised a composite system of legged and aerial robots. Each agent featured our autonomy and multi-modal perception solutions. The legged agents, in addition, could deploy communication-extender modules to increase the wireless network range. Together, these features enabled resilient navigation, exploration, mapping, and search within complex, large-scale, sensor-degraded, and communications-denied subterranean environments. Moreover, a tethered roving platform assisted the team and acted as a communication hub. Figure~\ref{fig:cerberusrobots} presents most of the types of robots Team CERBERUS developed for the SubT Challenge's Final Event. To gather the necessary expertise, Team CERBERUS involved an international partnership of the University of Nevada Reno, ETH Zurich, Norwegian University of Science and Technology, University of California Berkeley, University of Oxford, Flyability and Sierra Nevada Corporation. Team CERBERUS competed in the System Track of the SubT Challenge and was one of the DARPA-funded teams since the beginning of the competition. On September 23 2021, Team CERBERUS won the Prize Round of the Final Event of the DARPA Subterranean Challenge after three years of intensive developments and an exciting competition among top performers. 

The remainder of this paper is structured as follows: related work is outlined in section~\ref{sec:relatedwork}; a detailed description of the CERBERUS robot and autonomy technologies is provided in section~\ref{sec:cerberustech} followed by a comprehensive overview of the results during the DARPA Subterranean Challenge Final Event in section~\ref{sec:results}; lessons learned are outlined in section~\ref{sec:lessons}, while conclusions are drawn in section~\ref{sec:concl}. 

%% file: 02_related_work.tex
In order for our work to be put into perspective, in this section we attempt to summarize notable contributions in subterranean robotics and autonomy developed both before and outside of the DARPA SubT Challenge, as well as some of work of other competing teams based on the currently available literature. 

\subsection{Autonomous systems in subterranean environments and related settings}

Subterranean settings have long been enticing to the robotics community due to the uniqueness of the environments and the operating conditions~\cite{murphy2009mobile,Tardioli2019GroundRI}. One of the main conclusions from the survey in~\cite{murphy2009mobile} was that robots with navigation, localization, and mapping capabilities developed specifically for the underground scenarios were required. The work in~\cite{Novk2017TelerescuerR} described the concept of a reconnaissance mobile robot for inspecting underground coal mines including the motion, sensing, mapping, and communication modules.~\cite{Loupos2018Tunnel} presented a robotic platform consisting of a robotized boom lift, a high precision robotic arm, advanced computer vision systems, a 3D laser scanner, and an ultrasonic sensor for autonomous tunnel inspection. Additionally, the authors in~\cite{Colas2013pathplanning} proposed a path planning method to allow ground robots to navigate in 3D environments and demonstrated stair climbing. Departing from the need to fully reconstruct 3D maps of the environments, several works explored other directions to allow the robots to autonomously explore underground settings. The authors in~\cite{silver2006topological} described a system based on the Groundhog robot platform which was capable of autonomous topological exploration of mine environments. On the other hand, the contribution in~\cite{nikolakopoulos2019autonomous} presented a \ac{MAV} system that used darkness contour detection to command its heading angle based on the direction of the underground tunnels. Another contribution presented in~\cite{Tolga2016penstocks,Tolga2017penstocks} described an \ac{MAV} system that could fly along the center-line of tunnel-like dam penstocks using range and vision sensors. To mitigate the high bandwidth requirement of transmitting a high-resolution map from the robots to the operators,~\cite{Tabib2021cave} proposed a method for real-time occupancy reconstruction from \ac{GMM} with depth sensor observations. Furthermore, the work in~\cite{qin2019heterogeneous} utilized a collaborative team of an \acp{UAV} and an \acp{UGV} where the \ac{UGV} performed fast autonomous exploration to construct 2.5D coarse maps and the UAV conducted 3D fine mapping of the environment. While most previous work focused on solving one or several challenges in the underground environments or related settings, the SubT Challenge pushed the robotics community to develop novel technologies in all four technical areas: Autonomy, Perception, Networking, and Mobility. For the SubT Challenge, fragile autonomy, frequent and demanding supervision or teleoperation, over-specialization to one or the other type of underground environments, sensitivity against sensor degradation, or dependence on always-available communications were not options. The teams were tasked to present a solution that could demonstrate efficient subterranean exploration and mapping with the highest degree of resilient autonomy across environment configurations or operational conditions. 

\subsection{Other teams in the SubT Challenge}
\label{sec:related_work_other_subt_teams}
In response to the SubT Challenge, several international teams were assembled, each developing highly advanced and complex solutions. Below we outline key elements of the solutions of the other teams competing in the challenge finals. As at the time of writing not all technical overviews are available, we provide a summary based on already published work of the teams and information provided during the Summit that took place after the Final Event. 

Team Coordinated Robotics used custom wheeled, tracked and flying robots, leveraging several open-source mapping frameworks for the ground vehicles~\cite{shan2020lio_sam,shan2018lego_loam} and the aerial robots~\cite{labbe2019rtab} depending on the specific sensor setup. To traverse through narrow areas, the robots employed a wall following mode or were driven manually by the Human Supervisor while frontier-based exploration mode was deployed to explore open areas. Furthermore, a mix of WiFi and Gigabit fiber ethernet devices running the advanced B.A.T.M.A.N routing protocol~\cite{Johnson2008batman}, in addition to the droppable wireless nodes, created the communication system for the robotic team. For the robots to detect artifacts, YOLOv4~\cite{yolo_v4_alexey} network was run with onboard camera images while the gas sensor and bluetooth detection were also utilized to detect artifacts. Team Coordinated Robotics was a self-funded team. 

Team CoSTAR developed a fleet of Spot legged robots, Husky wheeled robots, Balto RC cars, Telemax hybrid wheeled/tracked vehicles and custom flying as well as hybrid ground/aerial robots for the Final Event. Their approach was based on an uncertainty-aware autonomy framework, dubbed NEBULA~\cite{nebula2021}, which constructed the belief space over the robot and world states and performed reasoning and planning in that belief space. NEBULA was an extensive architecture including several modules: (i) robust, resilient \ac{SLAM} front-end~\cite{palieri2020locus} and back-end~\cite{ebadi2020lamp} for large-scale environments, (ii) semantic understanding and artifact detection modules~\cite{Terry2020ObjectAG}, (iii) uncertainty-aware traversability analysis as well as risk- and constraint-aware local planning~\cite{fan2021step,thakker2021offroad} modules, (iv) efficient global exploration and coverage motion planning in belief space using a hierarchical planning approach~\cite{Kim2021PLGRIMHV}, (v) high-bandwidth and resilient network solution for a multi-robot system~\cite{Ginting2021chord}, (vi) learning-enabled adaptation module~\cite{fan2020tubempc,fan2020bayesian}. The autonomy stack was implemented for several robot types including wheeled, legged, flying robots, and field-tested in various environments. Team CoSTAR was a DARPA-funded team from the beginning of the challenge. 

Team CSIRO Data61's approach was presented in~\cite{hudson2021csiro} in which a team of BIA5 OzBot tracked robots, Spot legged robots, and Emesent Hovermap drones utilized the Wildcat SLAM system~\cite{milad2022wildcat}. The robots explored the environments by choosing the most informative 3D frontier in an efficient manner~\cite{williams2020frontier}. Moreover, for the UGVs to navigate through challenging unstructured terrain, a common local navigation module optimized for negative obstacles was developed in~\cite{hines2020virtual} as well as a deep reinforcement learning approach to allow the tracked robots to traverse through narrow gaps in the environments~\cite{tidd2021narrow_gap}. Additionally, the work in~\cite{Tychsen2018fitness} described a novel method used to obtain highly competitive localization accuracy with reasonable inference time for artifact detection and localization. A mesh network of radios formed by the modules on the ground robots, \acp{UAV}, and droppable nodes was utilized to share data between nearby robots and the Base Station when connected, as described in~\cite{hudson2021csiro}. Additionally, the artifacts were detected using onboard RGB cameras, and by measuring Bluetooth/WiFi signal strengths and gas concentration. The detection was then relayed back to the Human Supervisor for reviewing before submitting the artifact report to the scoring server (DARPA Command Post). Team CSIRO Data61 was a DARPA-funded team from the beginning of the challenge. 

Team CTU-CRAS-NORLAB's multi-robot heterogeneous exploration system included Clearpath Husky wheeled robots, Absolem tracked robots, PhantomX Mark II crawling hexapods, and custom flying robots~\cite{roucek2020darpa,rouek2021systemFM} in addition to several Boston Dynamics Spot, HEBI robotics LiLy, and custom SCARAB II legged robots as well as Superdroid Marmotte HD2 tracked robots. Their mapping solution was based on a variant of the \ac{ICP} algorithm as per~\cite{pomerleau2013libpointmatcher,pomerleau2014long}. The frontier-based exploration strategy described in~\cite{bayer2019autonomous} was deployed on the UGVs to explore the environments while the elevation mapping framework for the ground robots was described in~\cite{bayer2019autonomous,bayer2020modelling}. Additionally, the state estimation and control pipelines used for the aerial robots were presented in~\cite{baca2021mrs} and the path planning as well as frontier-based exploration strategy were detailed in~\cite{kratky2021autonomous}. Furthermore, three different communication systems were utilized: short-range link (WiFi), mid-range link (Mobilicom mesh network), and long-range link (droppable ``Mote'' modules) supported communication with different levels of reliability, bandwidth, and usage, as detailed in~\cite{roucek2020darpa}. Moreover, artifact detection was performed onboard every robot using a customized version of YOLOv3~\cite{yolo_v3_redmon}, then a Kalman filtering over multiple detections that was temporally consistent was utilized to estimate the final position of the detected artifact. Team CTU-CRAS-NORLAB was, for the first two competition events of the SubT Challenge, a self-funded team but became a DARPA-funded team due to their performance. 

Team Explorer’s entry to the Final Event of the SubT Challenge included a combination of custom wheeled \acp{UGV}, \acp{MAV}~\cite{Scherer:2021} in addition to a Spot legged robot with custom payload. Their mapping framework, illustrated in~\cite{zhao2021superodom}, employed an IMU-centric approach to combine the advantages of tightly-coupled and loosely-coupled methods. Specifically, in the SubT Challenge, the LiDAR-inertial version of this framework was utilized. The planning module used a hierarchical framework, described in~\cite{cao2021tare}, including a global planner level which maintained data sparsely and computed coarse paths at the global scale, and a local planner which planned kinodynamically feasible paths in the local area for the vehicle. Additionally,~\cite{fan2021far} developed another planning approach that dynamically updated a global visibility graph. This planner was capable of dealing with navigation tasks in both known and unknown environments as well as dynamic obstacles. For the robots to communicate with each other and with the Base Station, droppable radio nodes were deployed by the ground robots according to a cost function using distance, line-of-sight and RSSI conditions, as presented in~\cite{Scherer:2021}, or a predictive mapping approach, as per~\cite{tatum2020thesis}, to maximize communications coverage. Moreover, a multi-modal artifact detection and localization system was developed in~\cite{vasu2019thesis} to help provide situational awareness of the environments. Team Explorer was a DARPA-funded team from the beginning of the challenge. 

Team MARBLE deployed Spot legged robots and Husky wheeled robots in the Final Event with the SLAM pipeline based on the tightly-coupled lidar-inertial odometry framework described in~\cite{shan2020lio_sam}. For volumetric mapping, the team used a modified version of~\cite{hornung13auro} for efficient map sharing as well as incorporating traversability and stairs semantic information. In the Final Event, a graph-based global planner combined with a local reactive controller~\cite{michael2021multi_agent} was deployed while the multi-agent coordination strategy and artifact detection pipeline were illustrated in~\cite{michael2021multi_agent}. Additionally, a custom-developed ROS transport solution based on UDP and a Meshmerize mesh network that prioritized reconnection times were utilized to create effective communication systems in subterranean environments as detailed in~\cite{michael2021multi_agent}. Team MARBLE was a DARPA-funded team from the beginning of the challenge.

Finally, team Robotika used only custom wheeled robots which were equipped with 2D LiDARs and Realsense RGBD cameras for SLAM, traversability estimation, and obstacle avoidance as well as LoRa and WiFi modules for communication purposes. Team Robotika was a self-funded team. Finally, it is worth mentioning that additional teams participated in earlier phases of the SubT Challenge and the above discussion is limited to those competing in the Final Event. 

%% file: 03_00_cerberus_systems_of_systems.tex
This section outlines the CERBERUS' robots, their system design, and specifics with respect to the key technical areas that characterized the SubT Challenge: Mobility, Perception, Autonomy, and Networking. Additionally, the last sections show how the robots detected the artifacts placed in the environment and how the Human Supervisor interacted with the deployed agents.

\subsection{Mobility}
\input{03_01_mobility}

\subsection{Perception}
\input{03_02_perception}

\subsection{Autonomy}
\input{03_03_autonomy}

\subsection{Networking}
\input{03_04_networking}

\subsection{Artifacts scoring}
\input{03_05_artifacts}

\subsection{Single human supervisor}
\input{03_06_single_human_supervisor}\label{sec:single_human_supervisor}

%% file: 03_01_mobility.tex
In this section, we summarize the main robotics platforms deployed by Team CERBERUS at the Final Event. We introduce the ANYmal C SubT and the subterranean aerial robots in terms of their hardware development and main software components. Finally, we present a roving robot used to support underground operations.

\subsubsection{Subterranean walking robots - ANYmal C SubT}\label{sec:anymal_c_subt}
Team CERBERUS developed and deployed a team of legged robots specialized to operate in underground environments. In the Tunnel and Urban events, the team relied on the ``ANYmal B300'' quadrupeds by ANYbotics \cite{tranzatto2022cerberus}, which were replaced with the ``ANYmal C100'' series quadruped from the same company for the Final Event. The ANYmal C100 series quadrupedal robot is a robust and agile legged system tailored for autonomous, long-endurance tasks in challenging environments. This robot is water-proof and dust-proof (\ac{IP} \num{67}), able to operate in humid and dusty conditions, and designed for robustness for long-term operations. It features dedicated software and hardware interfaces to make it possible to extend it for application-specific purposes. We customized this platform with several hardware and software modifications to create a specialized version: the ANYmal C ``SubT''.

\begin{figure}[h]
    \begin{subfigure}{.48\textwidth}
      \centering
      \includegraphics[keepaspectratio,height=6cm]{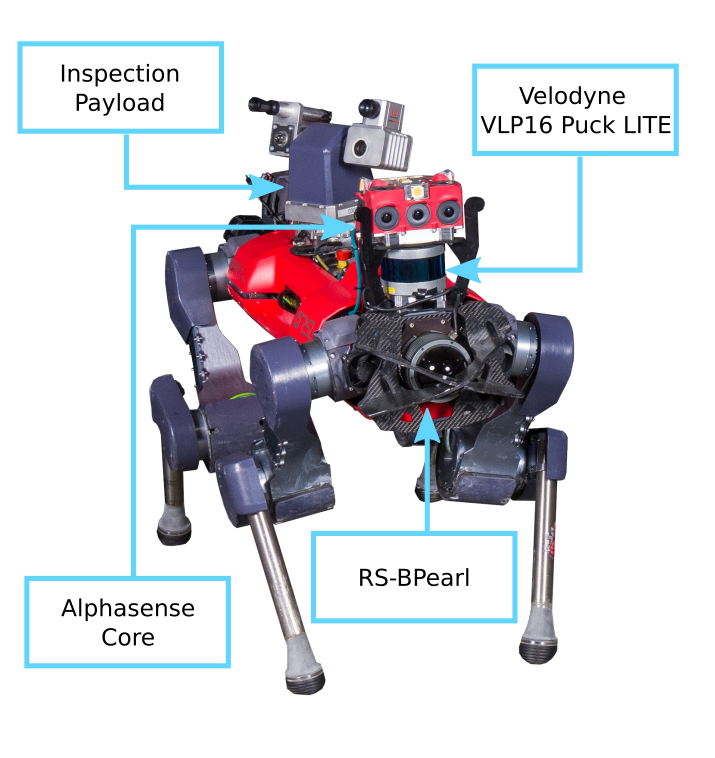}
      \caption{ANYmal C SubT Explorer}
      \label{fig:anymal_c_subt_explorer}
    \end{subfigure}
    \begin{subfigure}{.48\textwidth}
      \centering
      \includegraphics[keepaspectratio,height=6cm]{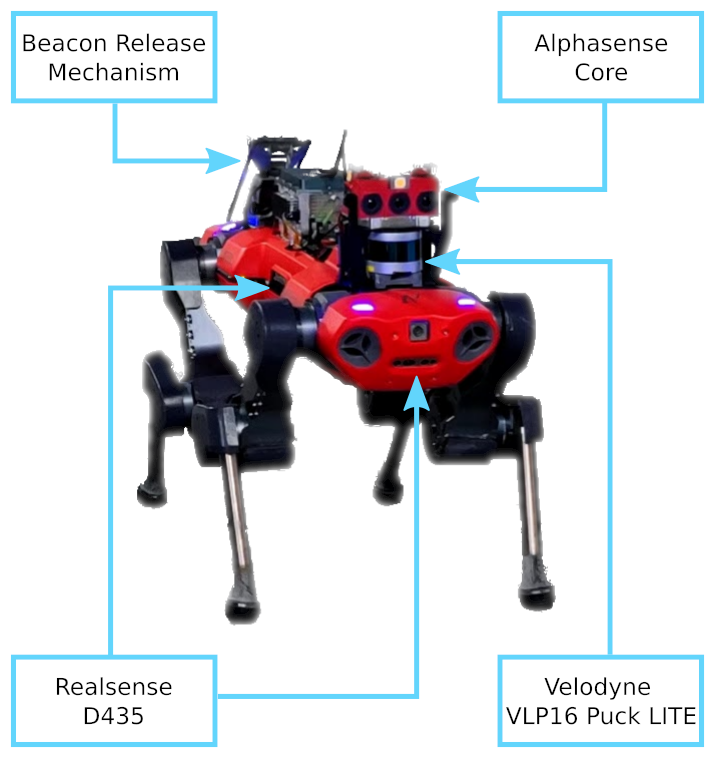}
      \caption{ANYmal C SubT Carrier}
      \label{fig:anymal_c_subt_carrier}
    \end{subfigure}
    \caption{ANYmal C SubT robots used in the Final Event.}
\end{figure}

The ANYmal C SubT platform weighed \SI{55}{\kilo\gram} including payload and could operate for \SI{80}{\minute} while continuously walking. There were two types of ANYmal C SubT robots: the ``Explorer'' (Figure~\ref{fig:anymal_c_subt_explorer}) and the ``Carrier'' (Figure~\ref{fig:anymal_c_subt_carrier}) robots, which differed in their sensor configurations and mission role. The Explorers' goal was to proceed deep in the subterranean environments, while detecting as many artifacts as possible. The Carrier robots had fewer sensors and instead supported underground operations by carrying and deploying the communication-extender modules to increase the range of the wireless network.

\begin{figure}[h]
    \centering
    \begin{subfigure}{.4\textwidth}
      \centering
      \includegraphics[keepaspectratio,height=5cm]{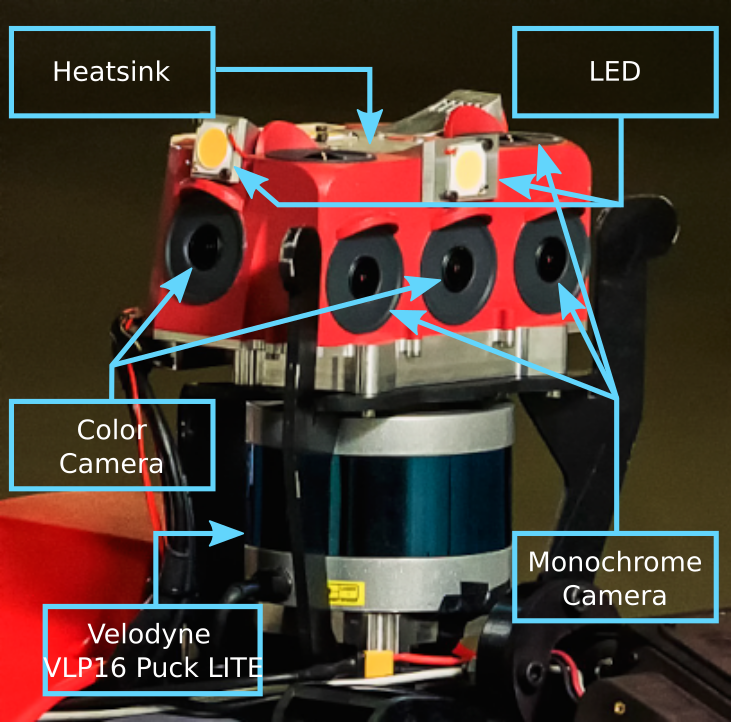}
      \caption{Common sensor suite}
      \label{fig:anymal_c_subt_sensor_head}
    \end{subfigure}
    \begin{subfigure}{.4\textwidth}
      \centering
      \includegraphics[keepaspectratio,height=5cm]{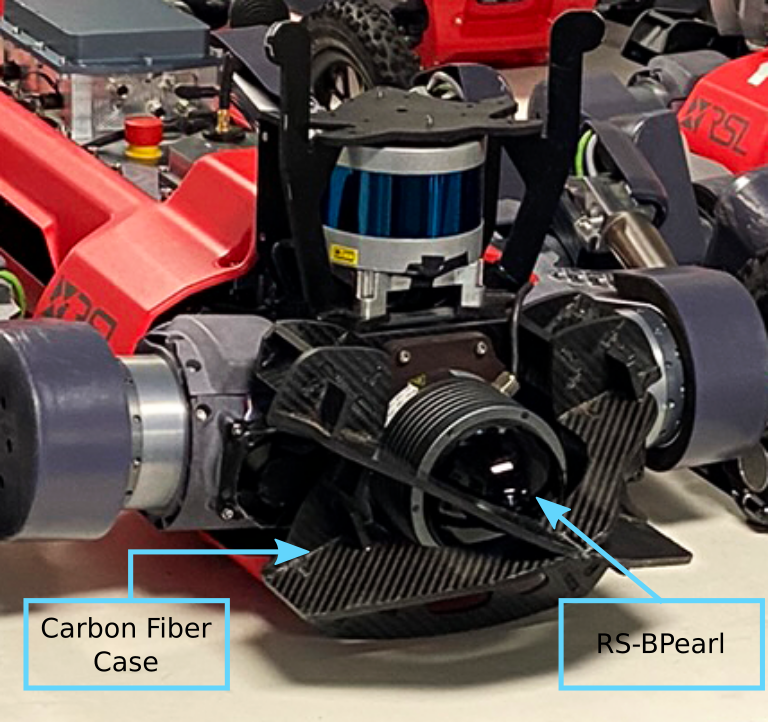}
      \caption{RS-BPearl LiDAR and its protection}
      \label{fig:anymal_c_subt_bpearl_mount}
    \end{subfigure}
    
    \bigskip
    \begin{subfigure}{.4\textwidth}
      \centering
      \includegraphics[keepaspectratio,height=5cm]{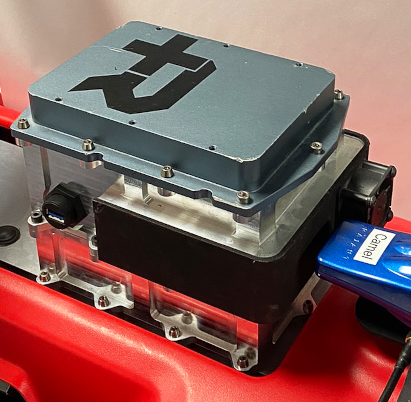}
      \caption{\ac{GPU} Payload}
      \label{fig:anymal_c_subt_jetson_payload}
    \end{subfigure}
    \begin{subfigure}{.4\textwidth}
      \centering
      \includegraphics[keepaspectratio,height=5cm]{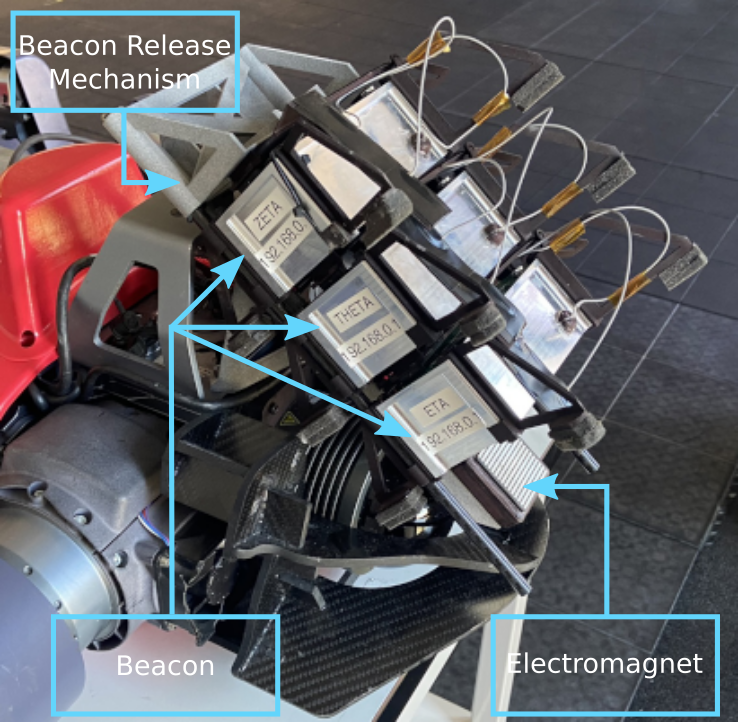}
      \caption{Beacon release mechanism}
      \label{fig:anymal_c_subt_dropping_mechanism}
    \end{subfigure}
    \caption{Hardware customization for the ANYmal C SubT. (\subref{fig:anymal_c_subt_sensor_head}) Common sensor suite included the VLP16 Puck LITE LiDAR sensor and the Alphasense Core unit.
    (\subref{fig:anymal_c_subt_bpearl_mount})RS-BPearl LiDAR for elevation mapping enclosed in its protective and lightweight carbon fiber case. (\subref{fig:anymal_c_subt_jetson_payload}) Jetson Xavier and additional peripherals mounted on top of the Carrier robots, inside an actively cooled case. (\subref{fig:anymal_c_subt_dropping_mechanism}) Beacon release mechanism with three loaded WiFi beacons held by electromagnets.}
\end{figure}

\textbf{Hardware modifications}\label{sec:anymal_modifications}\\
\emph{Sensors} - The ANYmal C SubT robots featured a common sensor suite located on one side of the platform for the Final Event. This suite included a ``VLP16 Puck LITE'' LiDAR by Velodyne and an ``Alphasense Core'' visual-inertial sensor by Sevensense (Figure~\ref{fig:anymal_c_subt_sensor_head}).
The Alphasense Core encapsulated three monochrome and four color cameras, giving each robot a wide field of view (front, left, right, and upwards directions) to increase the chances of detecting artifacts while exploring unknown areas. Each camera has a \SI{0.4}{\mega\px} sensor and mounts a lens with opening angles of \SI{165.4}{\degree} x \SI{126}{\degree} x \SI{92.4}{\degree} (DxHxV).
The Alphasense was fixed on a rigid, custom-made CNC milled aluminium mount to position the cameras precisely and tilt the side cameras to have clear field of view without including any part of the robot. To protect the unit against water and dust the entire camera array was enclosed in a sealed PA12 Multijet Fusion Case with flexible SLA gaskets around the cameras.
Each sensor unit was equipped with four high power, air cooled LEDs. Each of the four LEDs pointed in one of the camera directions (left, right, front and upwards) illuminating the entire field-of-view.
Additionally, each ANYmal featured a ``SDC30'' gas sensor by Sensirion and a Bluetooth sensor.
In addition to the common sensors suite, the Explorers were mounted with a pan-tilt unit head (Inspection Payload) for artifact detection. This head featured a \si{10}{x} optical zoom camera, a thermal camera, a microphone, and a spotlight. Its yaw and pitch angles could be controlled to inspect desired areas around the robot, where the main cameras could not detect potential artifacts due to distance or darkness. Each hardware modification was designed to respect the robot's Ingress Protection Code, thus resulting in the ANYmal SubT version to maintain the \ac{IP} \num{67} rating.

\emph{Terrain mapping sensors} - Reliable terrain perception is critical for safe navigation and locomotion, especially during autonomous missions. Therefore, terrain mapping sensors should be accurate and reliable in challenging environmental conditions and cope with different difficulties, such as darkness or reflective surfaces. The Explorer and Carrier agents presented different sensor configurations reflecting their mission roles.
Carrier robots used four ``RealSense D435'' active stereo sensors by Intel. This is a relatively lightweight and compact sensor (\SI{90}{\milli\meter} x \SI{25}{\milli\meter} x \SI{25}{\milli\meter}, \SI{72}{\gram}), allowing each Carrier to additionally ferry on its back four communication-extender modules. 
On the other hand, the Explorers presented two ``RS-BPearl'' LiDARs by Robosense (\SI{100}{\milli\meter} x \SI{111}{\milli\meter} x \SI{100}{\milli\meter}, \SI{920}{\gram}), each mounted on one end of the robot (Figure~\ref{fig:anymal_c_subt_bpearl_mount}). The RS-Bpearl is a dome-LiDAR that measures the reflection times of emitted laser pulses using a hemispherical scanning pattern. It offers a \SI{90}{\degree} field of view but presents a sparse point-cloud compared to other laser sensors (such as the ``Ouster OS0'' LiDAR). It is robust against sunlight and absorbent material, especially if compared to active stereo sensors. We chose it for the Explorer's configuration because of its sturdy performance in different underground settings, featuring dust, water, and reflective materials, thus making it a good fit for the role of these robots.

\emph{Parallel computing} - The terrain mapping algorithms can be efficiently parallelized using a \ac{GPU}, while visual object-detection also benefits from running neural-networks on a \ac{GPU}. A ``Jetson AGX Xavier'' by Nvidia was used on the legged agents. The Explorer robots featured an on-board \ac{GPU} inside their main body. In contrast, we developed an extra hardware module (Figure~\ref{fig:anymal_c_subt_jetson_payload}) containing a \ac{GPU} and additional electronics for the Carriers to ensure all the ANYmal C SubT robots had the same computational hardware configuration. We developed an air-cooled, water-proof, dust-proof, milled case to house the Jetson module and the additionally needed electronics.

\emph{Beacon release mechanism} - To deploy the communication-extender modules (also referred to as WiFi breadcrumbs or beacons) described in Section ~\ref{sec:deployable_breadcrumb_nodes}, the Carrier ANYmals featured a release mechanism (Figure~\ref{fig:anymal_c_subt_dropping_mechanism}). It could carry four breadcrumbs, each held in place by an electromagnet. To deploy a beacon, the robot tilted its base and deactivated the electromagnet. The beacon slid along the low friction carbon fiber tubes, out of the mechanism, to land in an upright position on the ground.

\textbf{Locomotion controller}\label{sec:anymal_locomotion}\\
To walk stably on challenging terrain and ensure that the ANYmal robots could explore deep in the course, our locomotion controller had to fulfill the following conditions.
First is the ability to walk on a variety of challenging terrains. The course contained tunnel, urban, and cave sections simultaneously, presenting different characteristics. Our controller was able to handle these diverse challenges posed by different settings without further input from the supervisor, i.e., it was not necessary to specify whether the controller had to use a specific gait to climb stairs or to locomote over rough terrain.
Next is robustness. Both sensor malfunctioning and measurement degradation can occur during a mission. On the one hand, sensors themselves could stop working for several reasons, such as connection failure or physical damage, making new measurements unavailable. On the other hand, sensors’ measurements can degrade due to external factors such as smoke, mist, fog, etc. The controller must be robust even under these conditions.
We deployed a learning-based perceptive locomotion controller trained in simulation~\cite{miki2022learning} to satisfy these requirements.
The robot learned to walk over randomly generated terrains within the simulation environment using reinforcement learning. It receives geometric information from an elevation map to walk smoothly over large steps or stairs.
In addition to this, we used domain randomization during learning, such as the mass of the robot, the friction coefficient of the ground, and external forces so that it can respond to various situations that robots face in the real world.
To enable the robot to respond to these cases, we added various distractors during learning to train the robot not to rely on map information when it becomes unreliable or unavailable.
As a result, a single controller was able to handle all course areas during the competition.
We also trained the body tilting action to deploy the wifi beacon on sloped or rough surfaces.

\subsubsection{Subterranean aerial robots}\label{sec:aerial_robots}
\begin{figure}
    \begin{subfigure}[b]{.48\textwidth}
      \centering
      \includegraphics[scale=0.6]{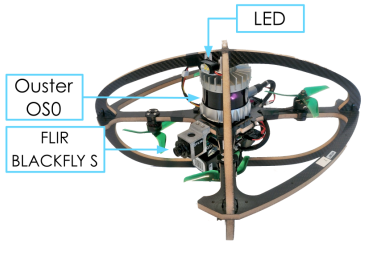}
      \caption{RMF-Owl}
      \label{fig:rmf_owl}
    \end{subfigure}
    \begin{subfigure}[b]{.48\textwidth}
      \centering
      \vfill
      \includegraphics[scale=0.1]{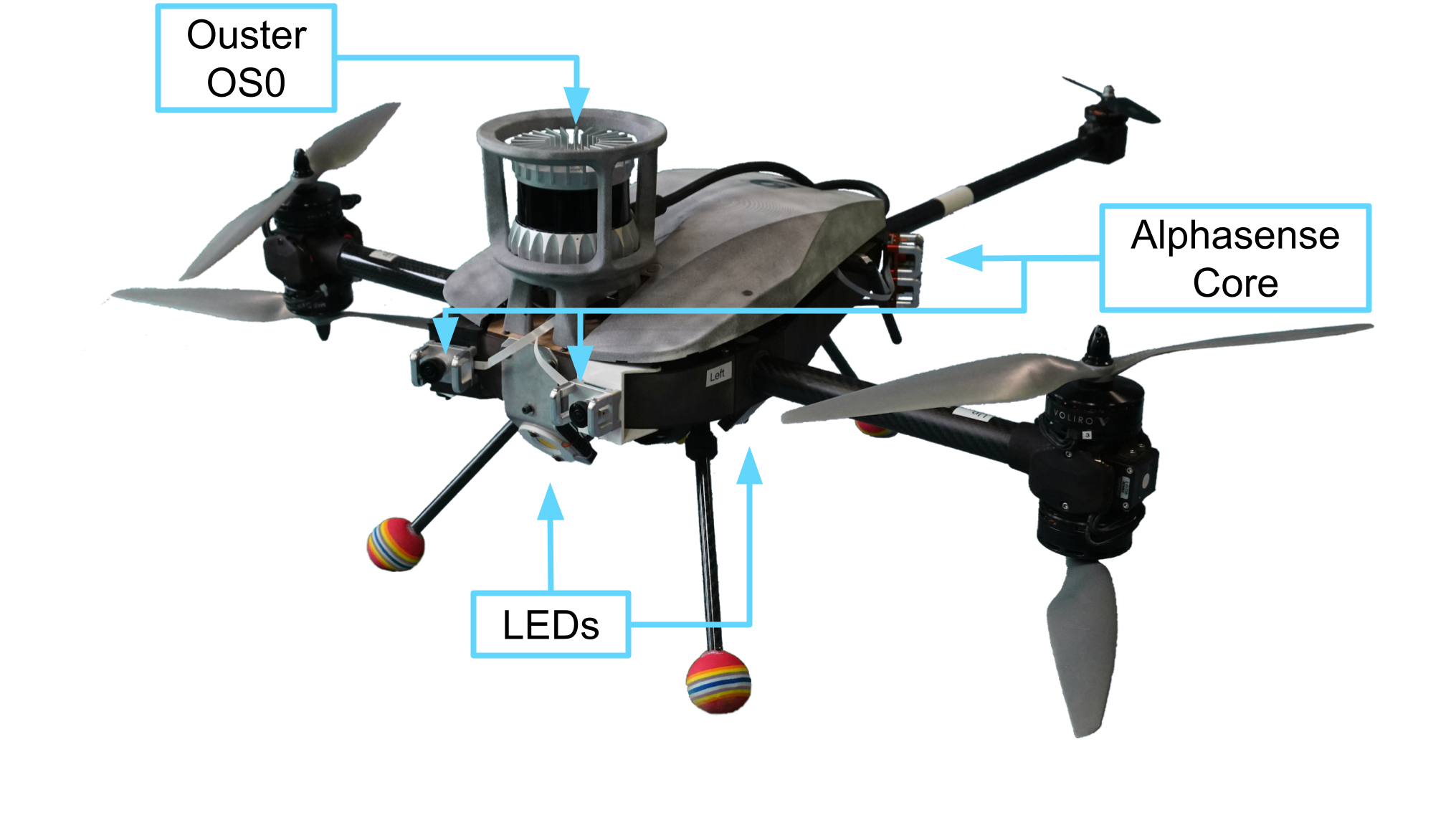}
      \caption{Voliro-T Kolibri}
      \label{fig:kolibri}
    \end{subfigure}
    \caption{The new aerial robots developed for the Final Event.}
    \label{fig:rmf_kolibri}
\end{figure}

Alongside the team of legged robots, Team CERBERUS developed and deployed a team of ``Aerial Scouts'' to provide rapid exploration capabilities, especially in regions of the course which were inaccessible by ground robots. In the Tunnel and Urban events, the team relied on traditional multirotor systems and collision-tolerant designs. First, a set of systems based on the DJI Matrice $100$ platform were developed and modified to host a $3\textrm{D}$ LiDAR sensor alongside color and thermal cameras. One of such platforms is depicted in Figure~\ref{fig:charlie}. A collision tolerant platform called Gagarin, developed in collaboration with Flyability, was also used for the Tunnel and Urban events~\cite{tranzatto2022cerberus}. Identifying the need for systems capable of navigating more constrained passages, we designed and developed a new collision-tolerant platform, named RMF-Owl~\cite{depetris2022rmfowl}, to traverse extremely narrow settings. In addition to RMF-Owl, we developed the larger and longer range Kolibri flying robot for the Final Event, although it was not deployed due to the narrow settings of the environment. The robots developed for the Final Event are depicted in Figure~\ref{fig:rmf_kolibri}. Subsequently, we detail RMF-Owl and Kolibri, while the other two types of robots are documented in~\cite{tranzatto2022cerberus}.

The RMF-Owl was custom-designed to have a compact form-factor and carried a Khadas VIM3 Single Board Computer incorporating \si{4}{x} \SI{2.2}{\giga\hertz} Cortex-A73 cores, paired with \si{2}{x} \SI{1.8}{\giga\hertz} Cortex-A53 cores, alongside a Neural Processing Unit (NPU) as the main computation unit. This computer was used to run the entire autonomy stack with modified lightweight versions of our localization and mapping framework (c.f. section~\ref{sec:compslam}), the exploration planner (c.f. section~\ref{sec:gbplanner2}) and device drivers for the various sensors alongside the networking components. The neural network for artifact detection was executed on the NPU to gain a speedup during inference (c.f. section~\ref{sec:artifact_inference}). Its sensor payload included an ``Ouster OS0-64'' or ``Ouster OS1-64'' LiDAR (different robot versions were developed) featuring wide vertical \acp{FOV} of \SI{90}{\degree} and \SI{45}{\degree} respectively and a horizontal \ac{FOV} of \SI{360}{\degree}. Other sensors included an IMU, and a FLIR ``Blackfly S'' \SI{0.4}{\mega\px} color camera with a horizontal and vertical \ac{FOV} of $[85,64]^\circ$, the latter used for object detection. The outer dimensions of the robot were \SI{38}{\centi\meter} x \SI{38}{\centi\meter} x \SI{24}{\centi\meter} and thus alongside its sensing suite was rendered suitable for going through staircases, vertical shafts, and narrow passages while detecting artifacts along the course. Additionally, to traverse and detect objects in dark environments, RMF-Owl was equiped with an off-the-shelf LED lighting system, increasing the visual range and enhancing the environment texture. The total weight of the RMF-Owl Aerial Scout, shown in Figure~\ref{fig:rmf_owl}, was \SI{1.46}{\kilo\gram} with a maximum flight time of \SI{10}{\minute}.

The larger Kolibri, shown in Figure~\ref{fig:kolibri}, is based on a \num{5}DoF multi-directional tricopter platform from Voliro, which we customized with full autonomy for long-range and vertical exploration missions. Its front propellers are fully-actuated, enabling it to pitch independently of translation and affording good flight stability near surfaces. With a takeoff weight of \SI{6}{\kilo\gram}, it carried a sensor payload similar to that of the ANYmals, using an ``Alphasense Core'' unit with three color and three monochrome cameras (each with \SI{0.4}{\mega\px})  for \SI{270}{\degree} artifact detection and multi-modal mapping. We paired these with the same high-power LEDs. However, to conserve power and weight for the aerial robot, the LEDs were flash synchronized with its cameras through the Alphasense Core, which also obviated the need for active cooling. It also carried a wide vertical \ac{FOV} ``Ouster OS0-128'' LiDAR. To support the computing needs of our large sensor payload, the robot hosted a 16-thread AMD 4800U with \SI{64}{\giga\byte} of RAM, as well as three Intel Neural Compute Stick 2 modules (one per color camera) for artifact detection. The maximum flight time of Kolibri was \SI{22}{\minute} with the full competition sensor payload.

\subsubsection{Subterranean roving robot}

\begin{figure}
    \begin{subfigure}[b]{.48\textwidth}
      \centering
      \includegraphics[width=0.85\textwidth]{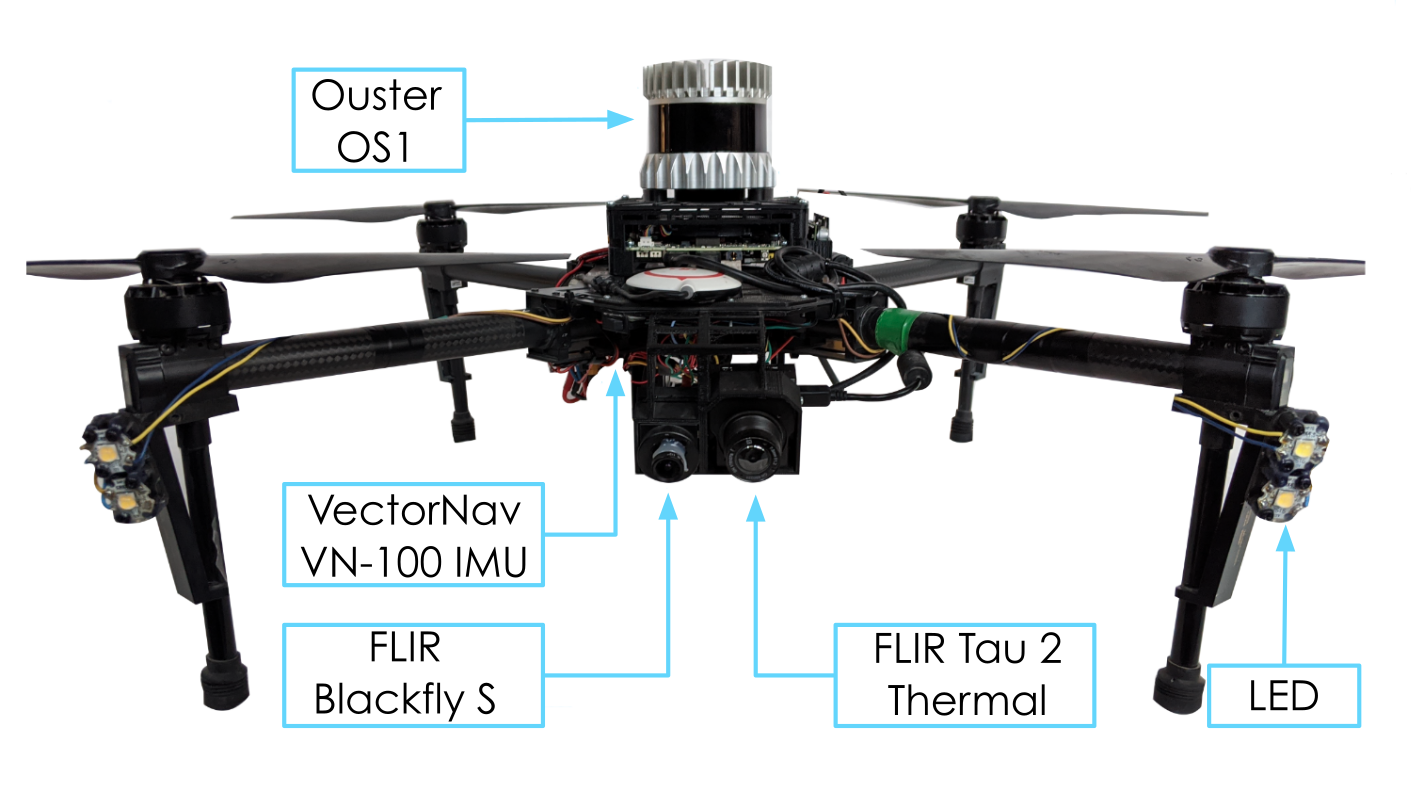}
      \caption{Aerial scout robot platform}
      \label{fig:charlie}
    \end{subfigure}
    \begin{subfigure}[b]{.48\textwidth}
      \centering
      \includegraphics[width=0.85\textwidth]{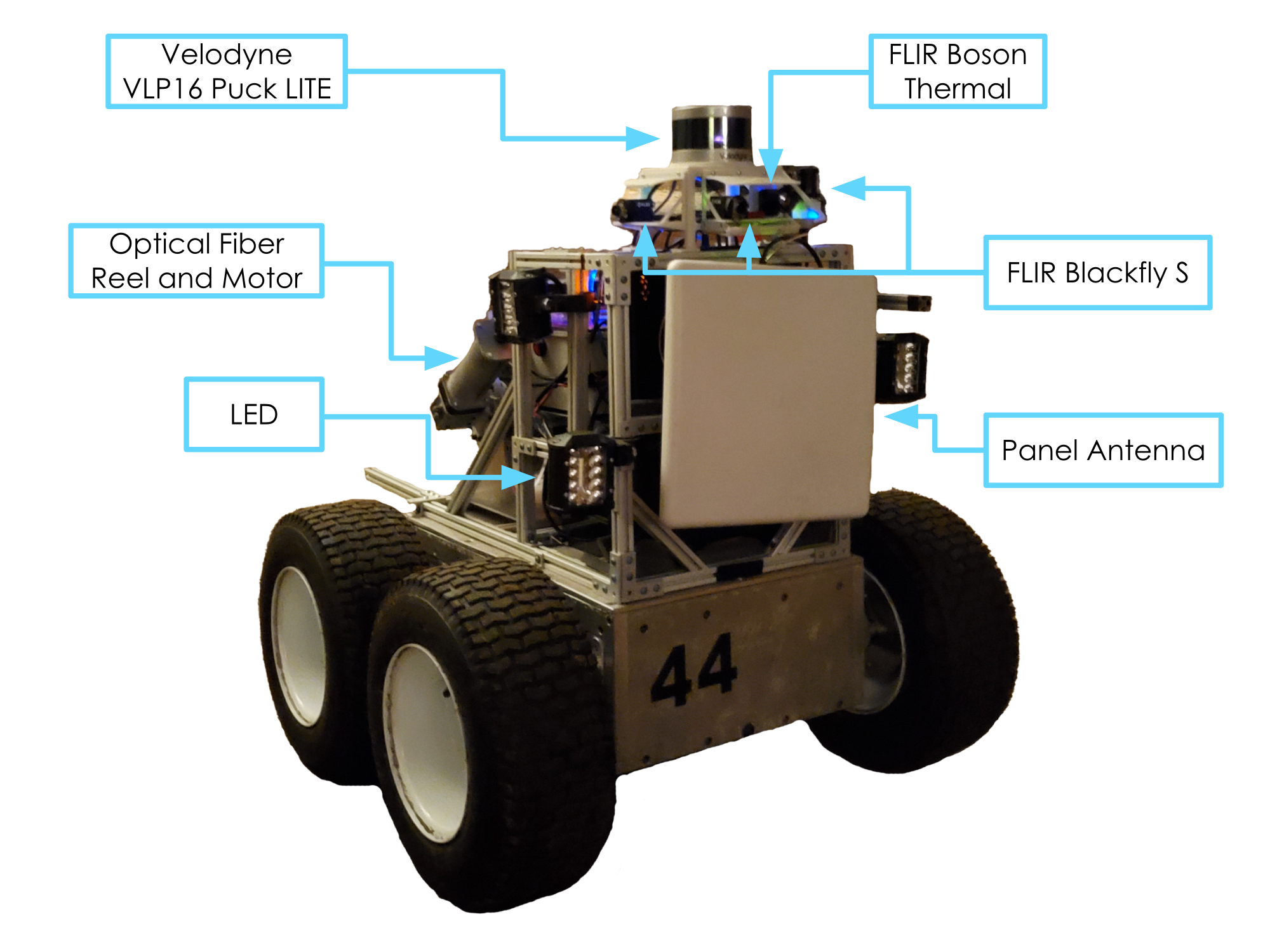}
      \caption{Roving robot platform}
      \label{fig:smb}
    \end{subfigure}
    \caption{The DJI Matrice $\textrm{M}100$-based aerial scouts, and the roving robot were used throughout the competition for all three events (Tunnel Circuit, Urban Circuit, and the Final Event).}
    \label{fig:smb_scouts}
\end{figure}

Finally, we developed a wheeled platform to act as a supporting agent to all the other robots and provide extended network connectivity into the course. Shown in Figure \ref{fig:smb}, this robot was based on an Inspector Bots Super Mega Bot (SMB) robot platform, with a modified chassis to mount equipment for extending network connectivity, a variety of sensors, an onboard computer, light sources, and a battery to power the components. The \SI{135}{\kilo \gram} robot hosted a Zotac VR Go Backpack PC having a $6^\textrm{th}$ generation Intel i7 CPU and an NVIDIA RTX $1070$ GPU inside a cage-like structure. We manufactured a two-level sensor mount to house four ``Blackfly S'' color cameras by FLIR, two ``Boson LWIR'' thermal cameras by FLIR, alongside a ``VN-100'' IMU by VectorNav, and a ``VLP16 Puck LITE'' LiDAR by Velodyne. The platform also featured a \SI{300}{\meter} long, ruggedized optical fiber cable, mounted on a controllable reel at the back of the robot. This cable was connected to the Base Station and the reel could automatically wind and unwind to prevent tangling into the wheels of the robot as it moved within the course. The wireless communication modules onboard this robot were connected to a \SI{5.8}{\giga\hertz} panel antenna, placed at the front of the robot. Accordingly, through the combination of the long optical fiber cable and the high-gain antenna, this system acted as a network range extender and provided high-bandwidth connectivity to the Base Station. In addition to the role of a network extender, the robot featured the necessary sensors and perception software, allowing it to operate independently as a backup platform. This robot was teleoperated by the Human Supervisor and did not operate autonomously.

%% file: 03_02_perception.tex
This section covers Team CERBERUS' perception pipeline, including onboard localization, multi-robot mapping, and terrain perception and classification. Firstly, we discuss the utilized sensor modalities with respect to how they were configured, synchronized, and calibrated.
Next, we detail the onboard complementary multi-modal localization and mapping as well as the elevation mapping pipelines.
Finally, we outline the multi-robot mapping approach combining the individual onboard estimations. It is noted that the competition necessitated that teams localize artifacts within a certain level of accuracy against groundtruth locations and report accurate maps. Alignment with the DARPA-defined coordinate frame (on which ground-truth is expressed) was achieved through the use of the ``DARPA Starting Gate'' which was at the entrance of the mission course. The DARPA Starting Gate offered multiple options for alignment including a) AprilTags, b) retro-reflective targetsm c) survey prisms and d) survey markers. Team CERBERUS performed alignment with the DARPA Starting Gate using only the AprilTags. 

\subsubsection{Sensor setup and calibration}
Since heterogeneous sensor setups can impose significant challenges on a multi-modal and multi-robot mapping approach, all legged robots as well as the Voliro-T Kolibri were equipped with the same sensory system (cf. section~\ref{sec:anymal_c_subt} and section~\ref{sec:aerial_robots}).
This included the same cameras, lenses, synchronization logic and illumination hardware.
In the context of robotic perception, this enables simpler parameter tuning as well as improved inter robot loop closure detection. Tailored to miniaturized system design and prolonged flight endurance for their overall size, the DJI Matrice 100 based aerial scouts and the collision-tolerant flying robots of our team utilized a LiDAR system, alongside a ``Blackfly'' color camera by FLIR and a ``VN-100'' IMU by VectorNav. One of the Matrice 100-based systems further integrates a ``Tau2'' thermal camera by FLIR.

Another important factor is the time synchronization between the sensors and the robots.
The ANYmal robots where equipped with a VLP-16 Puck LITE LiDAR that timestamped pointclouds based on arrival time, whereas the Voliro-T featured an Ouster OS0 which was synchronized to the computer clock via \ac{PTP}. 
Similarly, the Alphasense Core was also synchronized using \ac{PTP} on each robot. For the DJI Matrice 100 based aerial scouts and the collision-tolerant flying robots, a software-synchronized approach was employed in which the offset was estimated among the time-of-arrival of sensor packages. 

\textbf{Sensor calibration}\\
Reliable sensor calibration is vital to most sensing-based algorithms. Our calibration efforts targeted mainly intrinsic camera calibration as well as extrinsic camera-to-camera and camera-to-LiDAR calibrations.
Calibration-critical multi-sensory \ac{SLAM} applications are, for example, M3RM (Section \ref{sec:m3rm}) and CompSLAM (Section \ref{sec:compslam}). Furthermore good calibrations are important for the ray-casting of detected artifact locations into the volumetric map of the exploration planner (Section \ref{sec:artifact_filtering}) as well as camera-based AprilTag detection for anchoring the robot's state estimation in a global reference frame.

\emph{Intrinsic calibration} - To perform intrinsic camera calibration we used the visual-inertial calibration toolbox Kalibr~\cite{kalibr_furgale}.
Given the wide-angle field-of-view of the cameras that are part of both the Alphasense sensor mount and the visual cameras onboard the multirotor aerial scouts and collision-tolerant flying robots of our team, we found a large enough rigid calibration target to be key to get good calibration results in relevant distances from the camera. We obtained good results with a grid pattern of Apriltags~\cite{apriltags_olson} with \num{6} rows, \num{6} columns and tag side lengths of around \SI{10}{\centi \meter}.
Furthermore, the Kalibr library provides a useful tool to examine camera focus. The method is based on calculating entropy in the image frequency domain~\cite{kalibrfocus_kristian}.
Intrinsic calibration of LiDARs and IMUs were provided by manufacturers directly and were considered sufficiently accurate to be used out of the box. Intrinsic calibration of thermal cameras is achieved with the use of a specialized heated calibration target.

\emph{Extrinsic calibration} - For camera-to-camera extrinsic calibrations we also utilized the Kalibr toolbox. Alongside the tools for intrinsic camera calibration it allows for jointly estimating temporal offsets and sensor displacements using continuous-time batch estimation. It features options for single camera and IMU extrinsic calibration as well as multi-camera and IMU calibration. The latter is useful if camera \acp{FOV} are majorly overlapping and yields increased accuracy by performing co-optimization over all provided data. 
We considered using optimization based methods, such as presented in~\cite{LiDARcamcalib_huang}, to perform camera-to-LiDAR extrinsic calibration. In their work they show results using a \num{32}-Beam LiDAR. However we observed that with a \num{16}-Beam LiDAR the optimization problem tends to be insufficiently constrained. This is likely due to the sparsity of the LiDAR measurements. Accordingly, both for the robots with \num{16}-Beam LiDARs and for our overall set of systems, we chose to use a self developed manual calibration method that allows to visualize projected LiDAR points in the camera image and to thereon dynamically modify extrinsic LiDAR-to-camera transforms.

\subsubsection{Complementary multi-modal localization and mapping}\label{sec:compslam}
Robust localization and mapping are among the core components of the robot autonomy pipeline to ensure reliable operation in complex subterranean environments. They are responsible for providing consistent and low-latency localization to facilitate low-level, real-time operations, such as robot control. In addition, they are also responsible for producing an accurate and informative map to assist high-level autonomy tasks, such as path-planning in complex large-scale environments.

\begin{figure}[h]
    \centering
    \includegraphics[width=0.75\linewidth]{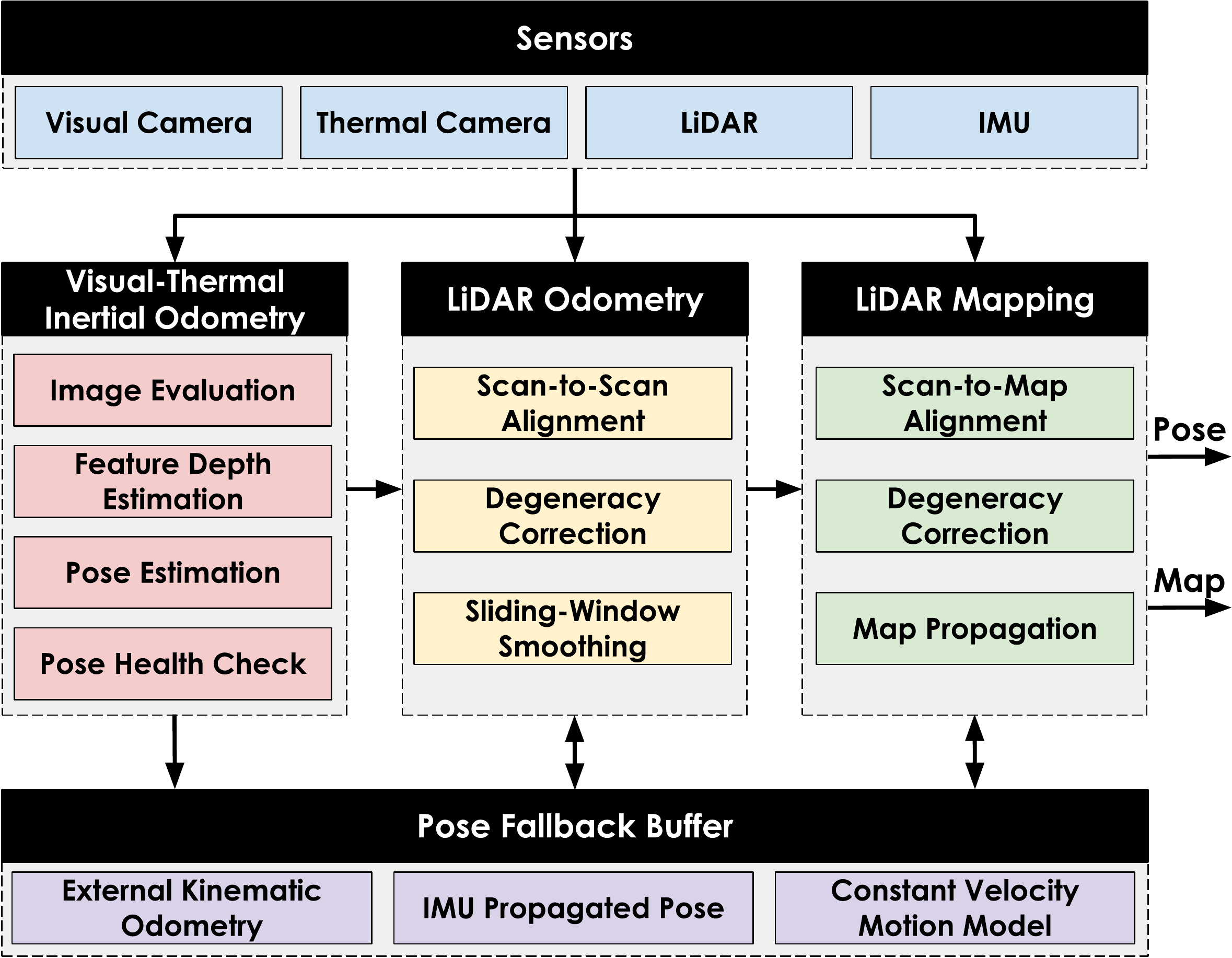}
    \caption{Overview of the onboard complementary multi-modal localization and mapping approach (CompSLAM).}
    \label{fig:compslam_overview}
\end{figure}

In response, the robots of Team CERBERUS deployed an onboard complementary multi-modal localization and mapping approach (CompSLAM)~\cite{Khattak2020sensorfusion}, that ensured reliable and independent robot operations without any dependence on communication with the Base Station. An overview of the proposed approach is shown in Figure~\ref{fig:compslam_overview}. CompSLAM utilizes a hierarchical sensor fusion strategy which refines individual or multi-modal pose estimates as they propagate through the framework, ensuring that a robot always has an updated pose estimate and a map available. Furthermore, CompSLAM's modular design made it flexible to adapt to the sensing and compute variances across the robots of Team CERBERUS. Hence, the method allowed to take advantage of the complementary nature of multi-modal sensors for operation resilience, while maintaining the ability to scale cognizant to the available robot compute capabilities to ensure real-time performance across diverse environments and robot configurations.

To provide redundancy against cases of sensor data degradation and estimation failures, CompSLAM performs fusion of visual and thermal imagery, LiDAR depth data, inertial cues, and kinematic pose estimates (when available) in a hierarchical manner. For the fusion of visual, thermal and inertial data, CompSLAM utilizes an Extended Kalman Filter (EKF) based estimator that extends the work of~\cite{Bloesch2017rovio} to simultaneously operate on both visual and full-radiometric thermal images, as described in~\cite{Khattak2019rotio,ktio2020}. 

The Visual-Thermal-Inertial Odometry (VTIO) estimator tracks and balances a fixed number of features across multi-spectral images by evaluating the quality of each image subject to a spatial and temporal entropy criterion~\cite{khattak2019visual}. This results in a small number of features required to be tracked as part of the EKF state-vector, keeping the computational costs tractable for real-time operation. Furthermore, to maintain low computational costs while facilitating robust scale estimation in degraded environments, sparse depth from instantaneous LiDAR scans is exploited for feature depth initializations and updates. For features enclosed by at least two or more LiDAR scans, a local-flatness check is performed to ensure depth consistency. If passed, features are initialized with the median neighborhood depth acquired from LiDAR scan and associated with a smaller variance on the feature depth to improve the overall scale convergence of the estimator. For features initialized without LiDAR depth, if a LiDAR scan sweeps over a feature due to robot motion during the mission, the feature depth and variance are updated in a probabilistic manner~\cite{limo}. Although, sparse in nature, the instantaneous LiDAR scans are used instead of the onboard map for depth association to build redundancy into the system in case of a mapping failure. 

For building the onboard map, the LiDAR scans along with the VTIO odometry are utilized by the Laser Odometry (LO) and the Laser Mapping (LM) modules. Before utilizing VTIO output as a prior, the LO module evaluates the VTIO health by ensuring that it is within the motion-bounds set by the robot controller. Furthermore, the covariance growth rate of the VTIO prior is evaluated to be within an acceptable margin by employing the D-Optimality criterion~\cite{dopt}. In case the VTIO health checks fail, the LO module reverts to a set of fallback solutions, in order, for the provision of prior - kinematic odometry (in case of ANYmal~\cite{tsif}), IMU propagated pose or a constant velocity motion-model. Using the obtained prior and consecutive LiDAR scans, the LO module performs a scan-to-scan alignment by minimizing point-to-line and point-to-plane distances, similar to the work of~\cite{Zhang2014loam}. The quality of scan alignment is checked by evaluating the minimum eigenvalues of the optimization Hessian matrix~\cite{jhang2016degeneracy}. If optimization degeneracy is detected, the ill-conditioned dimensions of the LO estimation output are replaced with the prior's estimates. The LO estimates are then fused with inertial measurements and smoothed using a sliding-window estimator scheme to improve pose consistency and gravity alignment. The sliding-window estimator is implemented as a factor-graph using the work of~\cite{gtsam}. Furthermore, when the robot is stationary, zero-velocity factors are added to improve the robot pose and IMU bias estimation.

The refined LO pose estimate is then provided to the LM module which performs a scan-to-submap alignment to produce an onboard map. For computational efficiency and, in particular, to improve the speed of establishing point correspondences, the onboard map is divided into blocks of volume $10\textrm{m}^3$. Each block is identified by a unique hash calculated from the location of each block's center with respect to the map origin. Furthermore, each map block is sub-divided into line and plane feature sub-blocks. Upon arrival of a LiDAR scan and associated LO pose estimate, all sub-blocks within the \ac{FOV} of the current scan are identified and independent KD-Trees are built for each block in parallel to allow parallel search for point correspondences. Once a scan-to-submap alignment is performed, a health check similar to that of the LO module is performed to determine the quality of the LM output. In case LM optimization is ill-conditioned, a previous healthy estimate in the sensor-fusion hierarchy is searched and utilized for merging the current scan to update the onboard robot map.

\subsubsection{Elevation mapping}
\begin{figure}[h]
    \centering
    \includegraphics[width=0.9\linewidth]{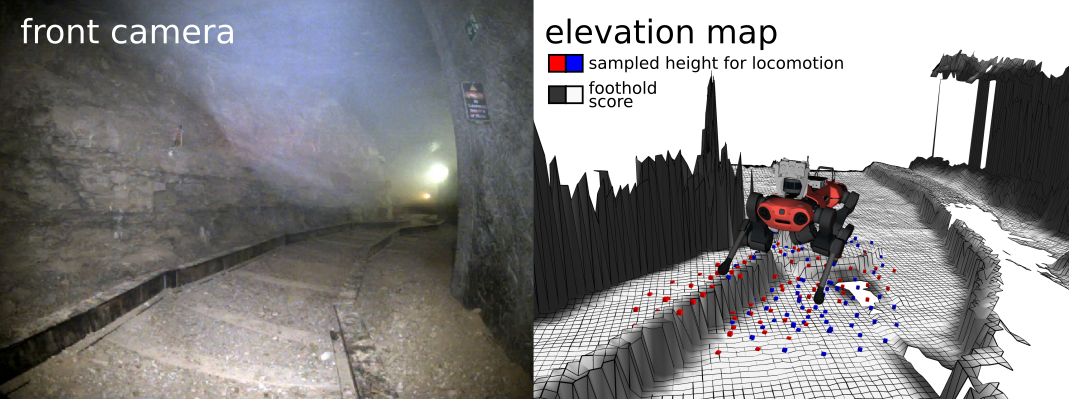}
    \caption{ANYmal's elevation map. The terrain map was built onboard the GPU and used by the learning based locomotion controller and the ANYmal navigation planner. The height information is used by the learning based controller to perform terrain-aware locomotion: blue and red dots indicate samples for left and right feet, respectively. The local planner utilizes the traversability layer (shown as foothold score) for path planning: white areas indicate patches considered as steppable, whereas black ones are not.
    }
    \label{fig:elevation_map_locomotion}
\end{figure}

An elevation map is a $2.5$D representation where each grid cell holds the height of the terrain. We used this map representation for local navigation and perceptive locomotion of the quadrupedal platform. It is built in a robot-centric fashion from depth sensor readings and robot odometry. We developed a GPU-based software for efficient processing~\cite{miki2022elevation}. In addition to point cloud registration it performs a visibility cleanup check to quickly remove map artifacts. The method performs raycasting from the sensor origin to each measurement point and clears cells if the ray passes below its stored elevation. To avoid issues with structures thinner than the map resolution, we do not clear a cell if it is currently observed by a measurement.
The terrain map used in the SubT Challenge finals had dimensions equal to \SI{8}{\meter} x \SI{8}{\meter} with a resolution of \SI{4}{\centi\meter}. These settings reflected the sensor limitations and represent a trade-off between accurately capturing the terrain's surface and keeping the computation cost to a reasonable level for onboard processing, thus making it possible to create a real-time terrain representation onboard the ANYmal's \ac{GPU} utilizing the onboard depth sensor measurements. The elevation map was computed at \SI{20}{\hertz} for Explorer robots, given each RS-Bpearl publishing at \SI{10}{\hertz}, and at \SI{45}{\hertz} for Carrier robots, given each depth camera publishing at \SI{15}{\hertz}.
The ANYmal's locomotion controller samples the terrain height in a circular pattern around each foot, as shown in Figure~\ref{fig:elevation_map_locomotion}. This height information is used by the controller to perform terrain aware locomotion.
We integrated a learning-based foothold score estimator~\cite{wellhausen2021rough}, which gives a foothold score value based on the local geometry and was used by our navigation planner. The foothold score is indicated by black (low) and white (high) color in Figure~\ref{fig:elevation_map_locomotion}. Furthermore, we enhanced the elevation map through improvements such as upper bound calculation similar to~\cite{hines2020virtual}, which helps planning in the presence of occlusions.

\subsubsection{Multi-robot mapping}\label{sec:m3rm}
Beyond individual onboard localization and mapping, Team CERBERUS further developed a solution for multi-robot mapping within the complex and large-scale underground environments. The long-term stability of onboard maps is a pertaining issue in \ac{SLAM}. Especially given the often limited understanding of the environment, choosing how to fuse new information and decide what is adequate to keep is not trivial.
A centralized server, accumulating all available onboard maps, can remedy this by providing a globally consistent environment estimate. 

\begin{figure}[h]
    \centering
    \includegraphics[width=\linewidth]{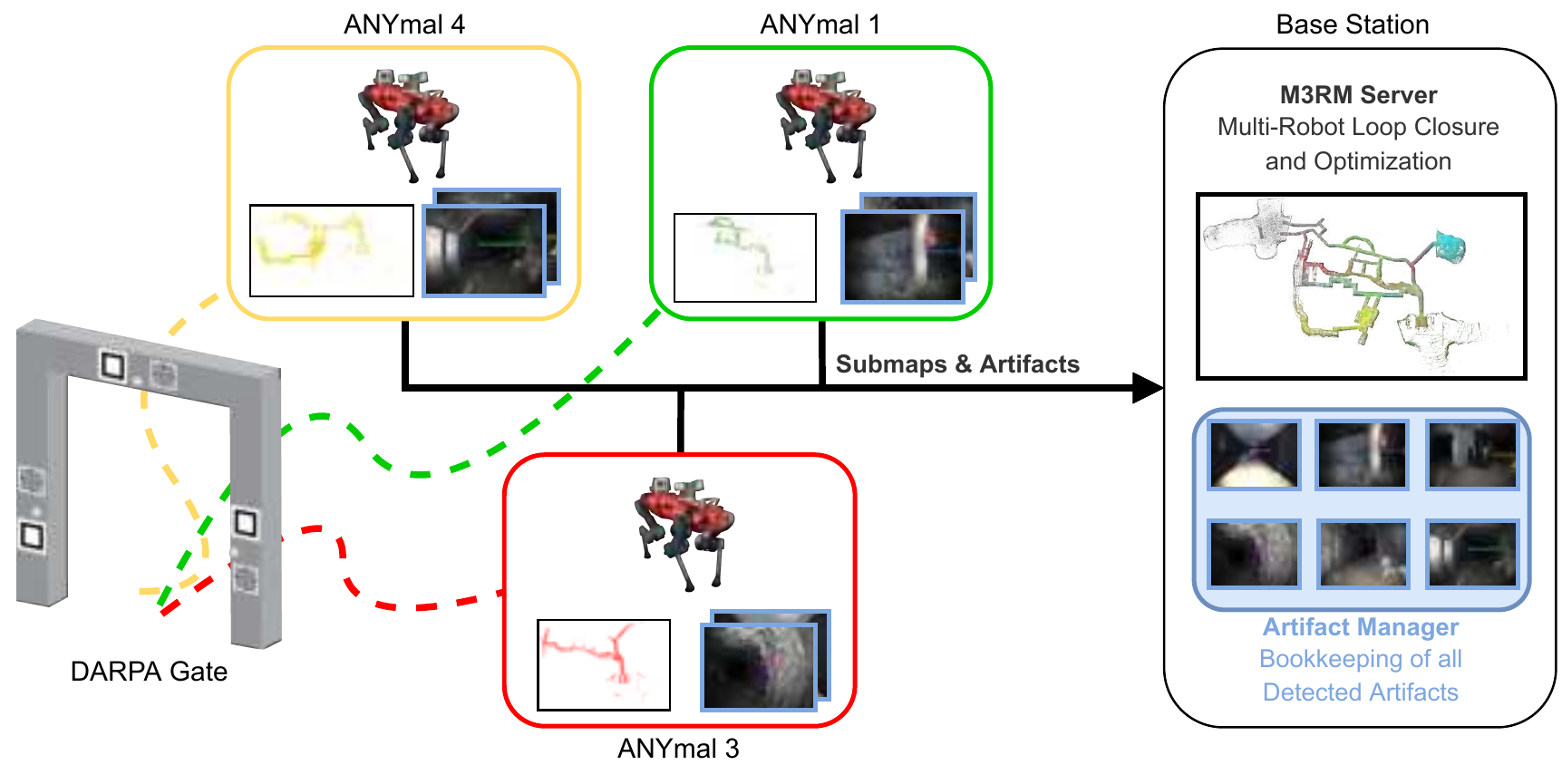}
    \caption{Overview of the employed centralized multi-robot mapping approach. Each robot aligns its estimations (poses and map) to the DARPA Gate and maintains its own onboard map independent of the other robots while exploring the environment. When in communication range to the Base Station, the onboard maps and the detected artifacts are transferred to the Base Station where the server merges the robot maps into the global multi-robot map. All artifact detections are accumulated at the Base Station and visualized to the Human Supervisor for verification.}
    \label{fig:mapping_overview}
\end{figure}

\textbf{Centralized mapping approach}\\
We designed and developed a multi-modal and multi-robot mapping (M3RM) framework using a centralized topology~\cite{tranzatto2022cerberus}. 
Overall, M3RM comprises of two core components, the M3RM node that creates a factor graph using the onboard complementary localization and mapping solution, and the M3RM server that maintains and optimizes the multi-robot global map. 
An overview of our developed mapping approach is given in Figure~\ref{fig:mapping_overview}.

Each M3RM node on each robot tracks visual features, i.e., BRISK~\cite{leutenegger2011brisk}, using all camera images and triangulates the tracked features to landmarks using the onboard localization. The resulting visual map is then represented as a factor graph and transmitted as chunks (submaps) to the M3RM server at the Base Station.
Before transmission, the point clouds are subsampled, compressed, and attached to the factor graph using their timestamp and extrinsic calibration. Accumulating all point clouds per submap, the required size is approximately \SI{2}{\mega\byte} per transmission. The M3RM server used these point clouds to register spatially close poses with each other. 

Finally, robots with an established connection to the Base Station transmitted submaps consisting of the onboard odometry estimates and subsampled sensor data to a centralized mapping server.
A synchronization logic handles partial transmissions and connection losses and ensures that only completed submap transmissions will be integrated into the multi-robot map. 
Moreover, the submap transmissions can be batched and compressed to increase the efficiency and to take the network's bandwidth limits into account.

The M3RM server then builds a multi-robot factor graph using all the available sensor modalities and performs intra-/inter-robot loop closure detection using vision and LiDAR. 
Specifically, visual loop closure searches globally for matching descriptor candidates, while LiDAR loop closures are searched locally between spatially close poses and submaps.
For the latter, additional LiDAR constraints are added to the factor graph by accumulating consecutive scans within a submap and aligning to spatially close submaps using ICP. Each finished submap transmission is processed on its own before being integrated into the global multi-robot map. 
Among other operations, this includes a visual landmark quality check, visual loop closure detection, LiDAR loop closure detection, and submap optimization.
The processing of each submap is independent with respect to other submaps and thus, the M3RM server will run multiple operations concurrently.  After processing the individual submaps, they are merged into the global multi-robot map which the M3RM server continuously processes and optimizes in a similar vein.
The M3RM server runs indefinitely a fixed set of operations, i.e., visual loop closure, LiDAR loop closures, and optimization.
However, in this case the operations consider the scope of the global multi-robot map.

Finally, the global multi-robot optimization is limited to a maximum of two minutes to guarantee fast iterations.
This allows the M3RM server to integrate the most recent
information that might contain crucial loop closures quickly and provide the most recent data to the Human Supervisor for ad-hoc mission planning and execution.

The M3RM server also enables to provide a localization feedback to the individual robots about their onboard position~\cite{bernreiter2022collaborative} to circumvent inconsistencies between them.
The robots and server can then exchange information in a bi-directional way to not only improve the onboard state estimation, but also serve as a catalyst for collaborative multi-robot planning.

\textbf{Long-term operation}\\
The plethora of sensor data being transmitted from the robots to the mapping server is filtered and subsampled to provide a high-quality and rapid global estimate of the environment. 
Nevertheless, the amount of data from multiple robots over a longer time span can still significantly slow the operation time.
Therefore, the Human Supervisor can choose between multiple performance profiles at the mapping server to adapt the global multi-robot optimization to the current situation. 
In situations where all robots are not within connection range, the mapping server can increase its computational load, whereas in scenarios where robots have re-established connections should lower its capacities to provide rapid updates.

%% file: 03_03_autonomy.tex
The autonomy solution of Team CERBERUS was tailored to the challenges of the subterranean environments. Our aim was to enable resilient autonomy with minimal human supervision in very large-scale settings with difficult terrain, narrow passages, and multi-branching, multi-level topologies. To realize this, all robots at their core utilized a Graph-based Exploration Path Planner (GBPlanner2)~\cite{GBPLANNER2COHORT_ICRA_2022} for efficient exploration of the environment. The exploration planner was designed to work as-is across all of the team's robotic platforms, with the only change being selected robot-specific parameters and features. Furthermore, robot specific motion planning and path tracking algorithms were used to follow the path provided by the exploration planner or track the waypoints given by the Human Supervisor. 

\begin{figure}[h]
    \centering
    \includegraphics[width=0.85\textwidth]{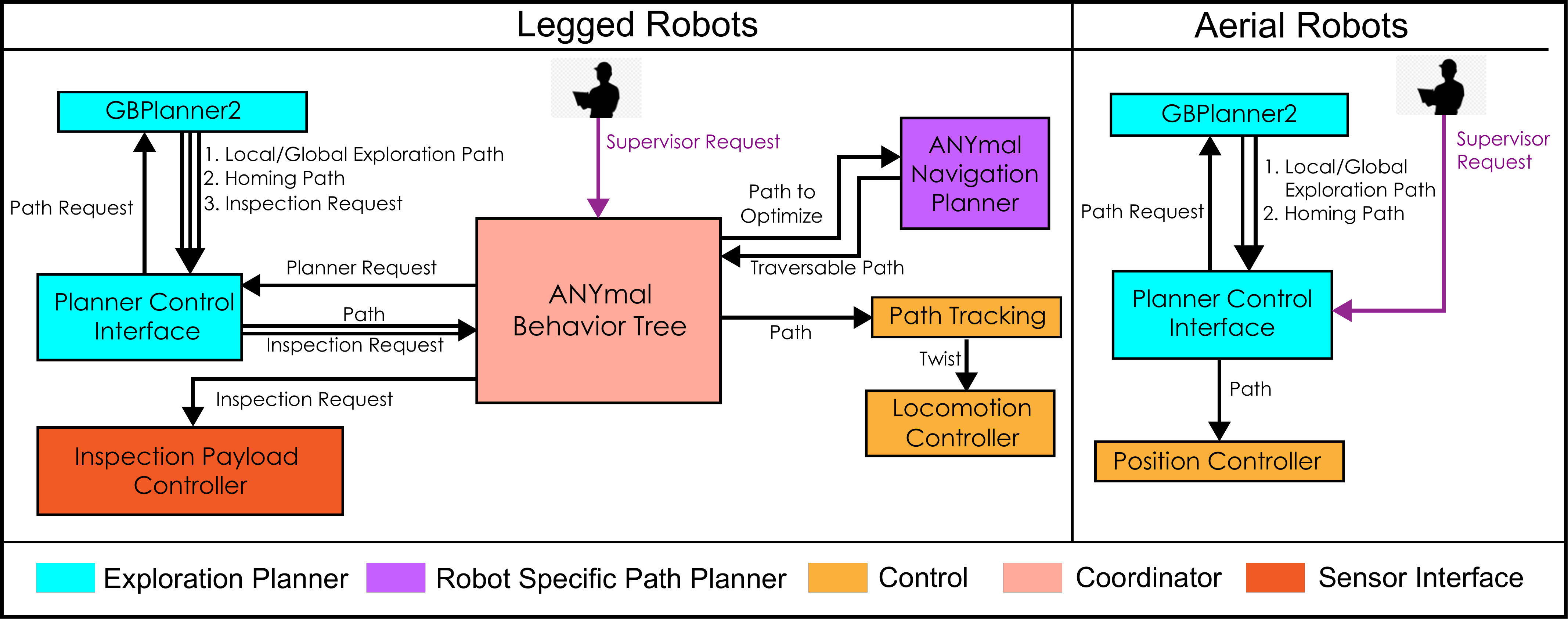}
    \caption{Overview of the Autonomy Stack of Team CERBERUS. The diagram shows a high level overview of the interaction between the sub-modules within the Autonomy Stacks for both legged and aerial robots.}
    \label{fig:autonomy_overview}
    \vspace{-2ex}
\end{figure}

The autonomy architecture of the legged and aerial robots is shown in Figure \ref{fig:autonomy_overview}. A coordination module, based on \ac{BT} (see section~\ref{sec:anymal_behavior_tree}), manages the switching between different tasks of the ANYmal's autonomy stack.
All the requests from the Human Supervisor are handled by this module. A robot specific \ac{PCI} provides APIs for accessing the functionalities of GBPlanner2. The autonomy stack of the aerial robots is relatively simpler with the Human Supervisor requests being handled directly by the \ac{PCI}.

\subsubsection{Exploration planner}\label{sec:gbplanner2}
At the core of the CERBERUS autonomy lies the Graph-based Exploration Path Planner (GBPlanner2)~\cite{GBPLANNER2COHORT_ICRA_2022} that uniformly guides both legged and aerial robots. GBPlanner2 extends the work of GBPlanner1~\cite{GBPLANNER_JFR_2020} (used in the Tunnel and Urban Circuit events) to better handle slopes and uneven terrain for legged robots, as well as by providing features for reduced computation times for computationally constrained \ac{MAVs}. The planner uses an incrementally built volumetric map using the onboard depth sensors. In the Final event, all robots of Team CERBERUS used ``Voxblox''~\cite{oleynikova2017voxblox} as the volumetric mapping framework.

\begin{figure}[h]
    \centering
    \includegraphics[width=0.9\textwidth]{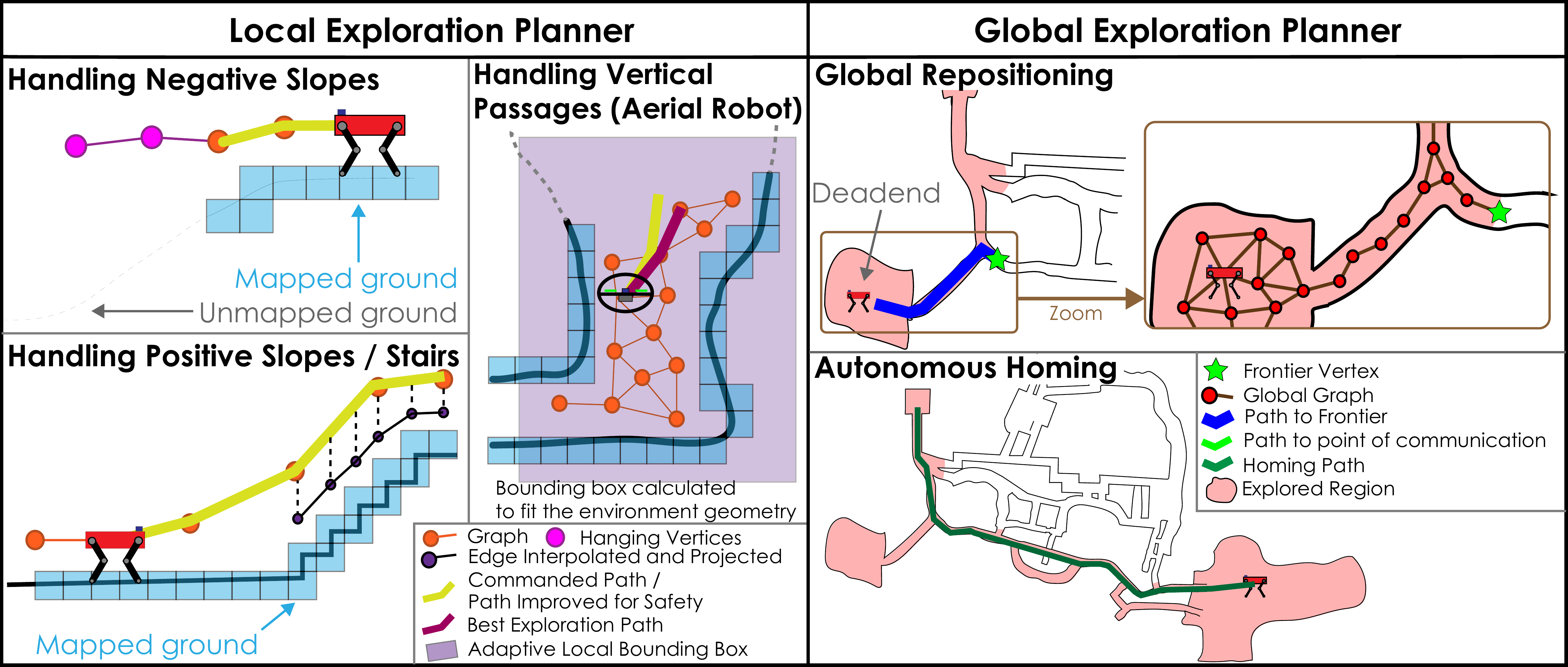}
    \caption{Graphical overview of the exploration planner's bifurcated local/global architecture. The main features involved in calculating the local and global exploration paths are shown.}
    \label{fig:gbplanner_fig}
\end{figure}

GBPlanner2 operates in a bifurcated local/global planning architecture. 
The local planner guides the robot to efficiently explore within a local volume around the robot. It first samples a collision-free graph (called ``local graph'') inside a local subspace of the map (hereafter referred to as the ``local bounding box'') that is calculated based on the local environment geometry. For legged robots, each vertex and edge in the graph should not only be collision-free but also have traversable supporting ground under it. This is evaluated by projecting the sub-sampled edge (including its end vertices) downward on the map and checking if a) occupied voxels exist in the map below the projected points and b) the inclinations of the segments of the projected edge are within the permissible limit for the robot (Figure~\ref{fig:gbplanner_fig}). To tackle negative obstacles, the planner retains certain vertices (called ``hanging vertices'') in the graph that are in free space but don't have supporting ground below them if at least one of the edges connecting to them has its inclination within the allowed limit. However, the edges connecting to these vertices are not commanded to the robot. Next, using Dijkstra's algorithm, the planner finds the shortest paths from the current robot location in this graph. The path that maximizes an information gain relating to the expected unknown volume that will be mapped by traversing that path is selected and commanded to the robot. \cite{GBPLANNER_JFR_2020} provides further details on the formulation of the information gain, alongside further steps to refine the path quality. Additionally, to reduce the computation time, GBPlanner2 also provides the functionality to calculate this gain only at the end vertices of the shortest paths as well as to cluster these vertices and calculate the gain only for one vertex in each cluster.

\begin{figure}[h]
    \centering
    \includegraphics[width=0.9\textwidth]{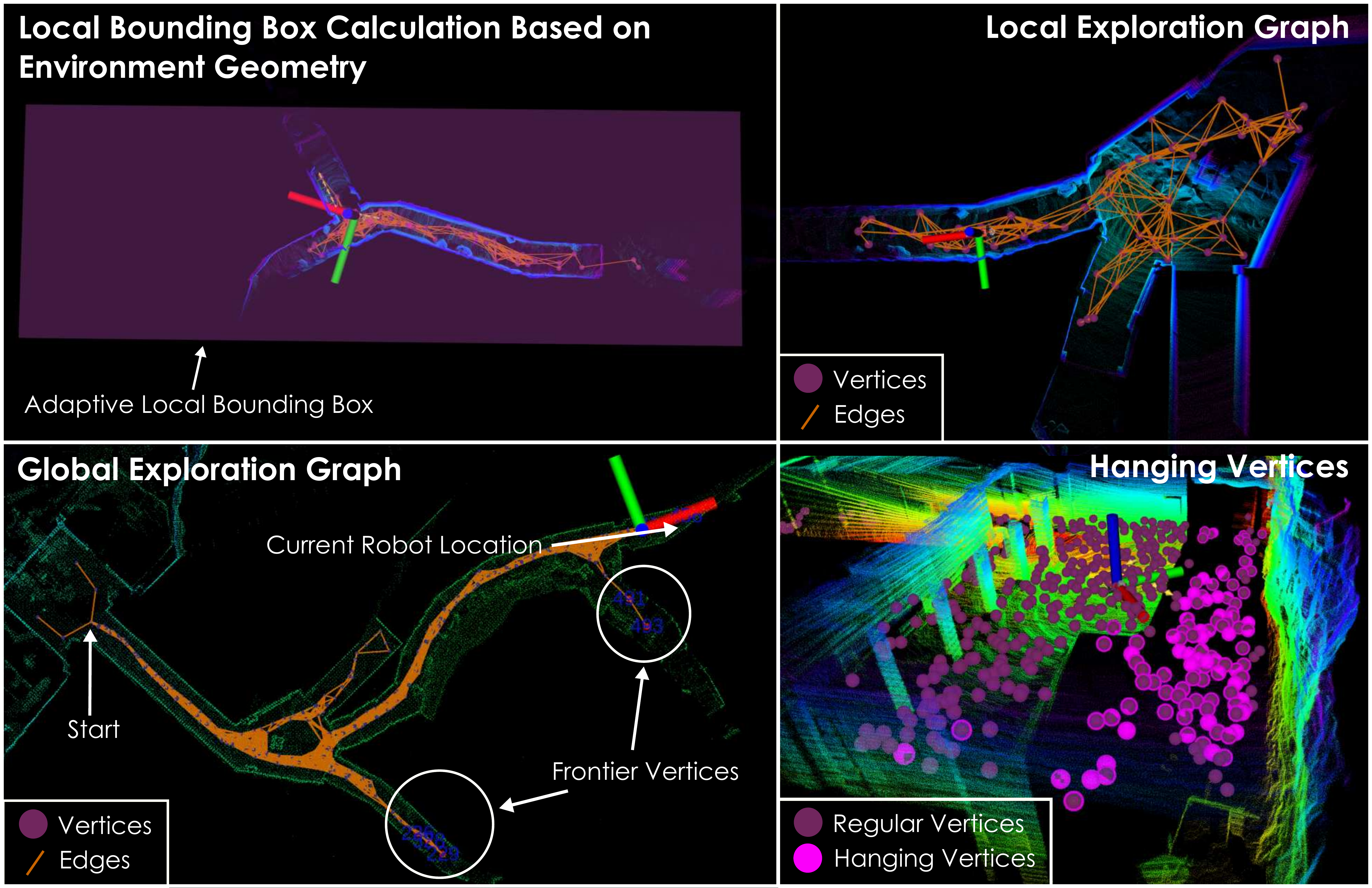}
    \caption{Visualizations of different components of the exploration planner on data. These include the adaptive local bounding box, the graph spanned by the local planner, the graph spanned by the global planner along with the frontier vertices, and the hanging vertices (feature for legged robot path planning).}
    \label{fig:gbplanner_fig_data}
\end{figure}

Due to the complex topology and challenging terrain of the subterranean environments, the local planner might not be able to find a path leading to further exploration. In such a case, the global exploration planner is responsible to re-position the robot to unexplored areas of the map. This is done by exploiting an incrementally built sparse graph (called ``global graph'') that spans across the known map and stores high information gain vertices called the ``frontier vertices'' along with their corresponding paths. Furthermore, at each local planning iteration the global planner calculates a path from the current robot location to the home location and commands the same to the robot if the remaining endurance is not enough to perform one more local exploration planning step and return home. Additionally, the global planner also supports calculating paths to goal points provided by the Human Supervisor or the ANYmal's behavior tree. Figure \ref{fig:gbplanner_fig} provides a graphical overview of the exploration planner. Similarly, visualizations of various components of GBPlanner2 on data are shown in Figure~\ref{fig:gbplanner_fig_data}.

\subsubsection{ANYmal navigation planner}

To avoid local obstacles and follow the exploration path or the exploration waypoints provided by
the Human Supervisor safely, we developed a navigation planner specifically for legged robots~\cite{wellhausen2021rough, wellhausen2022planner}.
This navigation planner could locally adjust
the exploration path to better cope with the properties of the surrounding terrain, given that legged robots have much higher mobility capabilities than wheeled or tracked platforms due to their ability to step over obstacles.
Therefore, classic navigation approaches, which compute a single traversability value per terrain patch independent of robot orientation or motion direction, are limiting.
Other teams which deployed both wheeled and legged systems deployed such a solution~\cite{fan2021step}.

\begin{figure}[H]
    \centering
    \includegraphics[width=0.9\linewidth]{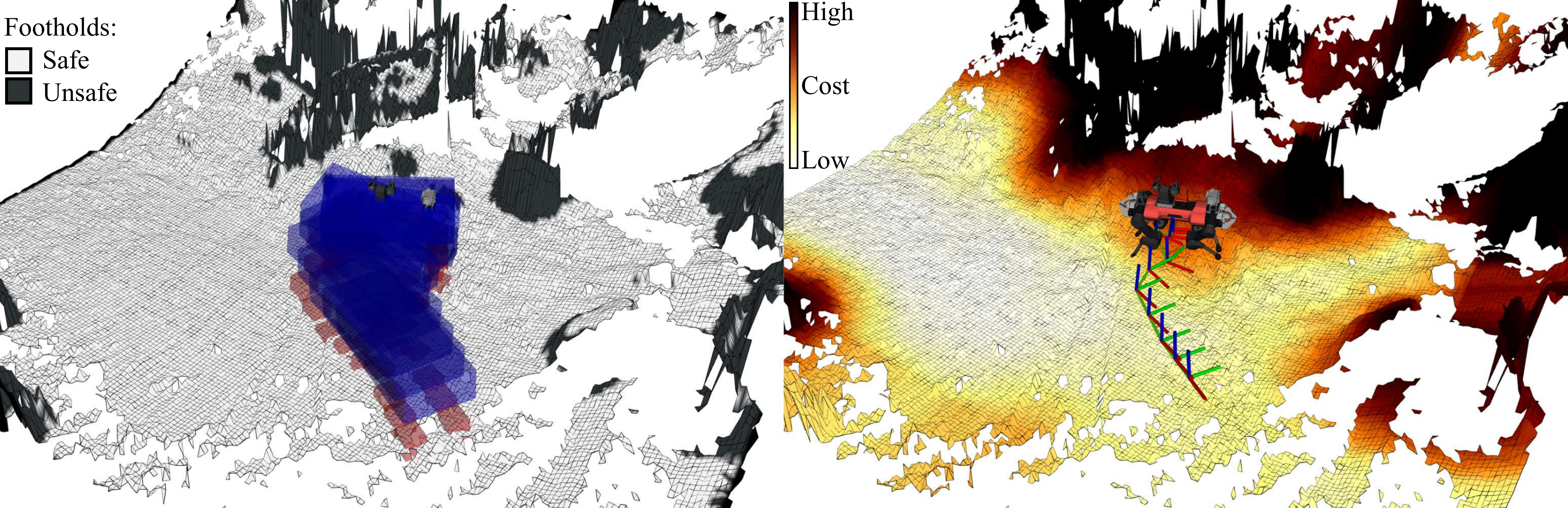}
    \caption{ANYmal navigation planner in the cave section during the Prize Round of the Final Event. (Left) Blue torso collision boxes need to be collision free while red reachability boxes need to be in contact with the environment. Black parts of the height map are considered invalid for footholds. (Right) A learned motion cost (white - low cost, black - high cost) is used to limit locomotion risk and optimize the path for both locomotion risk and traversal time.}
    \label{fig:art_planner}
\end{figure}

Our planner is based on LazyPRM*~\cite{hauser2015lazy} with a custom sampling scheme, which extends poses sampled in 2D to 3D using map information, and favors sampling in regions with low planning graph node density~\cite{wellhausen2021rough}.
It uses a reachability abstraction of the robot by decomposing it into a torso collision volume and one reachability volume per leg.
Since a legged robot needs to touch the ground to locomote, we assume that a valid robot pose needs to have all reachability volumes in contact with the environment, while the torso remains collision-free.
To avoid practically infeasible poses, like walking on a wall, we train a very small \ac{CNN} on manually labelled data to predict a foothold score from a height map. 
We can then exclude geometry with low foothold score from collision checking for reachability volumes to disallow stepping on invalid terrain.
Because the network only has \si{120} parameters, it can be trained on only \si{20} labelled height maps.
This generally produces feasible paths, but since a shortest path cost function would have no notion of risk, caused by uncertainty in the environment, imperfect control and path following, and various other sources, we additionally use a learned motion cost~\cite{wellhausen2022planner}.
To this end, we train a \ac{CNN} to predict the motion risk, energy and time cost given a height map and a relative $2$D goal pose~\cite{guzzi2020path, yang2021real}. 
Data is generated in simulation by deploying a locomotion policy on randomly generated terrain with random directional commands.
Energy consumption of the legged robots was of low concern during the Final Event, due to operation time greater than run's duration. However, a single navigation failure could be mission ending; thus we use a weighted sum of risk and time cost, with a five times higher weight on risk.

Although the ANYmal navigation planner plans towards a single goal pose, embedded into our autonomy stack, it receives a reference path from the exploration planner. 
We therefore iteratively try to plan towards the farthest pose in the exploration path, until we either succeed, or no valid goal pose can be found.
Because we use a roadmap-based planner, these path queries can be performed rapidly once a planning graph has been built.

\subsubsection{ANYmal behavior tree}\label{sec:anymal_behavior_tree}
\begin{figure}[]
    \centering
    \includegraphics[keepaspectratio,height=3.5cm]{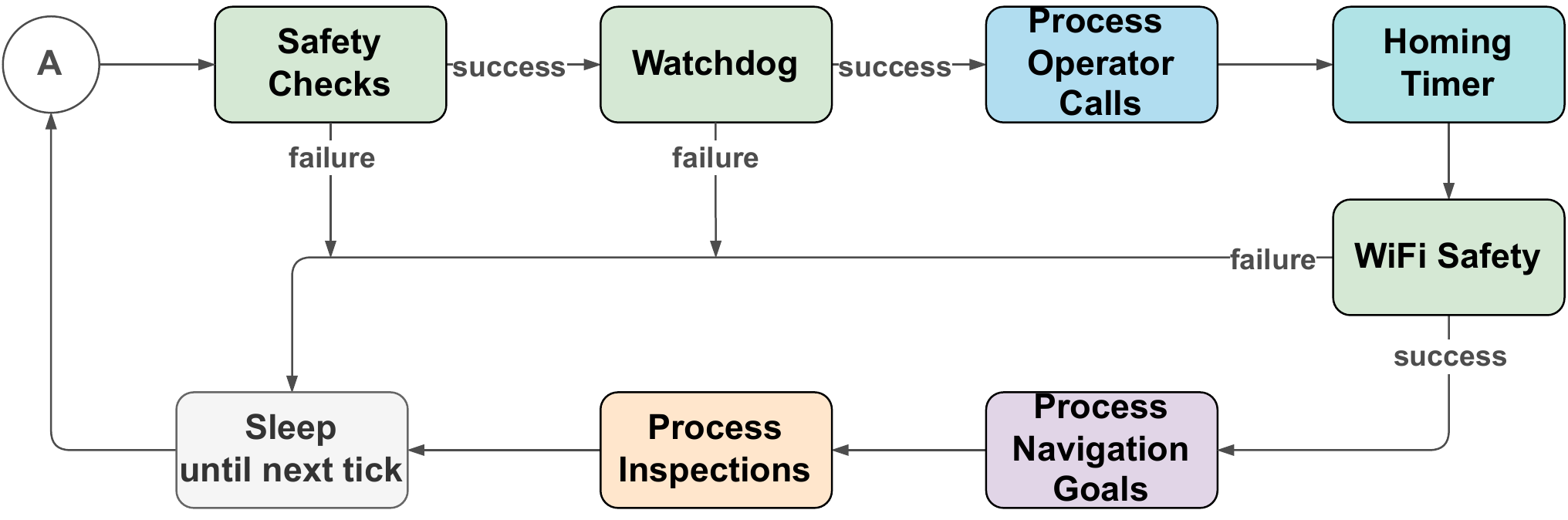}
    \caption{Overview of the ANYmal behavior tree module. Active blocks are displayed in green, blue, and purple. The grey block represents a sleeping block. Based on the given conditions, blocks can either return success and continue to the next block or fail. In case of failure or after reaching the last active block, the behavior tree sleeps until the subsequent execution.}
    \label{fig:anymal_behaviortree}
\end{figure}

\begin{figure}[]
    \centering
    \includegraphics[keepaspectratio,height=5.9cm]{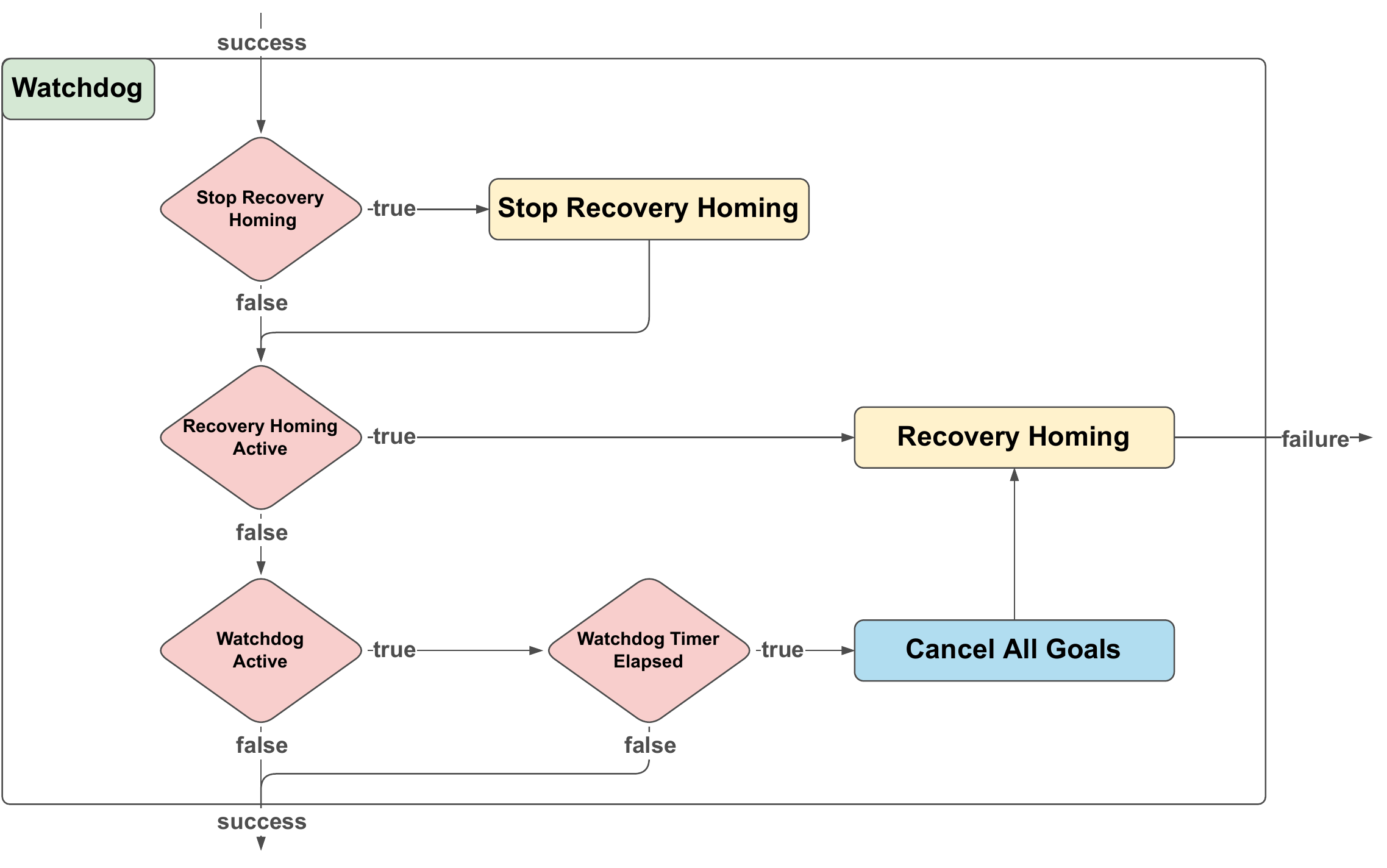}
    \caption{Watchdog module of the ANYmal behavior tree. This module triggers recovering actions if the agent is detected as not moving for a given amount of time.
    }
    \label{fig:watchdog}
\end{figure}

\begin{figure}[]
    \centering
    \includegraphics[keepaspectratio,height=5.9cm]{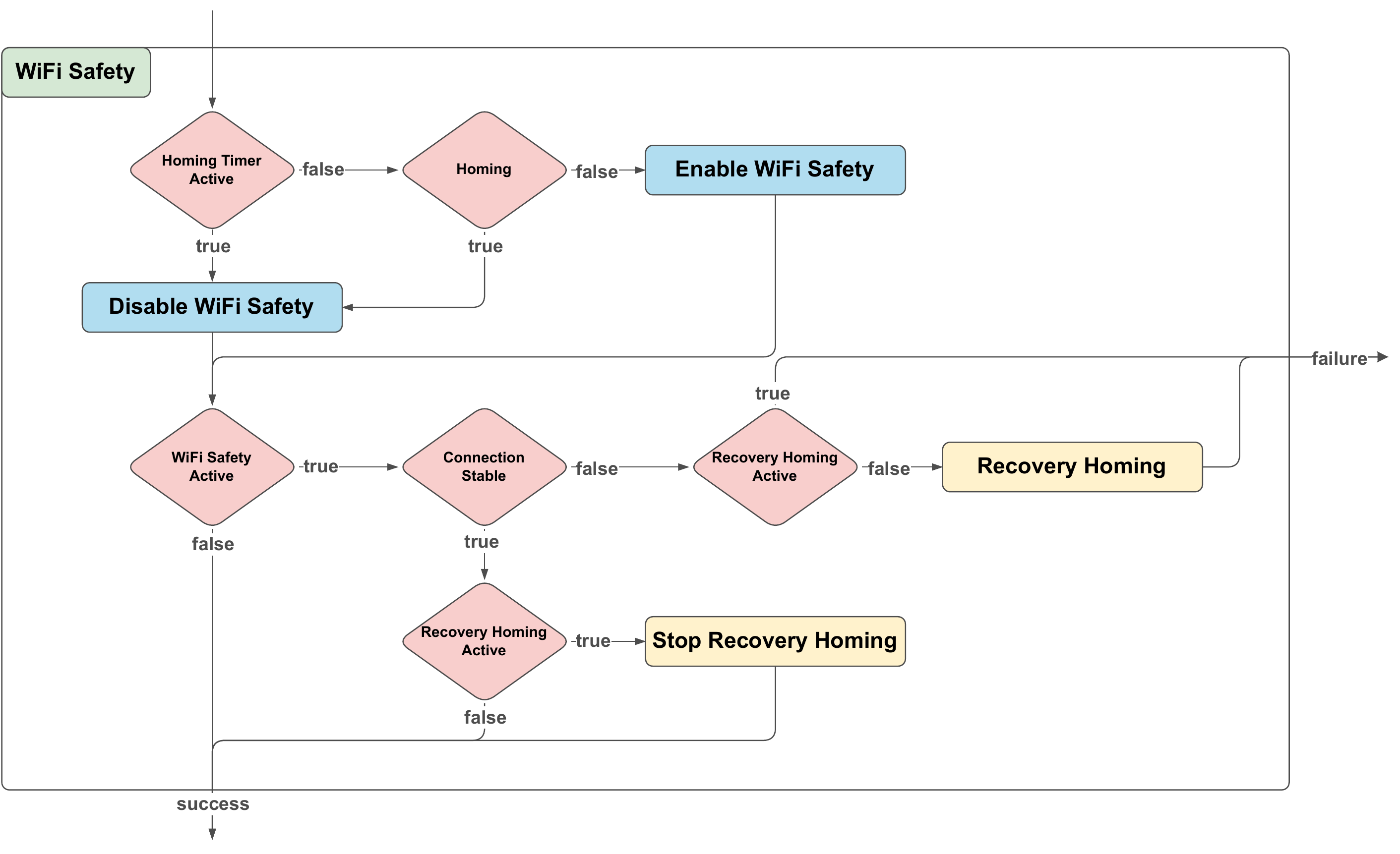}
    \caption{WiFi Safety module of the ANYmal behavior tree. This module checks if the agent is in connection to the Base Station, and if the agent is not assigned a time budget for exploration out of communication range, it commands the robot to backtrack its path until the connection is reestablished.
    }
    \label{fig:wifi_safety}
\end{figure}

The capabilities for coordinating different software modules to execute autonomous missions are provided by a behavior tree module running onboard the ANYmals. Moreover, the behavior tree receives commands from the Human Supervisor and executes recovery behaviors if it detects unexpected states during the mission execution. This module is based on an existing framework~\footnote{\url{https://github.com/BehaviorTree/BehaviorTree.CPP}} and its main components are depicted in Figure~\ref{fig:anymal_behaviortree}.
Specifically, the \emph{Watchdog} block (Figure~\ref{fig:watchdog}) triggers recovering actions if unexpected conditions are detected. It continuously checks the robot's position and triggers a backtracking of the traversed path, called ``Recovery Homing,'' in case the robot does not move for a predefined amount of time.
The \emph{Process Operator Calls} block handles the commands given by the supervisor, assigning them a higher priority compared to the other onboard modules' requests. %
If the Human Supervisor allocated a time budget for exploration out of communication range, the \emph{Homing Timer} block checks whether the given budget has elapsed and if so, it triggers the homing functionality of the exploration planner.
If the Human Supervisor did not allocate a time budget, the \emph{WiFi Safety} block (Figure~\ref{fig:wifi_safety}) checks if the agent is in connection with the Base Station. If the connection is lost (based on a threshold on consecutive, failed pings), it commands the robot to backtrack its path until the connection is reestablished.
The \emph{Process Navigation Goals} block handles navigation goals created either by the Human Supervisor or the exploration planner and implements additional mechanisms to manage possible pitfalls. It forwards the assigned goals to the navigation planner to compute a terrain-aware path or directly to the path tracking module if the Human Supervisor decides to bypass the navigation planner. Possible pitfalls handled by this block include (a) the navigation planner cannot find a valid path to any pose in the reference path, and (b) the navigation planner considers the robot's starting position invalid (for example, due to drift in the odometry estimate).
For the former case, depending on which module provided the request, either the supervisor gets informed about the inability to continue or an automatic replanning command to the exploration planner is issued. In the latter case, the module automatically resets the elevation map to remove any possible artifacts and issues again the last command that failed to execute. In either case, if the exploration planner provides a goal that cannot be achieved for a given amount of time, the robot is commanded to backtrack. While the robot backtracks its steps, the behavior tree triggers the exploration planner until a solution is found or the Human Supervisor stops the procedure. Lastly, the block \emph{Process Inspections} receives inspection requests, applicable only for the Explorer robots, that are sent either by the exploration planner or the Human Supervisor and puts the other requests on hold until the inspection task is completed.


%% file: 03_04_networking.tex
While the autonomy system of each robot is crucial to achieve subterranean exploration with minimal human supervision, simultaneously maintaining a reliable, decentralized wireless network is critical transmit the gathered information (such as the map, artifact locations, and robot poses) from the robots to the Base Station so as to provide situational awareness to the Human Supervisor. Wireless communication is challenging in underground settings due to the unpredictability of the signal propagation and the constrained environment geometry. Similar to the system described in our previous work, \cite{tranzatto2022cerberus}, the CERBERUS system-of-systems employed an ad-hoc wireless mesh network, composed of several physical ``nodes,'' spread of extended distances ensuring connection to the robots exploring deep in the unknown.

\begin{figure}[h]
    \centering
    \includegraphics[scale=0.25]{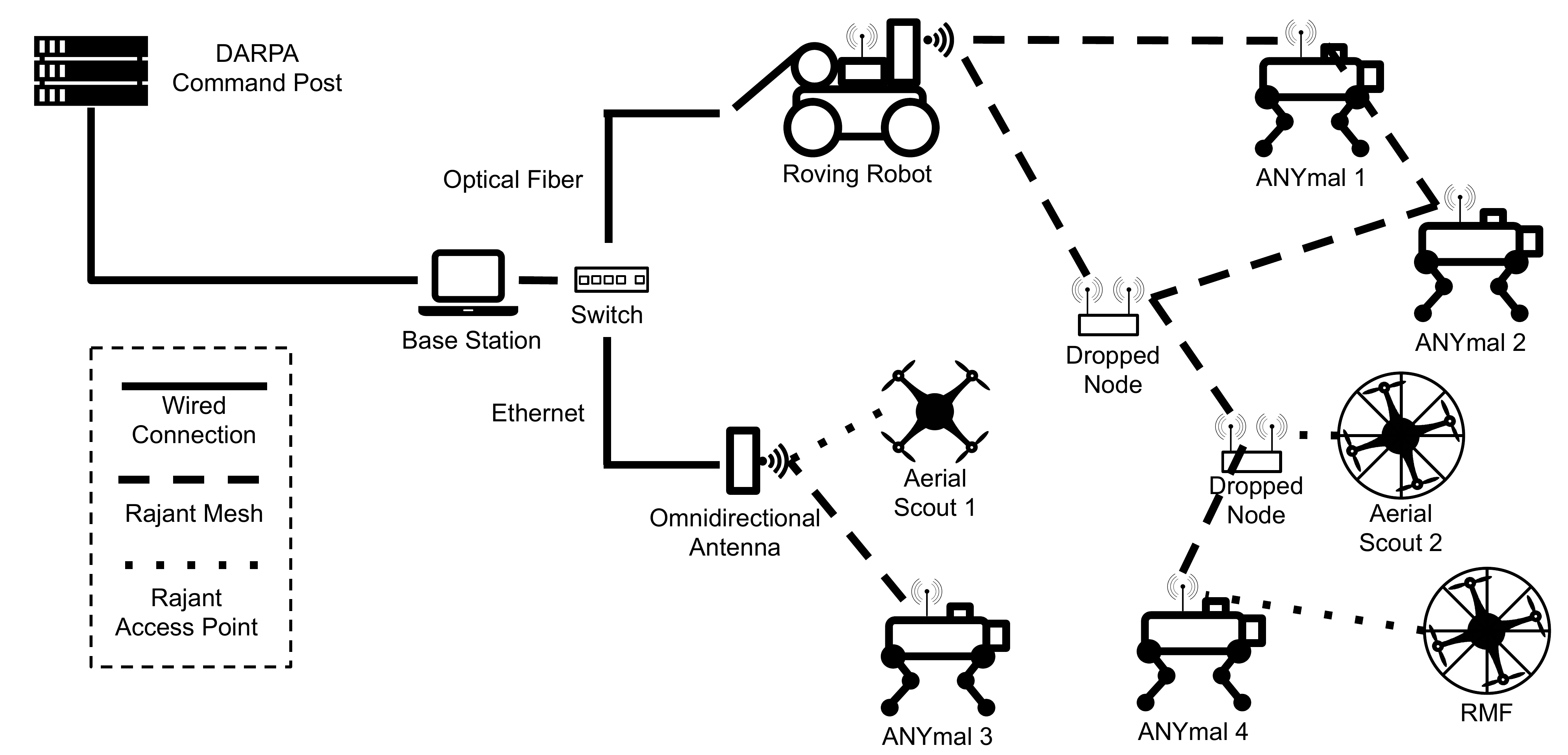}
    \caption{Block-diagram showing the networking infrastructure used during the Final Event. The Base Station, the quadrupedal robots, the roving robot, and the dropped communication-extender breadcrumb nodes created an ad-hoc mesh network, which established the backbone communication infrastructure. The roving robot was also equipped with a fiber optical cable to maintain a connection to the Base Station. Aerial scouts used a standard ($802.11$) WiFi network, acting as clients connecting to the backbone infrastructure.
    }
    \label{fig:network_diagram}
\end{figure}

The network operated in the \SI{5.8}{\giga\hertz} range and was composed of the Base Station, the legged and roving robots, which acted as nodes of the mesh, alongside the flying robots which acted as clients. Furthermore, Carrier ANYmals could ferry and deploy additional  communication-extender nodes, called ``breadcrumbs,'' to expand the reach of the wireless mesh network incrementally upon human command. The mesh network constituted the team's backbone communication infrastructure, which could also be accessed by the aerial scouts, acting as clients. Figure~\ref{fig:network_diagram} shows an overview of the communication architecture.
In the following part, we highlight the main differences of the system deployed in the SubT finals as compared to our intermediate publication.

\textbf{Mobile agent nodes}\\
The legged robots and the roving robot featured a commercial off-the-shelf radio (``Rajant BreadCrumb DX2'') that acted as a mesh point node and a WiFi access point. Each radio formed a mesh network when used in conjunction with other Rajant's BreadCrumb systems. In this mesh network, the DX2 radios connected directly, dynamically, and non-hierarchically to as many other DX2 radios as possible and cooperated to efficiently route data between clients. Moreover, each radio operated also as a WiFi access point, broadcasting a WiFi network with the same SSID. Aerial robots did not have a DX2 radio or their own - due to their limited payload capability and size constraints. Instead, they connected to the closest existing WiFi access point. In our previous setup for the Tunnel and Urban Circuit events the aerial robots could not connect directly to the legged robots; at the Final Event this limitation was removed.

\textbf{Base station nodes}\\
Our approach had two solutions for communications at the Base Station. The primary setup was a ``Rajat BreadCrumb ES1'' radio in the Staging Area, attached to an omnidirectional antenna to connect with the robots before entering the competition course. The ES1 radio can mesh with other DX2 radios. 
Second, the Base Station could be connected to a roving wheel robot via a \SI{300}{\meter}-long optical fiber cable that could unwind as the rover drove in the course while carrying a DX2 radio. This dual configuration allowed the robots in the Staging Area to connect to the ES1 radio when the roving robot was deployed deep in the course and out of wireless reach. Compared to our setup of the Tunnel and Urban events, the Rajant radios offered the possibility to connect both an ES1 and a DX2 to the Base Station via wired connection without creating any network loop, which was not possible with the previous technology.

\begin{figure}[h]
    \centering
    \begin{subfigure}{.4\textwidth}
      \centering
      \includegraphics[keepaspectratio,height=5.5cm]{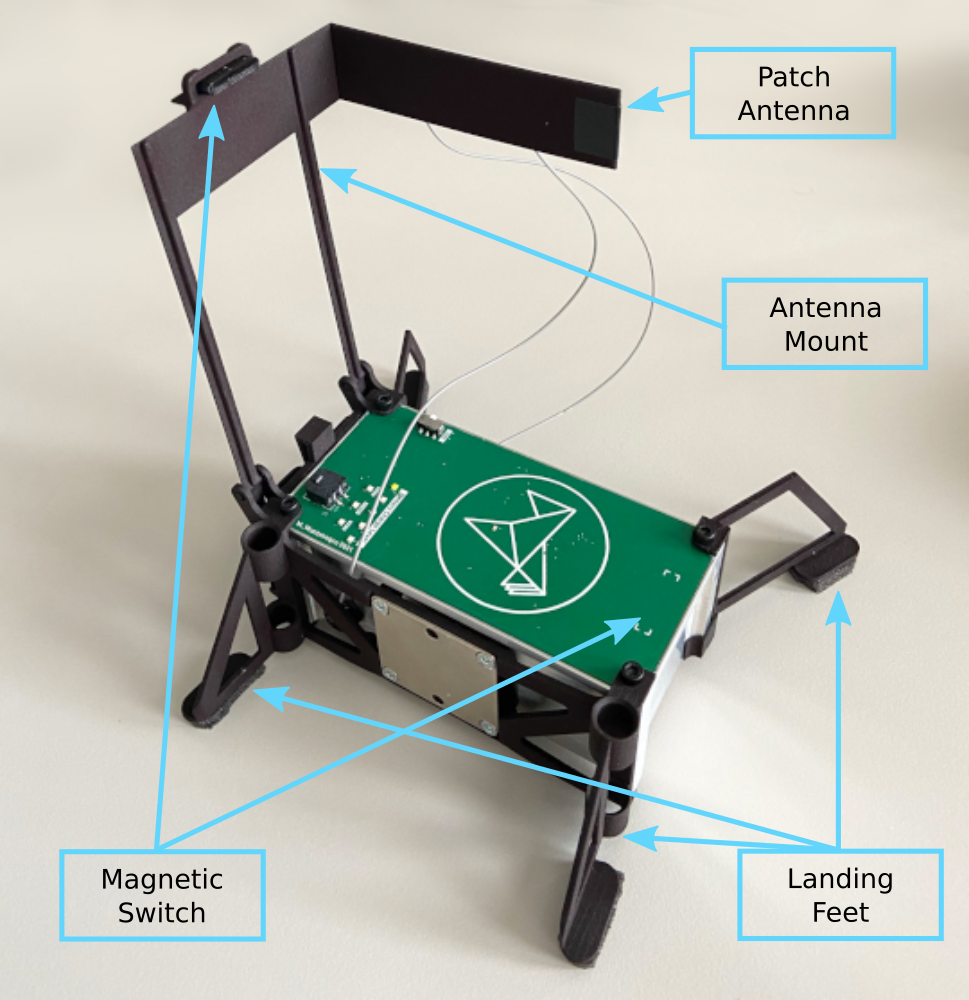}
      \caption{WiFi breadcrumb.}
      \label{fig:wifi_breadcrumb_detail}
    \end{subfigure}
    \begin{subfigure}{.4\textwidth}
      \centering
      \includegraphics[keepaspectratio,height=5.5cm]{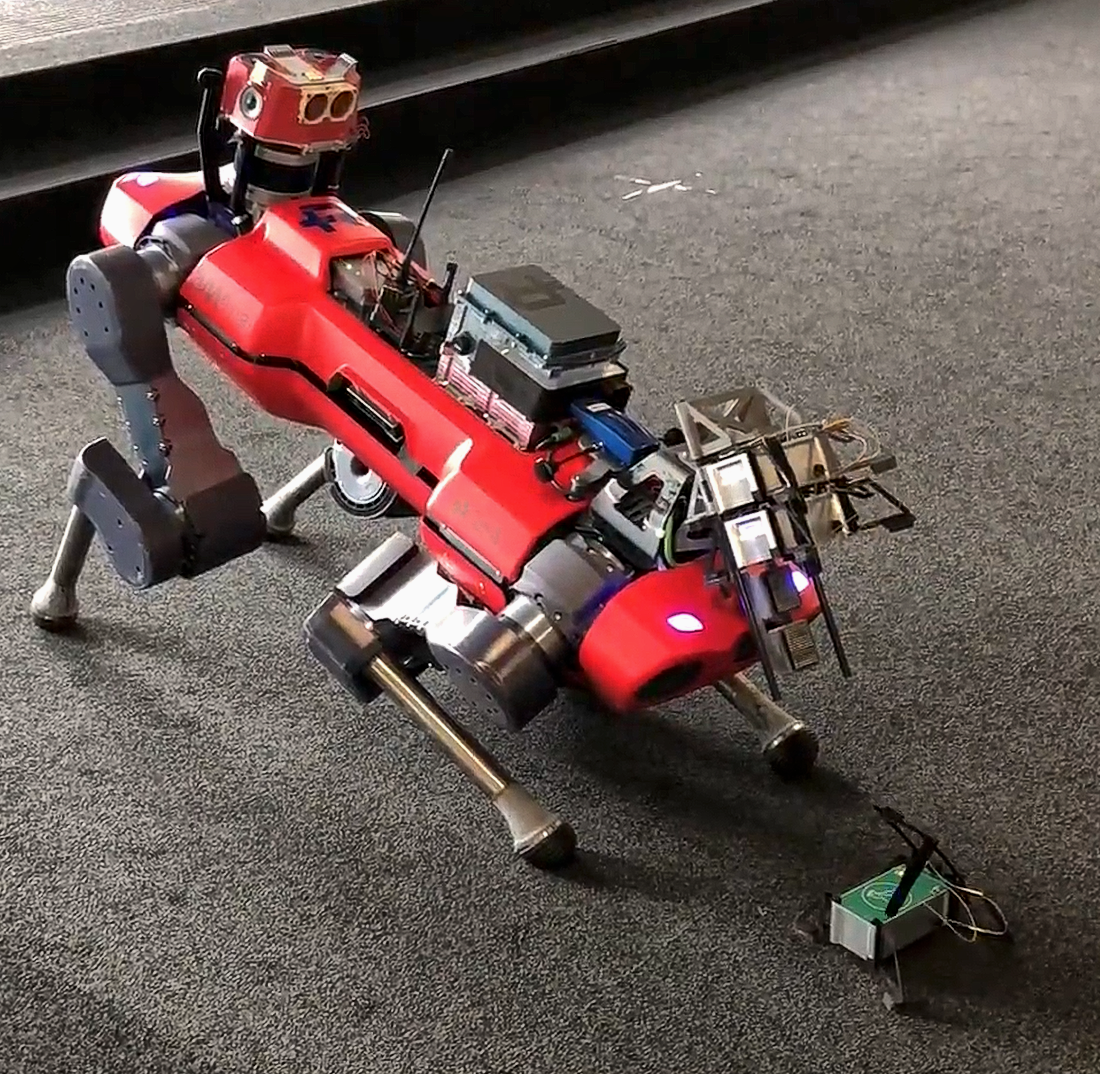}
      \caption{WiFi breadcrumb deployment.}
      \label{fig:wifi_breadcrumb_deployment}
    \end{subfigure}
    
    \caption{Communication-extender modules and ANYmal C SubT Carrier. (\subref{fig:wifi_breadcrumb_detail}) Assembled WiFi breadcrumb module. A WiFi patch antenna is attached to a compliant foldable mechanism. When loaded on the robot, the antenna is folded down and under tension. When being deployed, the compliant structure pushes the module out of its magazine. On the ground the antenna unfolded into position for better reception. (\subref{fig:wifi_breadcrumb_deployment}) Carrier ANYmal tilting its torso and deploying a WiFi breadcrumb on the ground.}
    \label{fig:wifi_breadcrumb}
\end{figure}

\textbf{Deployable breadcrumb nodes}\label{sec:deployable_breadcrumb_nodes}\\
Carrier ANYmal robots ferried additional WiFi breadcrumbs around the course and deployed them to incrementally expand the reach of the wireless mesh network. Carriers could transport four wireless breadcrumb modules and deployed them by lowering their torso and using an electromagnetic release mechanism (see Figure~\ref{fig:wifi_breadcrumb_deployment}), as described in Section ~\ref{sec:anymal_c_subt}. Each breadcrumb was a water and dustproof self-contained unit, consisting of a \SI{2}{\hour} battery pack, a battery management system, a magnetic switch (to switch the module on after releasing it), an LED to aid the recovery of the module after a mission, a WiFi patch antenna, and a ``BreadCrumb DX2'' radio (see Figure~\ref{fig:wifi_breadcrumb_detail}). The patch antenna was installed on a compliant, \num{3}D printed part that could be folded to allow the stacking of multiple beacons when loaded in the release mechanism.

Once the electromagnet holding a beacon was deactivated, the antenna mount would release tension, helping the module to slide down a carbon fiber tube and open in the upright position. The modules had four landing feet, with padding material on the bottom to absorb impacts when falling to the ground after release. Each beacon weighed \SI{245}{\gram} and was \SI{120}{\milli\meter} x \SI{60}{\milli\meter} x \SI{50}{\milli\meter} in size with the closed antenna mount. When standing on the ground, the antenna extended \SI{17}{\centi\meter} upwards.
We did not automate the deployment of these beacons. During the Final Event, the Human Supervisor made the expert decision about where and when to deploy the breadcrumb nodes as wireless signal propagation is often unpredictable in subterranean scenarios.

\textbf{Nimbro network for multi-master ROS networking}\label{sec:nimbro_network}\\
Above the physical layer, we employed the \ac{ROS} (version \num{1}) as the middleware for message passing between processes running on each robot, as well as across robots on the same network. Each robot ran their own onboard \textit{rosmaster} process, which monitored all internal communication connections. Given our use case scenario with multiple robots and the Base Station, each with their \textit{rosmaster}, we used the Nimbro Network\footnote{\url{https://github.com/AIS-Bonn/nimbro_network}} software package to allow multiple such instances on the same network.

This software package was configured to select individual data streams which were transmitted between the agents and the Base Station. We additionally implemented an option to distinguish between essential and non-essential topics. Essential topics included the robot's heartbeat and telemetry, artifact reports, and mapping data. Non-essential topics included camera streams and other data not strictly needed to be sent to the Base Station for the mission execution. All robots were allowed to stream essential topics to the Base Station continuously and could stream the non-essential ones only if selected by the Human Supervisor. This option, developed for the finals, prevented unselected robots from sending not-essential and bandwidth-hungry data when not specifically desired.

One possibility we considered was to use ROS\num{2} and its native support for multi robots systems and \ac{QoS} to manage the bandwidth requirements between robots. We decided against mixing ROS\num{1} and ROS\num{2} for simplicity.

%% file: 03_05_artifacts.tex
Each robot performed artifact detection onboard using a trained neural network similar to~\cite{yolo_v3_redmon}. The trained object detector provided bounding boxes identifying the correct class of the artifact from image streams collected by the cameras onboard the robots. Because of their small size or nature, the Phone, SubT Cube, and CO$_2$ gas artifacts were detected using dedicated sensors that measured the Bluetooth strength and CO$_2$ gas concentration. Multiple detections of the same artifact were fused to refine its estimated position. When sufficient confidence was reached, a single report was sent to the Base Station over the communication network.

\textbf{Training}\label{sec:artifact_training}\\
To train the object detector, we collected images using various camera sensors in urban basements, abandoned tunnels, mines, and natural caves, with lighting conditions ranging from low ambient lighting to complete darkness, depending only on the robot's onboard light sources. Datasets were acquired in the presence of obscurants and with a wide variety of background colors and textures. Different types of cameras and lenses were used on the robots when collecting data which contributed to a diverse common training dataset. We observed that training the detector with images obtained using different sensors at different resolutions improved the detection accuracy.

\begin{figure}[]
    \centering
    \includegraphics[width=\textwidth]{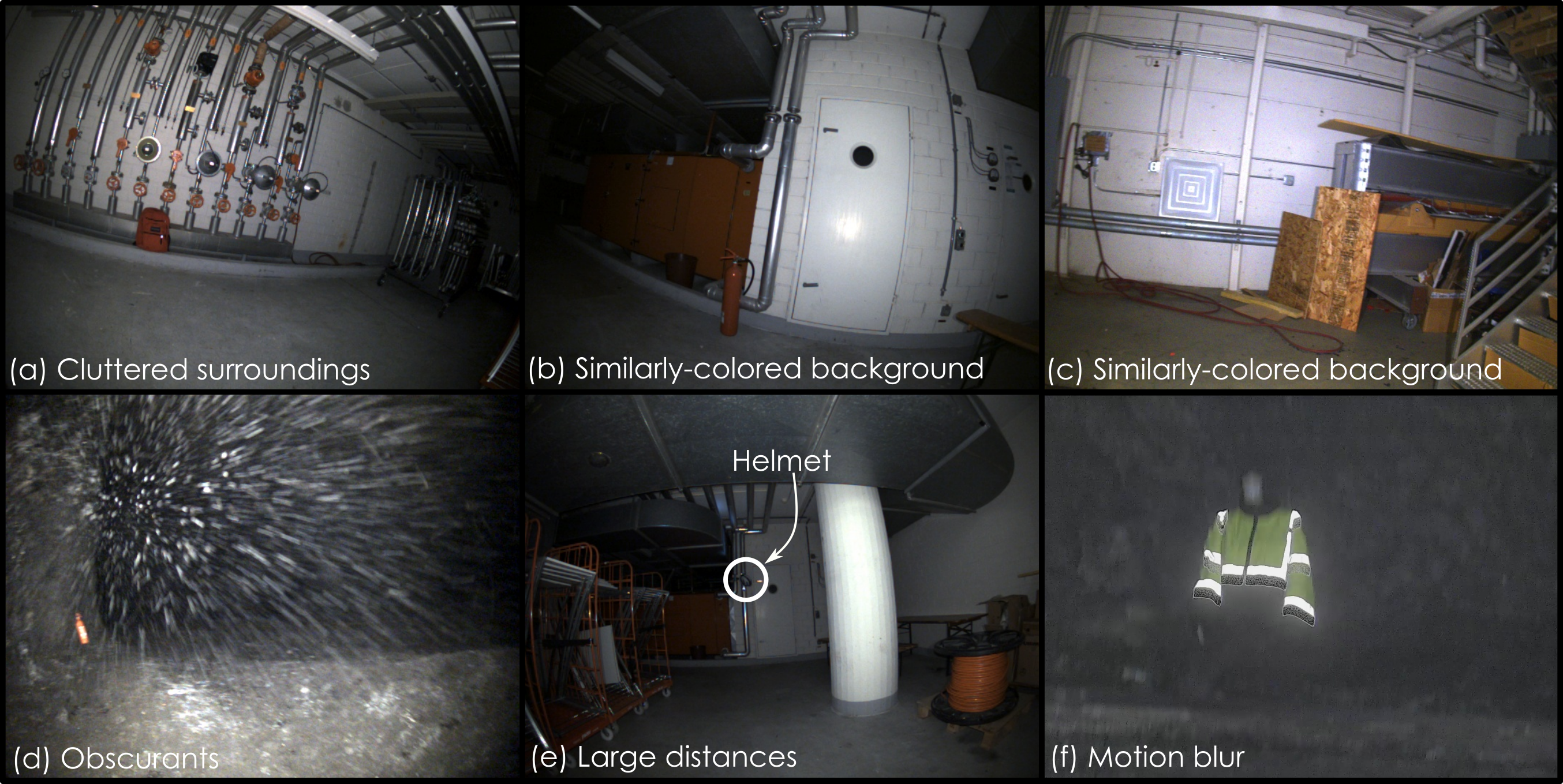}
    \caption{Specific examples where the detector regularly failed were collected to improve performance. Images were collected in cluttered surroundings (a-c) with distracting colors in the background or clutter similar to the artifacts to improve detection accuracy. Images were also collected in the presence of obscurants (d), with artifacts at large distances (e), and with the camera in motion to induce blur (f), as would be observed by the robot in real-life deployments.}
    \label{fig:artifact_training_variety}
\end{figure}

During the training process, special care was taken to identify conditions where the detector failed and to collect images for the detector to perform better in similar scenarios. Figure ~\ref{fig:artifact_training_variety} shows some of these cases. Examples included the detection of artifacts against similarly-colored backgrounds, presence of motion-blur, obscurants, and artifacts located distant from the robot and against cluttered surroundings. These cases were identified through continuous field testing and evaluation on datasets. The collected images were manually filtered to remove repetitive images to prevent over-fitting. The images that had significant amounts of obscurants or extreme motion blur were identified and removed. In total, \num{40007} images containing artifacts were labeled manually. Table \ref{tab:artifact_label_count} shows the number of labeled instances per artifact. Despite using labeled images for the ``Cell Phone'' in the training data, we used Bluetooth as the primary method to detect the artifact. In addition to these, we collected a set of \num{384} images containing no artifacts, but instead containing challenging environments and specific conditions that were found to cause false-positive detections. The training set consisted of \SI{90}{\percent} of all the images while the remaining \SI{10}{\percent} were used for validation. The neural network was trained on this set for up to \num{16000} epochs, and the best performing parameters for training were chosen based on their results on the validation dataset.

\begin{table}
\centering
\begin{tabular}{c|c}
\toprule
\multicolumn{2}{c}{\textbf{Training Images for the Artifact Detector}} \\
\midrule
\textbf{Artifact Class} & \textbf{Number of Labels} \\
\midrule
 Survivor & \num{6477} \\
 Fire Extinguisher & \num{5357} \\
 Drill & \num{4580} \\
 Backpack & \num{3415} \\
 Vent & \num{5350} \\
 Helmet & \num{6053} \\
 Rope & \num{5219} \\
 Cell Phone & \num{3556} \\
 Total & \num{40007} \\
\bottomrule
\end{tabular}
\caption{Number of labeled images used to train a neural network using YOLOv3 for each class of visible artifact.}
\label{tab:artifact_label_count}
\end{table}

\textbf{Inference}\label{sec:artifact_inference}\\
We adapted the trained neural networks to run on dedicated hardware on each robot. For the DJI Matrice 100-based aerial scouts and Kolibri platforms we used the Intel Movidius Neural Compute Stick 2 to perform inference at \SI{5}{\hertz} per camera stream. For the RMF-Owl, the weights were converted from 32-bit floating-point numbers to 8-bit integers to run on the NPU of the Khadas VIM3 computer at a rate of \SI{3}{\hertz}. ANYmal C used a Jetson AGX Xavier and NVIDIA's TensorRT SDK. This module used 16-bit floating-point numbers and batch processing, achieving an inference rate of \SI{3}{\hertz} for each of the four color cameras, while running alongside the elevation mapping pipeline on the Jetson. For the roving platform, the dedicated onboard GPU comprising of a GeForce \num{1070} graphics card from NVIDIA ran four instances of the object detection module, one for each of the onboard color cameras, and achieved an inference rate of \SI{5}{\hertz} per camera stream.

\begin{figure}[h]
    \centering
    \includegraphics[width=\textwidth]{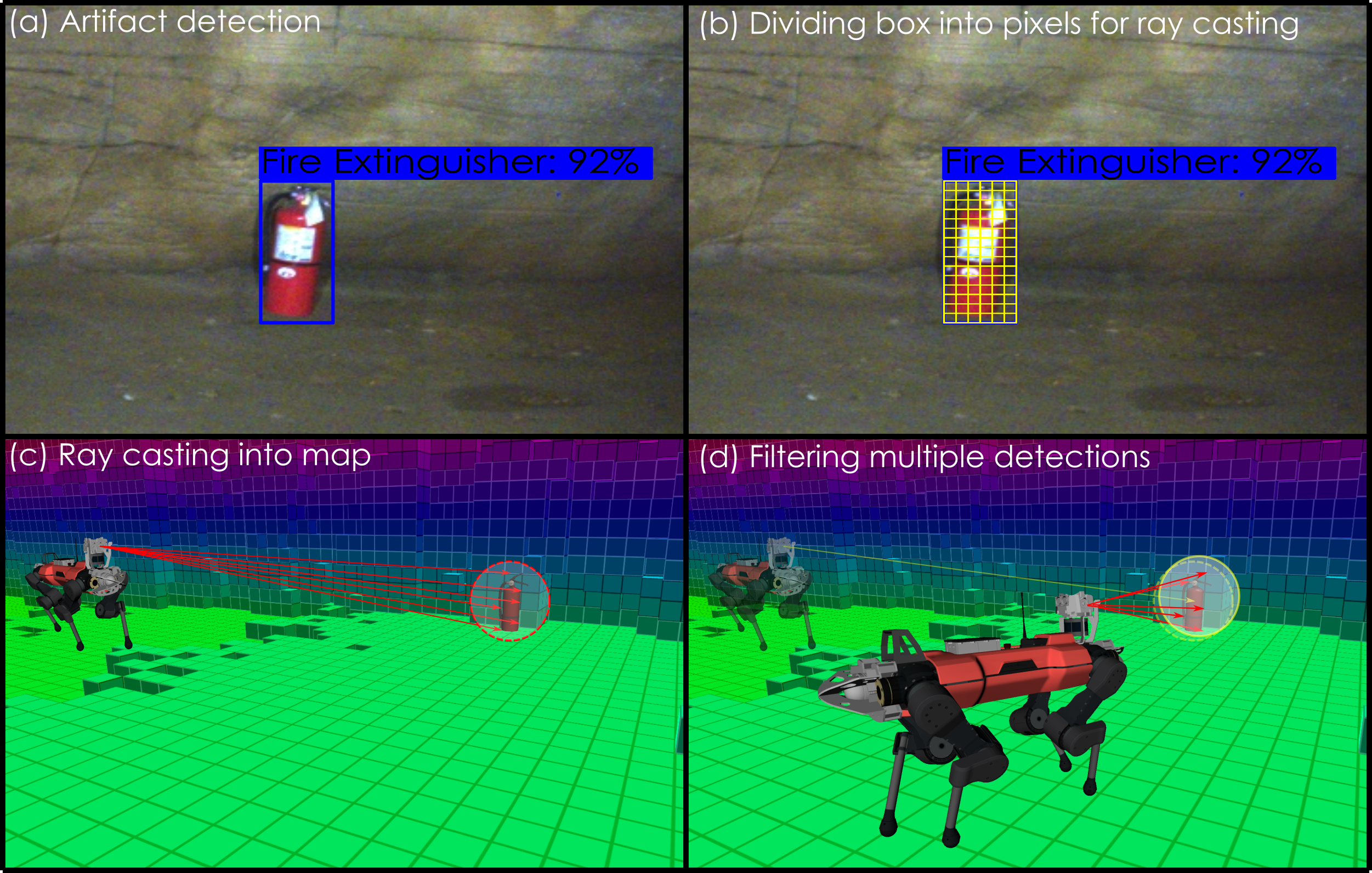}
    \caption{Artifact detection methodology employed by Team CERBERUS for artifacts detected by YOLOv3. (a) The artifact is detected with the bounding boxes around it and (b) the area inside the bounding box is divided into pixels. (c) For each of these pixels, rays are cast into the volumetric map of the robot to locate the artifact. (d) Multiple detection from the same robot are filtered to refine the position of the artifact.}
    \label{fig:artifact_detection}
\end{figure}

\textbf{Detection filtering}\label{sec:artifact_filtering}\\
For the operator, the object detector overlaid rectangular bounding boxes around the object with the probability of a correct detection as shown in Figure~\ref{fig:artifact_detection}. Because detections could be noisy or inconsistent detections, we filtered the artifact position estimate. Individual rays were cast into the volumetric map built by the exploration planner for each pixel inside the bounding box using the camera parameters. To estimate the position of the object, the median point of intersection of these rays with the map was calculated. A virtual sphere was constructed, centered at this median point. Further measurements from subsequent images that estimated the artifact's location within this sphere were considered to be detections of the same object. The sphere center's position was updated by averaging all the accepted detections. With each new detection, a separate binary Bayes filter recursively updated the probabilities for each artifact class. Once the probability for an artifact class exceeded a predefined threshold, the process was stopped, and the updated center of this sphere was reported to the Base Station as the position of the artifact. Subsequent detections of the artifact on the same robot were ignored to prevent duplicate reports. If the robot was out of communication range of the Base Station, the artifact's class and location was stored on the robot and was transferred to the Base Station once the communication was re-established.

\begin{figure}[]
    \begin{subfigure}{.65\textwidth}
      \centering
      \includegraphics[keepaspectratio,height=4.5cm]{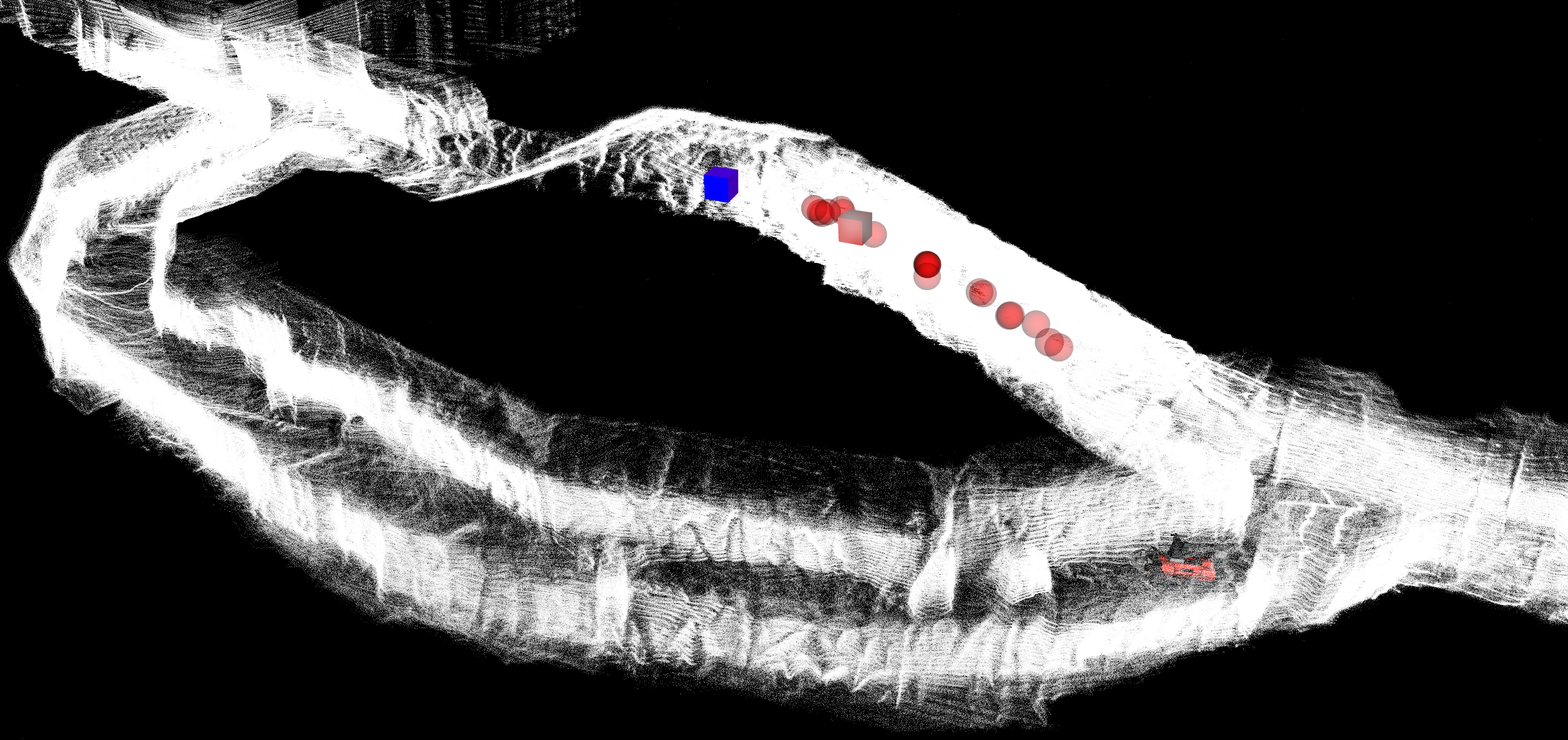}
      \caption{Clustering of measurements}
      \label{fig:clustering_bluetooth_measurements}
    \end{subfigure}
    \begin{subfigure}{.34\textwidth}
      \centering
      \includegraphics[keepaspectratio,height=4.5cm]{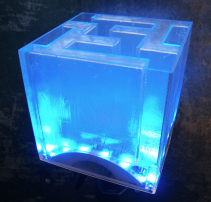}
      \caption{The SubT cube artifact}
      \label{fig:subt_cube_visual}
    \end{subfigure}
    \caption{Cube artifact detection. (\subref{fig:clustering_bluetooth_measurements}) The raw Bluetooth measurements are shown as red spheres, located on the positions where the robots sampled them. A brighter color indicates a higher signal strength at the marked position. The red cube shows the cluster position, whereas the blue cube displays the artifact's ground truth position. Because the robot didn't actually pass by the cube, its position estimate was not precisely accurate, nonetheless our estimate was still inside the \SI{5}{m} scoring radius resulting in a successful artifact report. (\subref{fig:subt_cube_visual}) The SubT cube artifact.}
    \label{fig:artifact_cube}
\end{figure}

\textbf{Bluetooth and CO$_2$ detection}\label{sec:artifact_bluetooth}\\
Bluetooth and $\textrm{CO}_2$ readings were processed by the same software module by normalizing the raw sensor measurements to a predefined range. Bluetooth measurements were clustered by their SSID identifier, whereas $\textrm{CO}_2$ measurements were clustered by distance. When a new raw measurement above a certain threshold was received, a new measure cluster was created. Each cluster's center was continuously updated using the largest measured values. A cluster's center was reported to the Human Supervisor when enough measurements about a predefined threshold were received. Both the SubT Cube and the Phone artifact were detected by Bluetooth localization; visual detections generally unreliable in the latter case. An example of clustered measurements used to obtain an estimate of a SubT Cube artifact position is shown in Figure~\ref{fig:artifact_cube}.

\textbf{Position optimization}\label{sec:artifact_Optimizations}\\
M3RM (Section \ref{sec:m3rm}) continuously estimated the optimized position of detected artifacts. Additionally to estimate the global artifact position estimated by CompSLAM (Section \ref{sec:compslam}) onboard the robot, each artifact report contained the position of the artifact in the occupancy map, expressed in the robot frame at the time of detection, as well as the respective timestamp.
For each artifact, M3RM then estimated the position of the reporting robot at the detection timestamp. The global optimized position for each artifact was calculated by transforming the position of the artifact with respect to the robot frame into the global frame using the optimized pose of the robot.
The position estimate for each artifact was updated after each full optimization cycle of M3RM.

\textbf{Inspection payload}\label{sec:inspection_payload}\\
On the ANYmal Explorer robots we employed the payload developed by ANYbotics' for inspection. This featured a \si{10}{x} optical zoom camera, a thermal camera, a directional microphone, and a spotlight. The unit can pan and tilt its sensor suite to provide a range of \SI{270}{\degree} for the pan joint and \SI{180}{\degree} for the tilt joint. The purpose of the moving sensor unit was to complement the static cameras of the Alphasense Core, which have a large field of view but cannot scan the distant surroundings and did not have thermal imaging. In contrast, the inspection payload's zoom camera had a smaller field of view but could spot far-off artifacts. Its thermal camera captured the thermal signatures of the survivor and the vent artifact.

\emph{Object detector}\label{sec:inspection_payload_object_detector} - We used the same image detector for the zoom camera images as for the Alphasense Core images, without any modifications. We implemented a dedicated detector for artifacts with thermal signatures (Survivor and Vent). For each received thermal image, the detector clustered pixels that were in the same temperature range as specified by DARPA for the Survivor artifact or \SI{10}{\degreeCelsius} above the ambient temperature (defined as the average of all temperature values of the last three images). If enough adjacent pixels were in a similar temperature range, a thermal artifact was considered detected and sent to the Human Supervisor for further inspection.

\emph{Scanning operation}\label{sec:inspection_payload_scanning_operation} - For any requested point to scan, the inspection payload's controller adjusted the yaw and pitch angles so that the inspection module's sensor suite pointed to the desired location. It changed the zoom level of the camera based on the distance to the area containing the point to scan so that if any artifact were present in the scanned area, it would have similar relative dimensions to the camera's resolution. Adjusting the zoom level based on distance was possible by performing ray-casting in the global volumetric map created by the exploration planner. The inspection payload featured four different operation modes: ``Operator Point Scan'', ``Operator Full Scan'', ``Autonomous Point Scan'', and ``Autonomous Full Scan''.
The Human Supervisor could trigger either ``Operator Point Scan'', to request the inspection of a given position or ``Operator Full Scan'' to request the scanning of the whole area (composed of multiple points scanned in sequence) surrounding the robot. The other two modes, ``Autonomous Point Scan'' and ``Autonomous Full Scan'', had the same functionalities as their ``Operator'' versions but were triggered by an autonomous scanning algorithm.

\begin{figure}[]
    \centering
    \begin{subfigure}{.99\linewidth}
      \centering
      \includegraphics[scale=0.2]{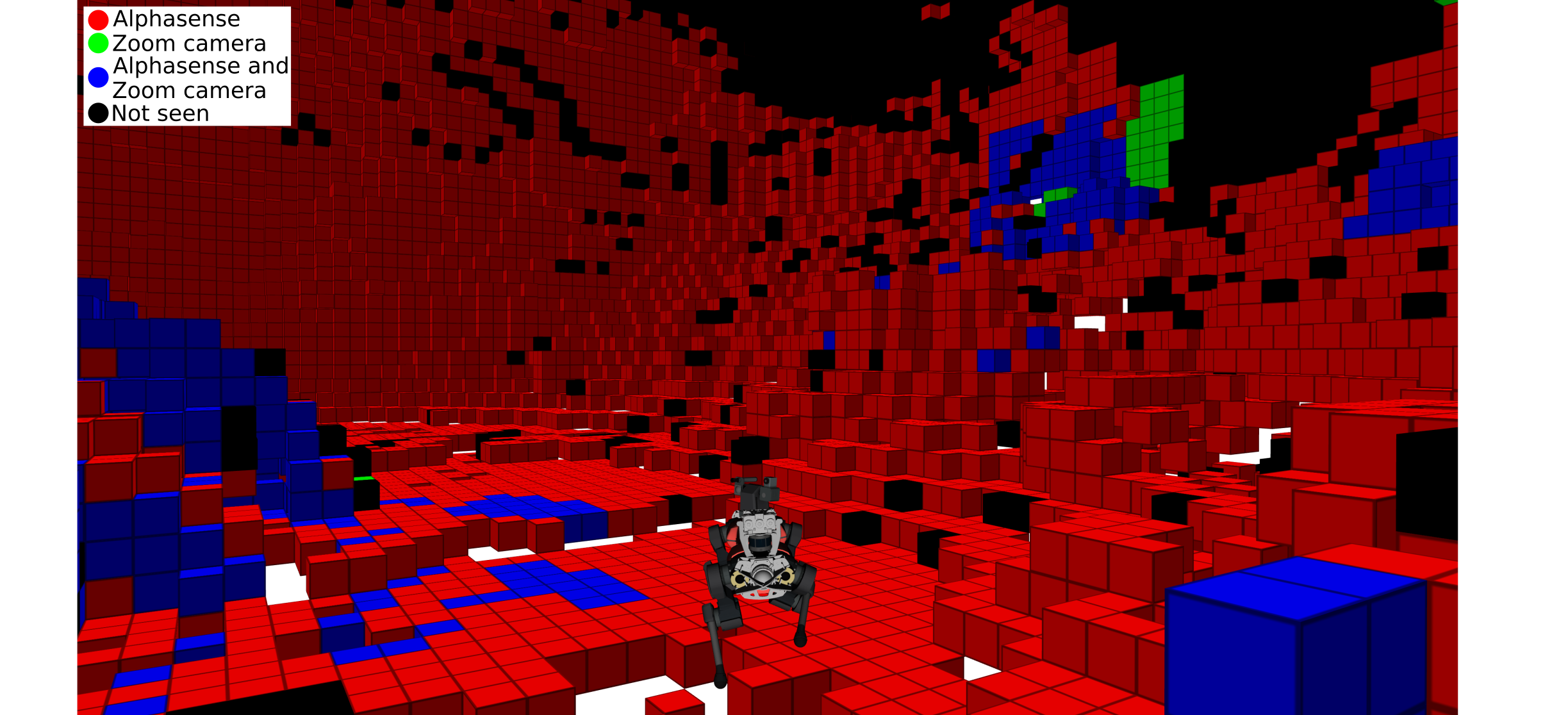}
      \caption{Volumetric map classified based on seen voxels.}
      \label{fig:inspection_payload_rviz}
    \end{subfigure}
    
    \begin{subfigure}{.99\linewidth}
      \centering
      \includegraphics[scale=0.2]{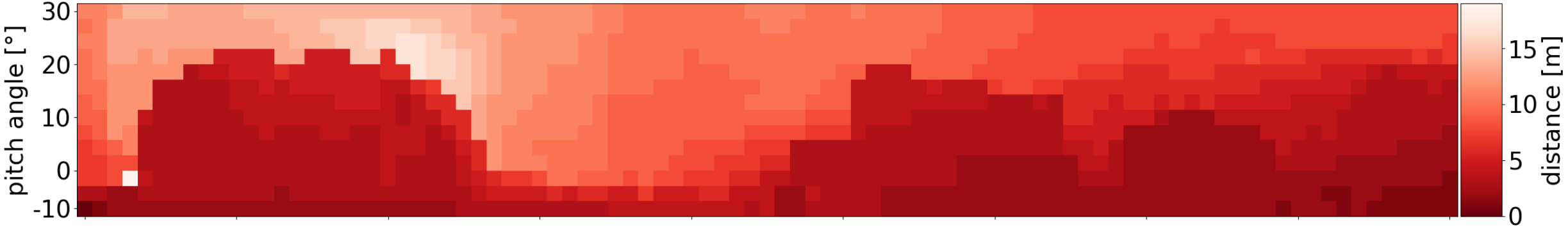}
      \caption{Robot's distance to voxels.}
      \label{fig:inspection_payload_distance}
    \end{subfigure}
    
    \begin{subfigure}{.99\linewidth}
      \centering
      \includegraphics[scale=0.2]{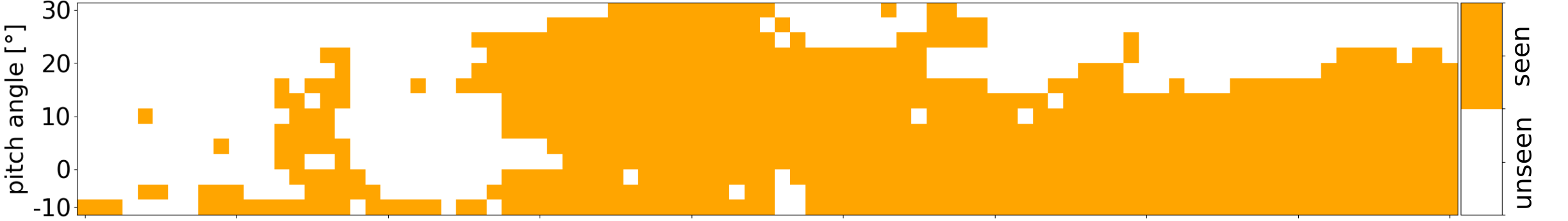}
      \caption{Already scanned area}
      \label{fig:inspection_payload_seen}
    \end{subfigure}
    
    \begin{subfigure}{.99\linewidth}
      \centering
      \includegraphics[scale=0.2]{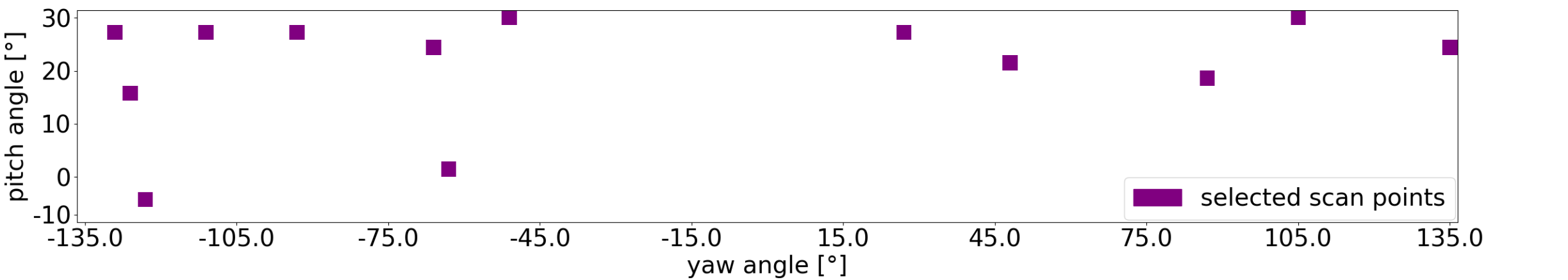}
      \caption{Scan points with seen area}
      \label{fig:inspection_payload_scan_points}
    \end{subfigure}
    
    \caption{Selection of scan points for the ANYmal C SubT Explorer's inspection payload. (\subref{fig:inspection_payload_rviz}) Visualization of the volumetric map created by the exploration planner: red voxels were seen by the Alphasense Core's color cameras, green voxels were already scanned by the zoom camera, and the blue voxels were inspected by both Alphasense Core and inspection payload, black voxels are considered as unseen. Voxels in $3$D are converted to a $2$D image. (\subref{fig:inspection_payload_distance})-(\subref{fig:inspection_payload_scan_points}) The $x$ and $y$ axes represent the inspection payload's workspace. (\subref{fig:inspection_payload_distance}) Distance from the zoom camera to the nearest voxel for each potential scan position. (\subref{fig:inspection_payload_seen}) Already inspected areas by the Alphasense Core and the zoom camera are marked in orange and unseen areas in white. (\subref{fig:inspection_payload_scan_points}) Final selected scan positions, displayed in purple,  are obtained considering the already scanned area and the distance to the nearest voxel.
    }
    \label{fig:inspection_payload}
\end{figure}

\emph{Autonomous scanning}\label{sec:inspection_payload_autonomous_scanning} - The autonomous scanning algorithm relied on the volumetric map built by the exploration planner to compute points to scan with the inspection payload.
Areas in the map closer than \SI{5}{m} and within the field of view of one Alphasense Core's color cameras were marked as seen. Afterwards, the algorithm computed unseen areas with respect to the inspection payload's joint space, which was set to \SI{-135}{\degree} to \SI{135}{\degree} for the yaw joint and \SI{-10}{\degree} to \SI{30}{\degree} for the pitch joint, with a resolution of \SI{3}{\degree} for both joints, resulting in a total number of \num{1170} potential scan points.
For each possible scan point, we performed ray-casting to determine where each ray intersected the global map. We marked each intersection with an identifier indicating whether it was unseen, and in that case we calculated the zoom camera's distance to that point. Both the zoom camera's distance to the nearest voxel and the already inspected area were considered to select the least number of scan points while still covering the entire workspace.  An example of the selection of new points to scan is depicted in Figure~\ref{fig:inspection_payload}, where it can be seen that no points are selected on already seen parts, and fewer scans are performed when the voxels are closer to the zoom camera. If the ``Autonomous Full Scan'' mode was enabled by the Human Supervisor, the scanning algorithm checked if the unseen area exceeded \SI{50}{\percent} of the reachable workspace and, in case, sent a request to the ANYmal's behavior tree to execute a full scan of the area. The behavior tree module then paused the currently running action, stopped the robot, scanned all points provided by the autonomous scanning algorithm, and resumed the previously running action. If the unseen area was less than \SI{50}{\percent} of the reachable workspace, the ``Autonomous Point Scan'' was executed, where the selected points were scanned while the robot was walking.

%% file: 03_06_single_human_supervisor.tex
A characteristic aspect for the SubT Challenge was the requirement to have a single operator, called the ``Human Supervisor'', as the only person permitted to use wireless (or other) communications to interact, coordinate, and manage the deployed robots during the competition run once they had entered the course. The single operator requirement was intended to minimize the amount of human intervention and pushed the competing teams to develop a high level of autonomy for remote mapping and navigation. The Human Supervisor was allowed to operate at a Base Station external but close to the course's entrance. The Base Station was a computer that served as an interface between the supervisor and the robots. Moreover, this computer directly connected to the DARPA Command Post. Each team had to send regular map updates and artifact reports to the Command Post. Artifact reports were automatically evaluated and scored if the reported positions were within five meters of the objects' ground truth.

\begin{figure}[]
    \centering
    \begin{subfigure}{.48\textwidth}
      \centering
      \includegraphics[scale=0.24]{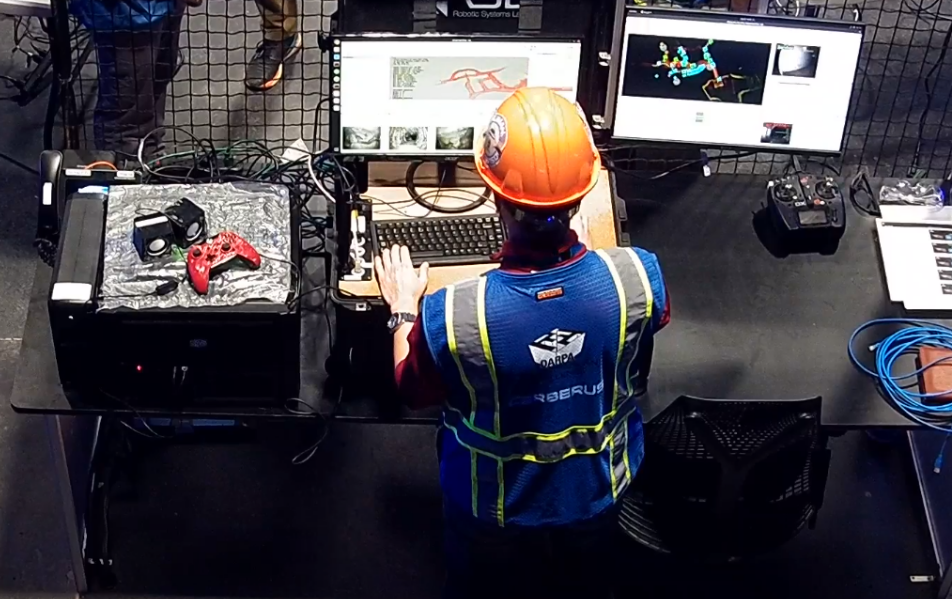}
      \caption{Base Station}
      \label{fig:base_station_setup}
    \end{subfigure}%
    \begin{subfigure}{.48\textwidth}
      \centering
      \includegraphics[scale=0.11]{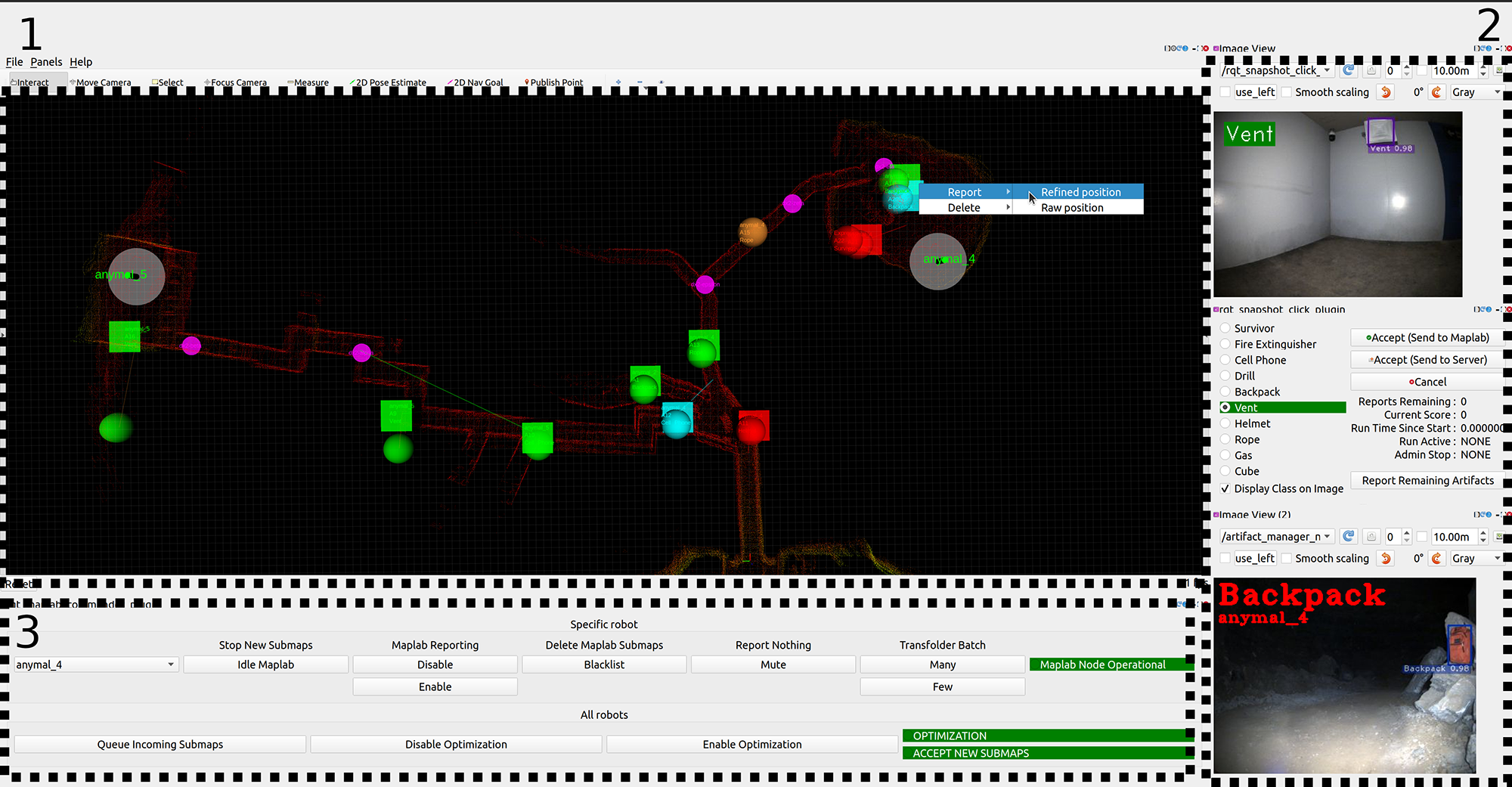}
      \caption{Monitor and Artifacts Reporting Interface (MARI)}
      \label{fig:base_station_mari}
    \end{subfigure}
    
    \bigskip
    \begin{subfigure}{.48\textwidth}
      \centering
      \includegraphics[scale=0.23]{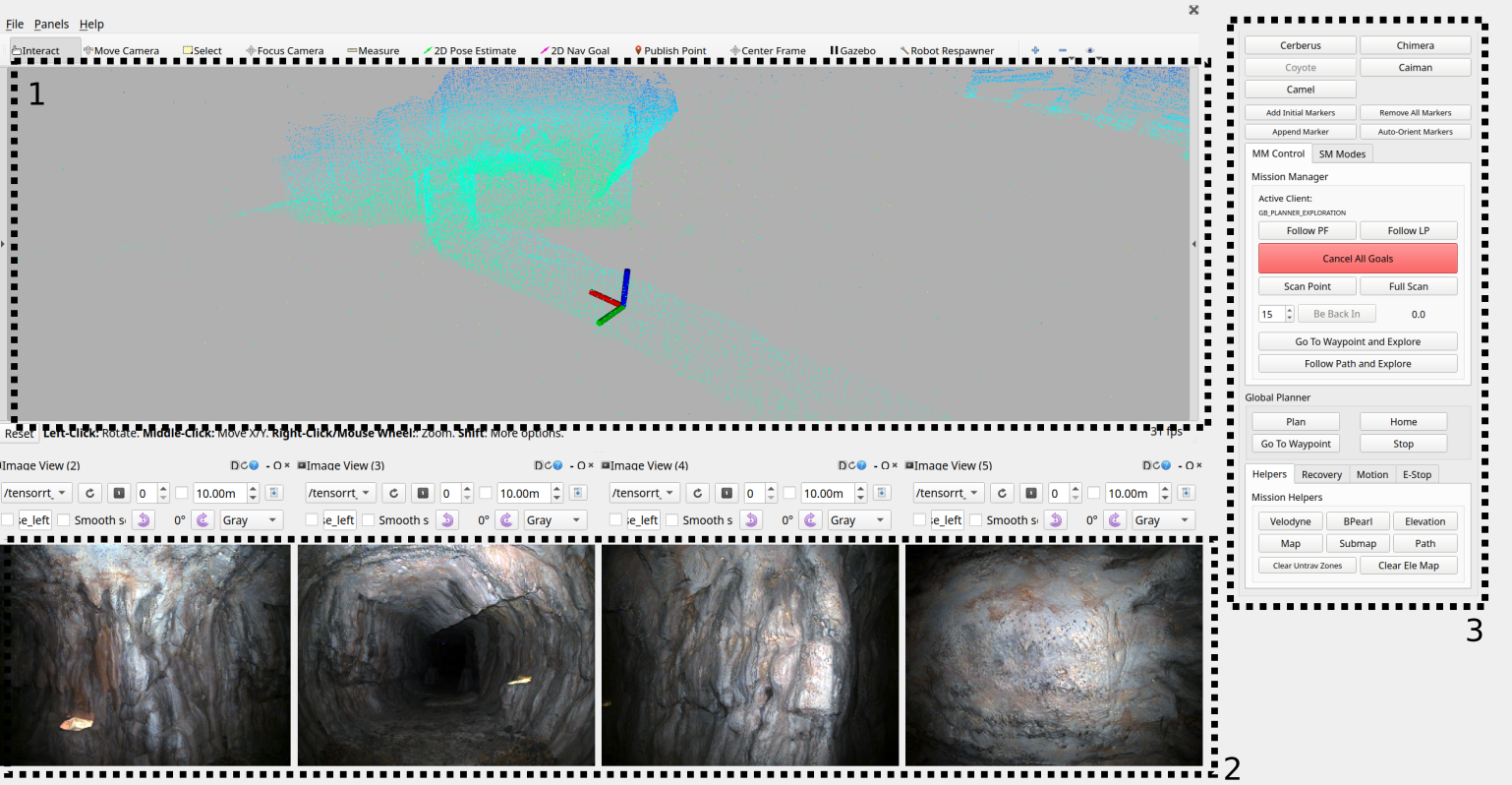}
      \caption{ANYmal Mission Control Interface}
      \label{fig:base_station_mci}
    \end{subfigure}%
    \begin{subfigure}{.48\textwidth}
      \centering
      \includegraphics[scale=0.11]{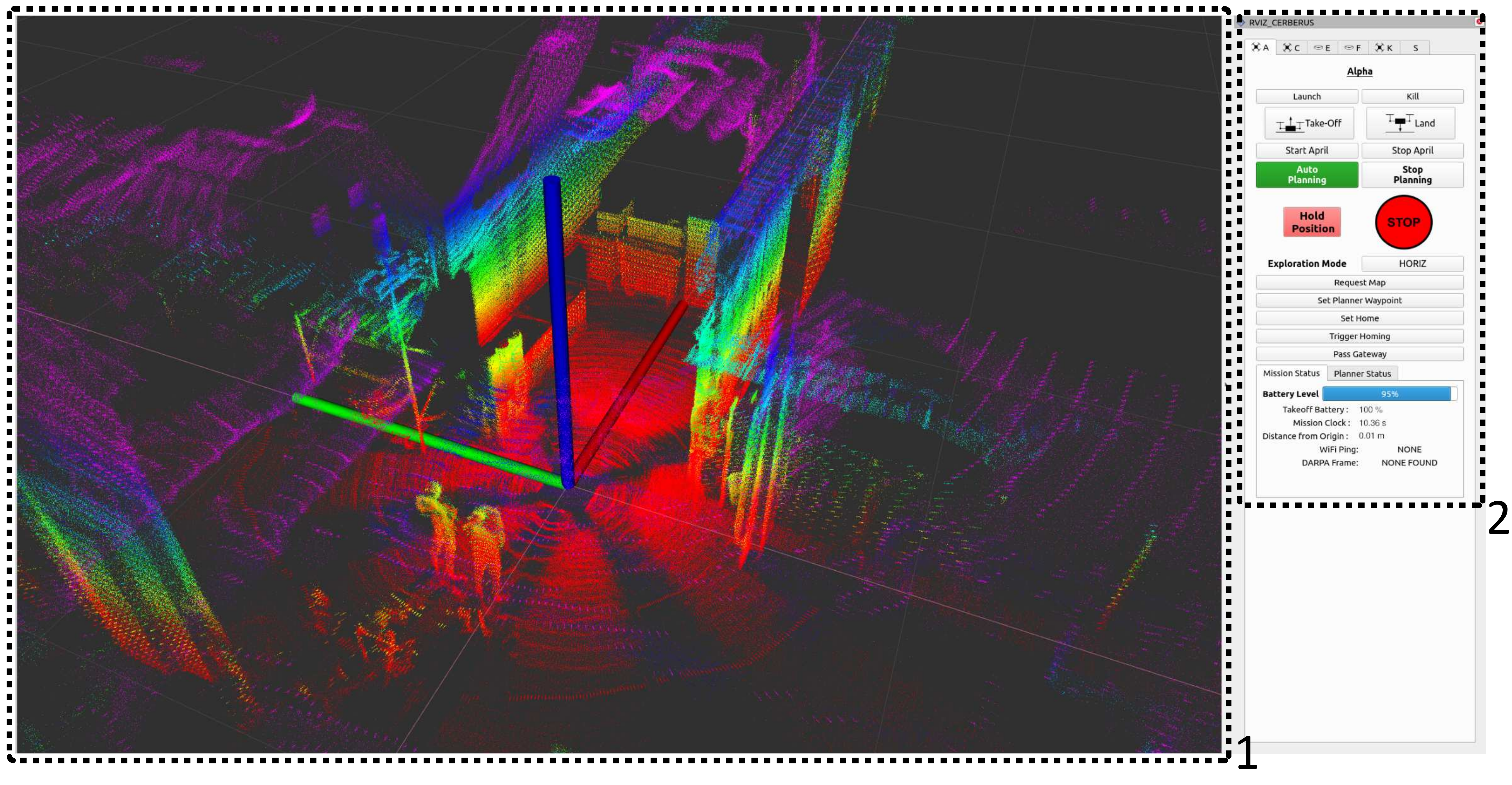}
      \caption{Aerial and roving robots Mission Control Interface}
      \label{fig:aerial_robot_mci}
    \end{subfigure}
    
    \caption{Base Station setup for the single Human Supervisor. (\subref{fig:base_station_setup}) Setup of the Base Station at the Final Event with a single computer and two screens, showing the two main GUIs used to supervise the deployed robots.
    (\subref{fig:base_station_mari}) Monitor and Artifacts Reporting Interface (MARI) was the main interface to provide situational awareness to the Human Supervisor. (\subref{fig:base_station_mci}-\subref{fig:aerial_robot_mci}) Mission Control Interfaces (MCI) used by the Human Supervisor to provide high-level commands to a selected robot, specialized based on agent's type.%
    }
    \label{fig:base_station}
    \vspace{-1ex}
\end{figure}

Our experience in the first two competition events showed that errors made by the operator were a significant fraction of mistakes during the runs (due to the general pressure the competition, controlling multiple robots via several computers and not enough practice operating in unknown environments). These findings are similar to those reported by several teams in DARPA Robotics Challenge~\cite{atkesonhappened}. In the Tunnel and Urban events, our Human Supervisor operated different computers to interact with different robots and to handle mapping and artifacts information. Building on the lessons learned from those events, we developed an improved version of the human-robot interface before the Final Event. A single computer was used to control all the robots and to aggregate mapping information and artifacts reports. The supervisor used two main software modules to interact with the deployed robots: a \ac{MCI} and a \ac{MARI}, exposed through a \ac{GUI} (see Figure~\ref{fig:base_station}). Both interfaces ran on the same computer at the Base Station and used a multi-master architecture with the Nimbro Network software package (described in section~\ref{sec:nimbro_network}) to communicate with the deployed robots.

\textbf{Mission control interface}\\
Through the \ac{MCI} (Figure~\ref{fig:base_station_mci},\ref{fig:aerial_robot_mci}) the operator could select and interact with a specific robot. Specifically, the supervisor could (a) give high-level commands, (b) visualize throttled and compressed image streams to either inspect an area or manually report artifacts, (c) request specific sensor data to evaluate the current situation better (``on-demand'' requests).
The supervisor had several options for high-level command and control. We include here a list of the most relevant ones: start autonomous exploration; define the exploration bounding box to restrict the area of interest; start homing procedure; cancel current task; relocate to a selected frontier and start exploration from there; scan surrounding area or selected point with the inspection payload (ANYmal C Explorer robots specific); move to a desired waypoint. Only the selected robot in the \ac{MCI} was allowed to send image streams to the Base Station, as specified in section~\ref{sec:nimbro_network}. Allowing only one robot to stream images helped the supervisor to focus only on one robot per time and also decreased the wireless bandwidth utilization.  Finally, the operator could request additional sensor data to better evaluate the robot's current state and surrounding area. These data included: CompSLAM map, LiDAR scans (finer resolution than the CompSLAM map) and elevation map snaphosts (legged robot specific).

Figure~\ref{fig:base_station_mci} shows the \ac{MCI} for the ANYmal C robots and highlights three main panels: an RVIZ view (1) showed essential data such as a frame representing the robot's pose in the map and the requested sensor data by the supervisor; a bottom panel (2) displayed the current image streams sent by the selected agent; a side RQT panel (3) exposed control buttons to the supervisor to interact with the selected robot.
Similarly, Figure~\ref{fig:aerial_robot_mci} shows the \ac{MCI} for the aerial and roving robots. Two panels are highlighted in this figure: an RVIZ view (1) similar to that of ANYmal; a side RQT panel (2) containing the control buttons for interacting with the robot.

\textbf{Monitor and artifacts reporting interface}\\
The \ac{MARI} (Figure~\ref{fig:base_station_mari}) was the main interface to provide situational awareness to the Human Supervisor. This \ac{GUI} was divided into two main parts, one for artifacts reporting and one for monitoring of the mission state. The artifacts part contained a panel notifying the supervisor of all new automatic artifact detections. If a robot sent a new artifact report to the Base Station, the Human Supervisor was notified with a sound. The associated camera image was shown for context, annotated with the artifact's bounding box and class. The operator could either accept, reject or modify (class type) the artifact report. The monitoring part presented a top-down view of the global map created during the mission and some extra markers. Cylindrical markers indicated the last received positions of all the robots. Spherical markers in the map showed the locations of the accepted artifact reports. The operator could click on each of these artifacts' interactive markers and either send them to the DARPA Command Post to try to score a point or delete them. Moreover, in the monitoring part, the operator could visualize where each robot was and what was its primary mode of operation (exploration, waiting for commands, homing, etc.). The goal of this view was to provide high-level information and help the operator's decision-making process.

Figure~\ref{fig:base_station_mari} highlights the \ac{MARI}'s three main panels: an RVIZ view (1) showed the global map, the last received position of each deployed robot and the locations of the accepted artifact reports, represented by sphere and cube pins. Spheres represented unoptimized locations reported by the agents, whereas cubes represented optimized positions. Each artifact report could therefore have one or two associated pins, based on whether its position had been already optimized or not. Every artifact report's pin was an interactive marker that the supervisor could use to report its location to the DARPA Command Post or delete it from the map. Artifacts that were not reported yet were colored blue if the optimized location was available and orange otherwise. After a report attempt, the associated pins turned green if a point was scored and red in case of an unsuccessful report. The locations of deployed breadcrumb nodes were indicated in purple; a side panel (2) showed incoming artifact detections (top). The user could accept or reject the detection and manually change the artifact class in case the detected class was not correct (middle). Upon acceptance, the interactive marker of the artifact was added to the global map (1). When the user selected an interactive marker in the global map, its corresponding detection image was shown in the lower right corner; (3) allowed enabling and disabling of specific robots in the optimization as well as changing several settings of it.

\textbf{Modes of operation}\\
Team CERBERUS envisioned a squad of highly autonomous robots coordinated by human reasoning to drive the exploration of unknown environments. As a result, the deployed systems worked in a ``Supervised Autonomy'' mode. Supervised meant that the single Human Supervisor used the information available at the Base Station to take high-level decisions and assign tasks to each robot toward one collective exploration strategy. Each system then undertook the execution of its task autonomously. Specifically, at each moment, the operator evaluated the current high-level situation (global map, position, and mode of operation of each robot) through the \ac{MARI}. If the supervisor wanted to get more information (camera streams, sensor data) of a specific robot, then they selected that agent in the \ac{MCI} and requested the additional information.

Furthermore, through the \ac{MCI}, the supervisor could assign or change each robot's task. First, a robot could be tasked to explore a specific unknown area while ensuring it remained within communication range. If it went beyond the communication range after a certain step, it would backtrack to a point with connectivity. Second, a robot could also be tasked to explore unknown regions away from the communication range, given a set time budget after which it would return to the connection range. A third mode allowed the Supervisor to command the robot to reach a frontier close to unmapped areas and trigger autonomous exploration. Last, as a last resort, the operator could provide a high-level waypoint in the map and request the robot to find an admissible path to reach it.

The part of the Human Supervisor shifted from an ``operator'' role during the Tunnel Event, telling each robot where to go and how to go there, more and more toward a ``supervisor'', coordinating a group of robots without focusing on the motion of a single agent. This mode of operation incorporated our vision of ``Supervised Autonomy'' for the SubT Challenge.

%% file: 04_00_results.tex
This section focuses on the results from the deployment of Team CERBERUS in preparation of and during the Final Event of the SubT Challenge.
\subsection{Individual experiments by subteams and preparation}
We performed a series of field tests to evaluate the performance of our walking and flying robots system-of-systems and prepare for participation in the Final Event. Reflecting the geographical organization of our team, we executed several field deployments spanning four different countries (USA, Switzerland, Norway, and UK). We explored different types of underground mines, subterranean urban infrastructure, and cave environments.

\input{04_01_individual_experiments}

\subsection{Final event}
\input{04_02_final_event}

%% file: 04_01_individual_experiments.tex
\subsubsection{Field deployment preparation}\label{sec:field_deployment_preparation}
From the beginning of the SubT Challenge, we continuously improved our field deployment and software testing procedures to ensure the robots were ready to execute periodic mock-up missions. We employed different tools and strategies to achieve that. On one hand, we used modern development tools such as ``git'' for version control and used \ac{CI} pipelines for automated testing of the individual software modules. While these software development approaches were important, field deployment is robotics specific. Our approach to field testing changed according to our experience in the three years of the competition. As CERBERUS was a diverse international team with expertise across countries and continents scheduling of major field testing events was of paramount importance to integrate disparate contributions and build a team spirit. The following paragraphs contain the most relevant findings we gathered during this time.

\textbf{Weekly shakeout}\\
Regardless of the frequency of the field deployments, we executed a general functionality test of core systems once a week. We referred to this event as a ``weekly shakeout''. We introduced this procedure after the Tunnel Circuit event and committed to testing the overall functionalities of our robotic systems once a week in a more straightforward but representative enough scenario. Weekly shakeouts usually took place on Fridays and involved a subset of all the robots. The involved members updated the software the robots used for testing. First, team members performed a functionality tests in the lab, where they verified the functionality of the basic features (autonomous planning, \ac{SLAM}, artifacts detection and reporting, etc.). After this first step, the team and robots moved to a more representative but still easy to access location, such as a university basement featuring multiple rooms, corridors or a garage. The testing team verified that the main software and hardware components worked during a full mission mock-up. Even though these environments were relatively small compared to the actual scale of the SubT Challenge environments, we could execute mission runs lasting circa \SI{20}{\minute} and collected valuable data about the overall status of our deployed solution.

\textbf{Field deployment}\\
To gather as much experience in the field as possible and to develop a unifying robotics solution for the different and diverse types of underground settings (tunnel-like, urban, and cave networks), we tested in various subterranean environments in preparation for the finals.
Figure~\ref{fig:environment_collage} shows a selection of the different environments we tested in, whereas Table~\ref{tab:field_deployments} includes the list of extended (daily) field deployments.
We divided the field deployment into two main phases: ``pre-deployment'' and ``deployment''.

\emph{Pre-deployment} - This phase started roughly one week before a scheduled field test. Firstly, the team agreed on which new features had to be tested in that deployment.
Secondly, we did an integration test of the new desired features one day before the field test as part of a mission shakeout. Each robot had to pass the shakeout individually to be cleared for the field test.
If the team detected a malfunction during the shakeout phase, they could either try to solve it or decide to use a previous, tested version of the same module, without the detected issue. If the team performed either a software fix or a downgrade, all the robots had to switch to the same software version and be tested again. 
Once every robot passed the shakeout test, it was marked as ``mission-ready'' and allowed no new hardware or software updates. With these constraints, the team ensured that on the field deployment day, they were ready to go to the selected underground location and start testing without any hindrances due to unforeseen issues related to the latest software or hardware updates. 
After all the robots were ready and packed, we discussed and agreed on the field deployment schedule and wrote down a testing plan.

\emph{Deployment} - We scheduled at least one field deployment in an underground setting per month in a different location to avoid overfitting to a specific site (underground mine, urban environment, cave networks). 
Each field deployment included several mock-up missions in a configuration as similar to the SubT Challenge runs as reasonably possible. 
Specifically, the team set up a mock-up gate with four Apriltags so that each robot could execute the ``DARPA-frame'' detection before starting its mission. 
Moreover, we prepared the Base Station, where the Human Supervisor coordinated the mission runs, as required during an actual competition event. Finally, we distributed the artifacts along the course to verify that the robots could detect them and report their locations. 
Before each mission test, the Pit Crew members prepared the robots with the same modalities used for the official missions.
In the year before the Final Event, we had two people training as the Human Supervisor.
Each field deployment happened despite the status of the features under development, and its primary goal was to assess the overall mission performance. 
Each executed mission was a snapshot of the system's readiness in a competition-like environment and provided essential information to the team. Furthermore, each field test contributed to the team's readiness for the final event.

\textbf{System launch}\\
Before each mission, the Pit Crew launched the software in each robot using ssh-shell sessions running \emph{tmux}\footnote{\url{https://github.com/tmux/tmux}}. Tmux is a terminal multiplexer used to create persistent sessions on each robot where the launched software kept on running even if the original ssh-shell session disconnected. Moreover, we could save the complete output history of each onboard session to file for post-processing analysis and debugging of every mission.

\textbf{Onboard data recording}\\
Robots recorded extensive data during their mission executions for post-mission analysis. These data included mission-related information (such as robot state and odometry, onboard map, planned paths, etc.), as well as raw sensor data (LiDAR packets, camera streams, etc.) and other onboard compute information (for example, CPU, memory, GPU usage, and temperature). Given the complexity of the software running onboard, we had to carefully select which data to record since recording all the available information for possibly more than \SI{60}{\min} (one entire mission duration) was not feasible due to onboard memory limitation and processing overhead created by the recording process itself. Therefore we focused on the information needed to perform post-mission analysis and debugging and privileged, when possible, ``raw information'' required to reconstruct the robot state in post-process.

\textbf{Mission data analysis}\\
After every mission test, we thoroughly analyzed the onboard sessions' logs and recorded data from each robot. This procedure included writing down a summary report highlighting the mission's main outcome and any detected issue that required further analysis. Additionally, we generated a series of plots using the onboard recorded data for each test. We found the tool \emph{plotjuggler}\footnote{\url{https://github.com/facontidavide/PlotJuggler}} to be extremely helpful for this purpose. For example, by visually inspecting these plots and looking for an unexpected trend, we could detect if any silent bug was introduced (such as a memory leak, abnormal CPU or GPU usage, etc.).

\subsubsection{Qualitative results from preparation tests}\label{sec:qualitative_deployments}

\begin{figure}[]
    \centering
    \includegraphics[width=0.95\linewidth]{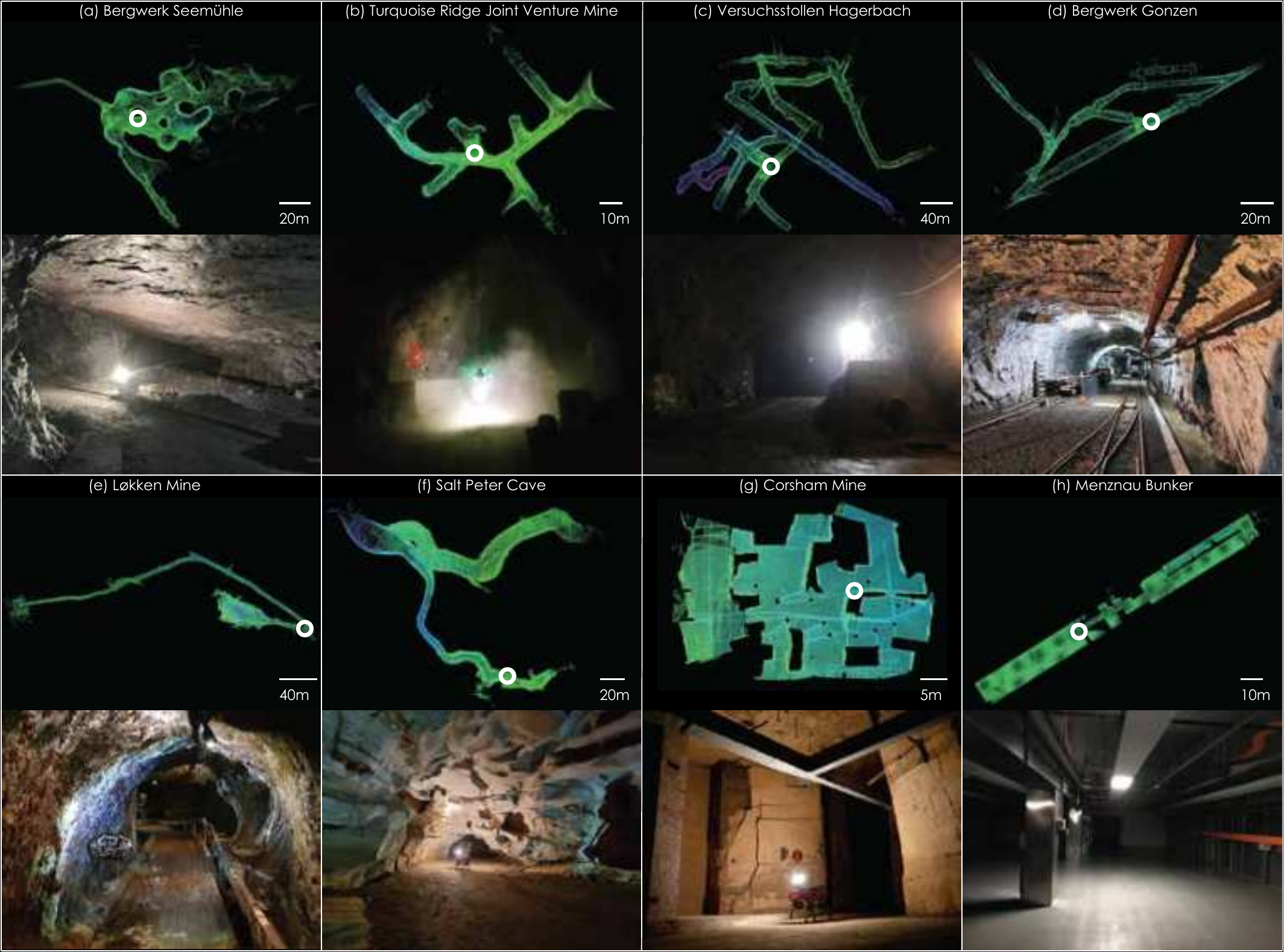}
    \caption{A subset of different underground environments used during the ``deployment'' phase of field-tests during our participation throughout the SubT Challenge. The first and third rows represent a top-down orthographic view of the reconstructed pointcloud maps of the environments while the second and fourth rows capture a third-person view in these environments. The locations of the image captures are presented in the corresponding maps by a white hollow circle. The challenges presented by each of the environments are discussed in section~\ref{sec:qualitative_deployments}}
    \label{fig:environment_collage}
\end{figure}

Different challenges were expected from the diverse environments of interest (tunnel-, urban- and cave-like) including a) geometric self-similarity, b) visual obscurants (e.g., smoke and dust), c) lack of visual texture, d) multi-level settings, and e) wide or extremely narrow environments. This guided our selection of a representative set of specific environments for conducting field tests. We expected challenges in terms of their complexity, traversability, effect to communications, scale and perception. To this end, we conducted field deployments in a set of environments that can be categorized as tunnel-, urban- or cave-like in preparation for the challenge. Table~\ref{tab:field_deployments} presents a complete list of all the field deployments conducted over the course of the competition (\num{2018}-\num{2021}) divided by category. A representative subset of these environments are shown in Figure~\ref{fig:environment_collage}, with a top-down view of the 3D maps and an image, respectively. The environments presented in Figure~\ref{fig:environment_collage}-(a-e) are representative of tunnel-like environments, while Figure~\ref{fig:environment_collage}-(f) and Figure~\ref{fig:environment_collage}-(g,h) are representative of cave-like and urban environments respectively. In terms of perception, these environments presented challenges to our entire exteroceptive sensor suite albeit not necessarily simultaneously. For LiDAR, Figure~\ref{fig:environment_collage}-(a, e) presented a sudden change in scale, Figure~\ref{fig:environment_collage}-(c) multiple levels and Figure~\ref{fig:environment_collage}-(a, h) geometric self-similarity while for a visual camera, Figure~\ref{fig:environment_collage}-(a, b, f, g) presented darkness, Figure~\ref{fig:environment_collage}-(a, b) heavy airborne dust and for a thermal camera, cave environments like Figure~\ref{fig:environment_collage}-(f) presented thermal flatness.

\begin{table}
\centering
\begin{tabular}{c|c|c|c|c}
\toprule
\multicolumn{5}{c}{\textbf{Field Deployments}} \\
\midrule
\textbf{Location} & \textbf{Environment} & \textbf{Country} & \textbf{Period} & \textbf{Deployments} \\
\midrule
Wangen a. d. Aare - ARCHE & Urban & Switzerland & 2019-2021 & 15 \\
Bergwerk Gonzen & Tunnel & Switzerland & 2019-2020 & 12  \\
Versuchsstollen Hagerbach & Urban/Tunnel & Switzerland & 2021 & 6 \\
Bergwerk Seem\"uhle & Tunnel/Cave & Switzerland & 2021 & 4 \\
Menznau Bunker & Urban & Switzerland & 2019 & 3 \\
Baustoffe Schollberg & Tunnel & Switzerland & 2021 & 2 \\
TRJV Mine & Tunnel & USA & 2018-2019 & 12 \\
Comstock Mine & Tunnel & USA & 2018-2020 & 6 \\
Black Diamond Mine & Tunnel & USA & 2019 & 1 \\
Wampum Underground & Tunnel & USA & 2019 & 1 \\
Applied Research Facility & Urban & USA & 2019-2021 & 12 \\
Moaning Caverns & Cave & USA & 2020 & 1 \\
Great Saltpetre Preserve & Cave & USA & 2021 & 1 \\
L\o kken Mine & Tunnel/Cave & Norway & 2020-2021 & 4 \\
Corsham Limestone Mine & Tunnel & UK & 2019-2021 & 6 \\
\bottomrule
\end{tabular}
\caption{Overall robotics field deployment of Team CERBERUS in preparation for the DARPA SubT Challenge. Every reported deployment counts as a full day, including several mission mock-ups and tests.}
\label{tab:field_deployments}
\end{table}

%% file: 04_02_final_event.tex
The SubT Challenge's Final Event took place at the Louisville Mega Cavern in Kentucky on $21-24$ September \num{2021} and involved eight teams for the Systems track and nine teams for the Virtual track. For the Systems track, as side from Team CERBERUS, the following teams competed: CSIRO Data61, MARBLE, Explorer, CoSTAR,  CTU-CRAS-NORLAB, Coordinated Robotics, and Robotika. The competition course included elements from all the subdomains (Tunnel, Urban, and Cave) combined into a unified, integrated environment. The course\footnote{https://s.ntnu.no/subt-finals-course} had a single entrance, composed of an inclined tunnel and a steep turn, followed by an opening containing three different access breaches leading to different sections for each subdomain. Multiple connections and crossover points existed between the subdomain areas. Thus, the final course had a complex, heterogeneous, multi-level morphology exposing the teams to all the different challenges that each subdomain offered. Accordingly, the Final Event represented a grueling challenge combining the mobility, perception, autonomy and communications challenges envisioned by DARPA. In the leaderboard of the Systems track's Prize Round, which took place on September \num{23}, both teams CERBERUS and CSIRO Data61 scored \num{23} points, followed by teams MARBLE (\num{18} points), Explorer (\num{17} points), CoSTAR (\num{13} points), CTU-CRAS-NORLAB (\num{7} points), Coordinated Robotics and Robotika (both with \num{2} points).
As both CERBERUS and CSIRO Data61 were tied with \num{23} points each, the tiebreaker rule had to be invoked. This rule denoted that the team which reported its last score the earliest would win: CSIRO Data61 reported their final point with under \SI{30}{\second} remaining, whereas CERBERUS reported our last artifact with more than \SI{1}{\minute} left. Team CERBERUS won the DARPA Subterranean Challenge as well as the associated \$\num{2} million (USD) prize reward.

\subsubsection{Preliminary runs}
Two test runs on September \num{21} and \num{22} \num{2021}, called ``Preliminary Runs'', preceded the Prize Round and served as a testing bench for the teams before the scoring final run. Each Preliminary Run lasted \SI{30}{\minute} and included \num{20} artifacts. The competition course was re-configured between each Preliminary Run and before the Prize Round. Each team's score during these runs was not used to determine the final ranking in the Prize Round, which was based solely on the Prize Round result. Teams were also not allowed to use any prior map information or other data collected from the preliminary runs.

The afternoon before each Preliminary Run, some team members gathered to discuss the deployment strategy for the upcoming run (how many robots to use, which robot to deploy first, etc.). In the meantime, the other members finalized data analysis and required fixes in software and hardware. Upon agreement on the deployment strategy, each subteam (walking, flying, autonomy, multi-robot mapping) worked together to finalize the system configuration for the run. This work included software updates on the agreed modules and final verification of each system, to be completed within the same evening. Selected team members checked each robot (for instance, to verify that all the LEDs turned on, sensor glass was cleaned, plugs correctly connected, etc.) and subsequently tested our overall system in a small mock-up mission to verify that all the significant functionalities were correctly executed. This mock-up mission resembled the tests performed during the ``weekly shakeout'' sessions highlighted in section~\ref{sec:field_deployment_preparation}.

\textbf{Preliminary run - day one}\\
Team CERBERUS deployed three ANYmal C SubT agents and the roving robot during the first Preliminary Run. Our overall strategy for this run was a trade-off between testing our robotic solution in the course while being cautious enough not to damage the agents by following more risky paths in terms of traversability characteristics. Specifically, we used rather conservative settings for the ANYmal's navigation planner to prevent the robots from tackling dangerous but still feasible routes.

We deployed an ANYmal C SubT Explorer, two Carriers, and the roving robot. The Human Supervisor assigned a different area of exploration to each ANYmal based on the information he received from the first deployed robot. The supervisor then drove the roving robot in the course and parked it at the intersection with the three subdomain openings to extend the wireless coverage of the agents and support underground operations. The supervisor tasked the Explorer with the autonomous exploration of what he perceived as the Tunnel subdomain area, accessible from the course entrance. Moreover, he allowed this agent to keep on exploring even if it went beyond the communication range. Afterward, the supervisor assigned the first Carrier robot to explore the Urban section. While this Carrier proceeded further, the third ANYmal robot was deployed and set to explore the Cave subdomain. The Explorer autonomously navigated through the Tunnel section and reached the intersection with the Urban area, where one of its feet got stuck in a railway track and caused an unexpected shutdown of the agent due to over-current demands while trying to proceed forward.

The Carrier sent into the Urban area also explored this section autonomously, apart from two occasions where the Human Supervisor explicitly provided waypoints. In the first case, the exploration planner provided waypoints to climb a short flight of stairs, but the ANYmal's navigation planner did not plan a path due to the conservative settings mentioned at the beginning of this paragraph. In the second case, a long descending flight of stairs connected the Urban and the Tunnel environment, but its bottom part was never seen by the LiDAR sensor placed on top of the Carrier, due to its limited field-of-view. Thus, the exploration planner could not provide any path to descend the staircases. Once the Human Supervisor realized that the robot could not proceed further, he provided a waypoint to command the robot downstairs directly. Lastly, the third deployed robot autonomously navigated in the narrow passages of the Cave section. During the first Preliminary Run, Team CERBERUS scored \num{7} points and led the Systems' preliminary round tie with team CSIRO Data61. Importantly, we learned that the environment of the DARPA Subterranean Challenge final event started with a narrow tunnel. We therefore identified that flying robot deployment from the staging area could be both risky and of limited value. 

\textbf{Preliminary run - day two}\\
For the second Preliminary Run, we decided to start with the same settings as the first run and apply a few modifications to evaluate further which configuration to use for the scored run of the day after. First, we did not allow any robot to explore beyond the communication range for this run. This constraint was an extra safety feature on top of the conservative tuning from the first day: we wanted to make sure to have all the robots available for the Prize Round. Second, we deployed the exploration planner with a modification to the global planning stage. Since the planner had a binary constraint on the inclination of the edges in the graph, we observed that the planner could provide paths passing areas too steep for the robot to traverse safely but satisfy the inclination constraints according to the low resolution volumetric map (\SI{0.2}{\meter}). The issue was prominent in the global planner due to the static nature of the global graph. Hence, a modification was made that calculated $k$ shortest paths towards the goal in the global graph and commanded the one having the least inclination value on its steepest edge.
We also re-tuned the detection thresholds for Bluetooth and gas artifacts.

We deployed two ANYmal C SubT Explorers, one Carrier, and the roving robot. The Human Supervisor assigned each ANYmal to autonomously explore different sections: the first Explorer to the Cave environment, the second Explorer to the Urban environment, and the Carrier to the Tunnel area. The deployed robots explored less unknown space than in the previous run due to the additional constraint of maintaining wireless communication with the Base Station. Team CERBERUS scored \num{6} points in this run, achieving third ranking in the overall Preliminary Round (determined based on
the sum of the team’s two runs).

\subsubsection{Prize round}
The Prize Round took place on September \num{23}, \num{2021} and consisted of one scored run, lasting \SI{60}{\minute}, where each team had to find as many of the \num{40} artifacts distributed along the course, with the total number of allowed artifacts reports equal to \num{45}. Each team was competing on the same course, one after the other, with particular caution by the DARPA team to ensure that team personnel would not communicate with each other to prevent gaining knowledge from another team's run. This was achieved by ``sequestering'' team personnel and systems from their garage area until they completed their scored run. As it turned out, the course was further modified from the two Preliminary Runs, again without the knowledge of the competing teams, with the addition of dynamic obstacles such as a hidden mechanism that would close an opening after a robot stepped in or a falling object from the ceiling triggered by a moving robot passing close by. 

The afternoon before the Prize Round, team members discussed the final software and hardware configuration and agreed on the deployment strategy. We decided to use settings similar to the ones used during the first Preliminary Run because we were confident about the overall performance of our robots with these configurations. Moreover, we extensively discussed whether to deploy some last software modifications (in the planning and \ac{SLAM} pipelines) that could have improved some detected issues, but we decided not to. We believe that these conservative decisions, avoiding last minute (and not extensively tested) changes was crucial for the overall performance during the scored run. We learned during the previous years of this competition how complex robotic systems are and how everything must be extensively tested before deploying in the field. Afterwards, we tested all the robots with a ``shakeout'' procedure, similar to that mentioned above.

At the Prize Round, we prepared the following robots for deployments: four ANYmal C SubT robots (two Explorers, ANYmal \num{1} and ANYmal \num{2}, and two Carriers, ANYmal \num{3} and ANYmal \num{4}), one RMF-Owl aerial robot, one DJI M100-based Aerial Scout, and the roving robot. First, the two Explorers were deployed: the Human Supervisor utilized online feedback regarding the map of the area and assigned ANYmal \num{2} to explore the Cave section and ANYmal \num{1} to explore the Tunnel section. When the first two robots were exploring their areas autonomously, the supervisor teleoperated the roving robot in the course and parked it close to the entrance of the three different sections, with its directional antenna pointing toward the Tunnel and Urban openings. Afterward, the first Carrier robot, ANYmal \num{4}, was deployed and assigned by the supervisor to explore the Urban environment. While the three ANYmal robots autonomously navigated their given bounding boxes, the last Carrier robot, ANYmal \num{3}, was deployed and sent to the Cave area.

With \SI{18}{\minutes} left in the run, the team decided to deploy the RMF-Owl to explore the higher area of the cavity in the cave section. However, after deploying, the robot went up exploring the upper part of the staging area. This was caused due to the fact that there was no provision to add $3\textrm{D}$ ``no-go zones'' within the staging area (to indicate to the planner that this area is out of consideration) and an existing but deprecated feature of our software for crossing the gate before starting the exploration was not used. The robot was manually landed by the safety pilot and not deployed in the remaining of the run. During the complete run, the Human Supervisor coordinated the mission execution by checking what each deployed robot was doing (autonomous exploration, frontier repositioning, etc.) and, if needed, requested additional actions. Most of the time the supervisor focused on processing the incoming artifact reports from the different robots. At the end of the \SI{60}{\minute} scored run, Team CERBERUS achieved \num{23} points and used all the \num{45} report submissions available.

\begin{figure}[]
    \centering
     \includegraphics[width=\textwidth]{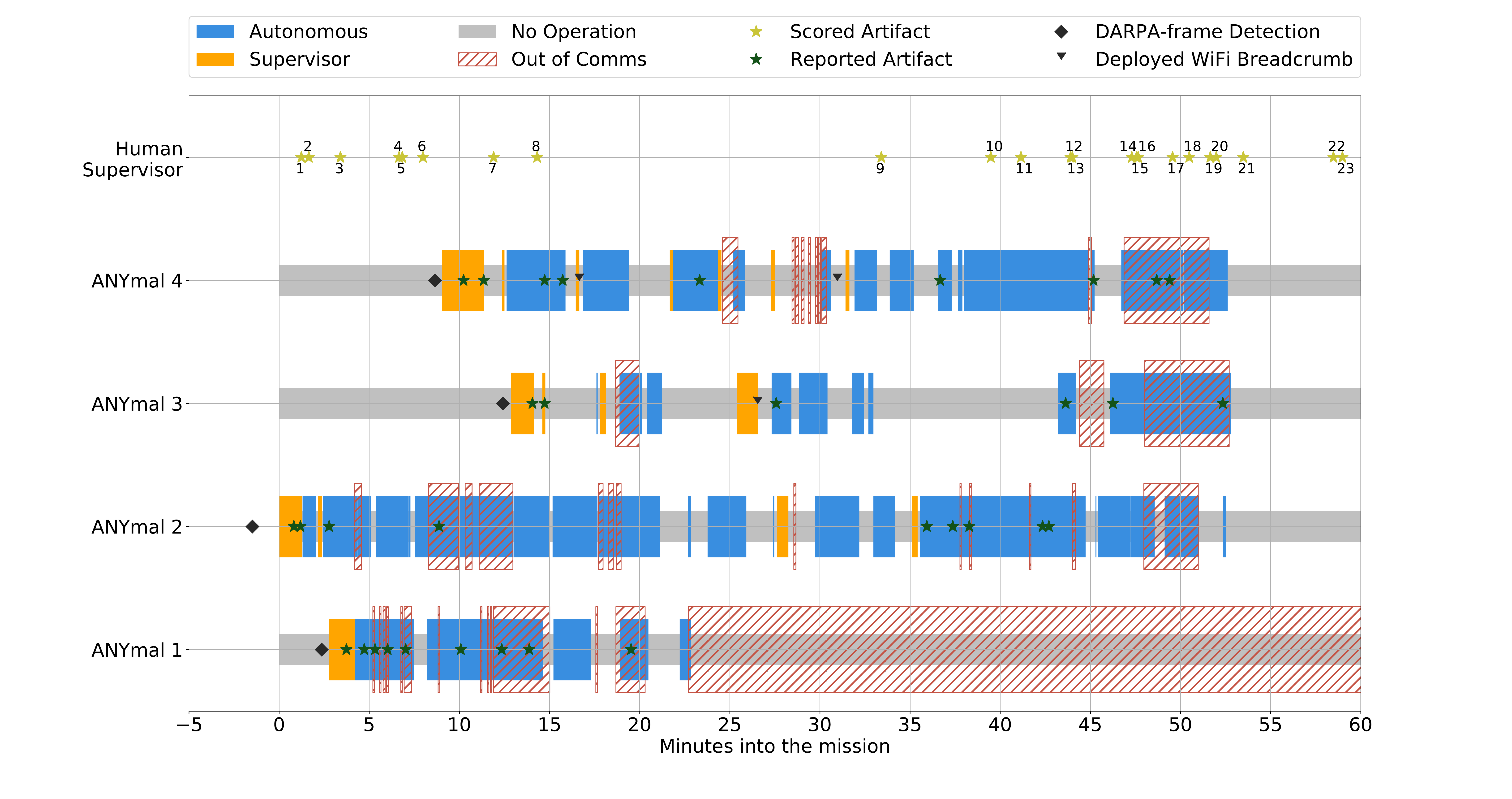}
    \caption{Mission analysis of the winning Prize Round of Team CERBERUS in the DARPA SubT Challenge. The plot shows in which mode each robot operated (``Autonomous'', ``Supervisor'', and ``No Operation'') and when major mission events occurred (``Reported Artifacts'', ``DARPA-frame Detection'', and ``Deployed WiFi Breadcrumb''). An agent was considered out of communication range (``Out of Comms'') if it could not ping the Base Station for more than \SI{3}{\second}. An extra line is used for the Human Supervisor to highlight when artifact reports sent to the DARPA Command Post were scored.}
\label{fig:mision_analysis}
\end{figure}

Figure~\ref{fig:mision_analysis} and Table~\ref{tab:prize_run_statistics} show an overview of Team CERBERUS' Prize Round. Figure~\ref{fig:mision_analysis} details in which mode each robot operated at any given moment in time: mode ``Supervisor'' indicates that the Human Supervisor directly provided a waypoint in the robot's proximity and asked the agent to go there (directly or querying the ANYmal navigation planner to find a traversable path). Mode ``Autonomous'' indicates that a robot was executing a given task completely autonomously and could, if necessary, decide to take different actions (for example, switch from exploration to homing, execute frontier repositioning, etc.). Mode ``No Operation'' shows when a robot was not moving / not operational.  An agent was considered out of communication range (``Out of Comms'') if it could not ping the Base Station for more than \SI{3}{\second}. A correct artifact report (artifact class and position) sent from a robot to the Base Station is marked with a dark green star in the robot's timeline. The Human Supervisor filtered the reports received by the different agents and decided which artifact (class and position) to report to the DAPRA Command Post. Successfully scored artifacts are marked with yellow stars in the Human Supervisor's timeline. Since each robot operated in the course solely based on the information collected onboard and on the commands received by the Human Supervisor, it was not aware of the data collected by the other agents. Therefore, if two robots crossed the same section, they might have reported the same artifacts to the Base Station (hence the number of ``Reported Artifacts'' exceeds ``Scored Artifacts'').

The Figure~\ref{fig:mision_analysis} also shows our overall strategy: we decided to let the robots go out of communication range for a limited amount of time in favor of getting them back in connection range, receiving their artifacts reports, checking the area they explored, and finally assigning them a new task. The assigned budget to each agent for exploration could be decided at the moment, but during the Prize Round, it ranged from \SI{5}{\minute} to \SI{13}{\minute}. A given time budget meant that the robot could explore for that time, and then it had to come back to the starting position, thus resulting overall in a time allotment for the assigned task with a duration double the allocated budget. We believe that such a strategy, involving short and frequent autonomous tasks, with the advantage of the Human Supervisor interacting with the robots at a reasonable frequency, paid off compared to letting every agent go for longer autonomous missions and waiting for them to come back only towards the end of the run. However, our approach still implied a slightly heavier load on Human Supervisor who had to cope with different agents at the same time while checking the artifact reports. Another insight about the mission can be noted in the plot between \SI{15}{\minute} and \SI{33}{\minute}, when the Human Supervisor did not score any points. Specifically, during this time, the supervisor had to control the deployed robots and, at the same time, check a higher number of artifact reports sent by ANYmal \num{4}. This agent sent to the Base Station \num{30} CO$_2$ artifact reports between \SI{10}{\min} and \SI{15}{\minute}. These reports accumulated in FIFO list, called artifacts queue, at the Base Station and had to be analyzed by the supervisor. Such a high number of reports for a single artifact type, detected in a short period, was
caused by erroneous tuning of the threshold used to check the CO$_2$ concentration. We set a low detection threshold for CO$_2$ concentration (\SI{1500}{ppm}, in contrast to the given official DARPA specification of \SI{3000}{ppm}), intending to detect the presence of gas leaks even if not particularly close to the source and relying on the supervisor to decide further where to send the agent in this case. Unfortunately, such a low value created many reports, even though measurements above the set threshold were grouped based on their positions. In fact, within each group, the estimated clustered position of the artifact was allowed to be updated if new measurements were received. Therefore, the more the robot moved close to the gas leak source, the more measurements were above the threshold, and the more often the clustered position was updated and reported to the Base Station. The plot shows only one ``Reported Artifact'' marker, despite of how many times its clustered position was updated.
A similar problem, even if less severe, occurred between \SI{37}{\min} and \SI{42}{\minute} when ANYmal \num{2} sent to the Base Station \num{11} CO$_2$ artifact reports. Finally, the plot shows that in the last \SI{7}{\minute} of the competition, the supervisor focused only on deciding which of the already received artifacts to report to the Command Post, thus he did not send any additional exploration goals to the agents.

\begin{table}[]
\centering
\begin{tabular}{c|c|c|c|c|c|c|c|c}
\toprule
\multicolumn{9}{c}{\textbf{Prize Round Statistics}}\\
\midrule
\multirow{2}{*}{\textbf{Robot}} & \multirow{2}{1.5cm}{\centering \textbf{Run Time}} & \multirow{2}{*}{\textbf{Distance}} & \multirow{2}{2cm}{\centering \textbf{Scored Artifacts}} & \multirow{2}{*}{\textbf{Operational}} & \multicolumn{4}{c}{\textbf{Autonomous}}\\
& & & & & \rotatebox[origin=c]{90}{Explore} & \rotatebox[origin=c]{90}{Home} & \rotatebox[origin=c]{90}{Reposition} & \rotatebox[origin=c]{90}{Recover}\\
\midrule
ANYmal 1 & \SI{57}{\minute} & \SI{240}{\meter} & \num{8} & \SI{31.1}{\percent} & \SI{75}{\percent} & \SI{6.9}{\percent} & \SI{4.5}{\percent} & \SI{13.6}{\percent}\\
ANYmal 2 & \SI{60}{\minute} &  \SI{687}{\meter} & \num{8} & \SI{69}{\percent} & \SI{60.6}{\percent} & \SI{6.6}{\percent} & \SI{25.8}{\percent} & \SI{7}{\percent}\\
ANYmal 3 & \SI{47}{\minute} &  \SI{311}{\meter} & \num{1} & \SI{34.1}{\percent} & \SI{53.4}{\percent} & \SI{11.9}{\percent} & \SI{20}{\percent} & \SI{14.7}{\percent}\\
ANYmal 4 & \SI{50}{\minute} &  \SI{500}{\meter} & \num{6} & \SI{57.5}{\percent} & \SI{81.3}{\percent} & \SI{12.9}{\percent} & \SI{0}{\percent} & \SI{5.8}{\percent}\\
\bottomrule
\multicolumn{1}{l}{} & \multicolumn{1}{l}{} & \multicolumn{1}{l}{} & \multicolumn{1}{l}{} & \multicolumn{1}{l}{} & \multicolumn{1}{l}{} & \multicolumn{1}{l}{} & \multicolumn{1}{l}{} & \multicolumn{1}{l}{}
\end{tabular}
\caption{Statistics of the winning Prize Round of Team CERBERUS in the DARPA SubT Challenge. For each robot, the following information are provided: ``Run Time'' indicates for how long each robot operated in the course; ``Distance'' holds the travelled distance by each agent; ``Scored Artifacts'' indicates how many artifacts of the total scored ones were reported by the agent; ``Operational'' indicates the operating percentage with respect to the agent's time in the course (Run Time); ``Autonomous'' holds the percentages, split by type, when the agent was in autonomous mode - percentage with respect to the operational time.}
\label{tab:prize_run_statistics}
\end{table}

\textbf{Artifacts scoring}\\
The artifacts during the Prize Round were detected by the robots using their onboard artifact detection framework. For each artifact, the detections were filtered and reported back to the Base Station. The Human Supervisor could see the detections on the \ac{MARI} as shown in Figure \ref{fig:base_station_mari}, which was then used to accept, or reject the reports or to change the class of a detected artifact. Team CERBERUS detected \num{23} artifacts during the Prize Round. These artifacts were detected by the four ANYmal robots, were initially localized based on the CompSLAM solution onboard each robot and the M3RM framework was (at instances) utilized to correct their position estimates on the Base Station. The Human Supervisor added each incoming artifact report from the robot to the map in the MARI and reported the artifacts visualized in the map. For some artifacts reported initially, the Human Supervisor chose to utilize the M3RM solutions, while others were reported using the onboard CompSLAM estimate. Since the M3RM framework required more time to optimize the position as the map grew, this was used in the beginning, whereas later in the mission, the onboard estimate was used.

\begin{figure}[]
    \centering
    \includegraphics[keepaspectratio,width=0.975\textwidth]{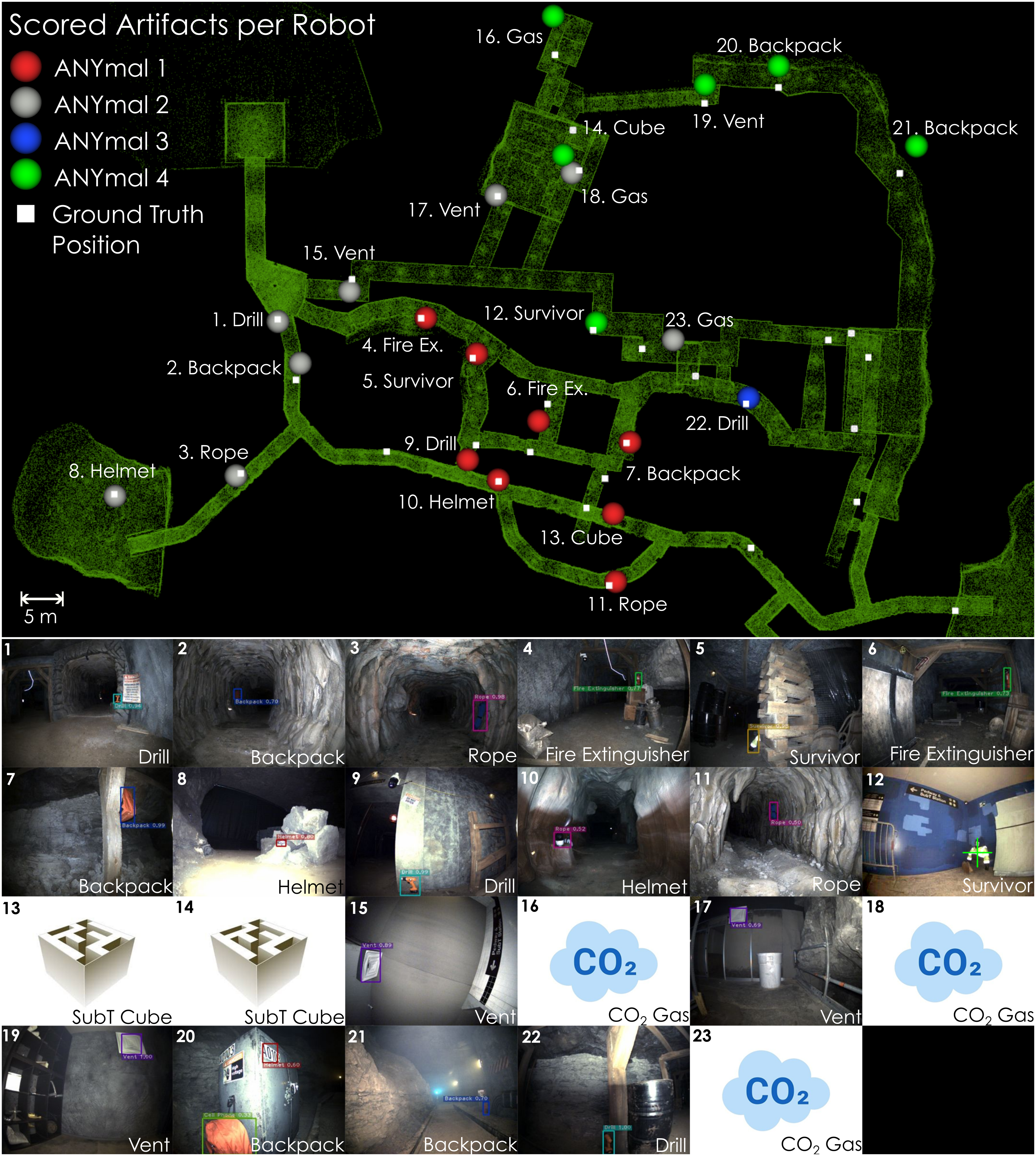}
    \caption{Scored artifacts during the Final Round. The colored circles represent the locations of the artifacts detected by each robot against the ground-truth (in white). The images of artifacts numbered \num{10} and \num{20} show erroneous classes. The supervisor accepted the reports from the robot, changed the class types and then sent the revised reports to the DARPA Command Post. The locations of fourteen artifacts were estimated using the robot's onboard CompSLAM output, while those of three artifacts were refined using the M3RM (artifact numbers \num{6}, \num{9} and \num{15}). Five artifacts involved Bluetooth or $\textrm{CO}_2$. Finally, the Human Supervisor manually detected one artifact (artifact number \num{12}) while checking the image streams sent from one agent. A video showing the Team's Prize Round is available at~\url{https://youtu.be/QON8IFc8cjE}}
    \label{fig:reported_artifacts_with_yolo_images}
\end{figure}

Figure \ref{fig:reported_artifacts_with_yolo_images} chronologically numbers the artifacts that were correctly reported to the DARPA Command Post visualized alongside the ground-truth pose of the artifacts. The ground-truth locations of the artifacts were made available by DARPA only after the completion of the Final Event. The ground truth map of the environment provided by DARPA is shown in green. The figure also shows the images captured by the onboard sensors of the respective robots at the time of detection. The Human Supervisor reported the refined positions of the artifacts numbered \num{6}, \num{9} and \num{15} as optimized by the M3RM, while the artifact numbered \num{12} was identified by observing the image streams on the MARI. Artifact localization was initiated onboard the robot for this instance by the supervisor by clicking on the artifact inside the image.

\begin{figure}[]
    \centering
    \includegraphics[keepaspectratio,height=7.0cm]{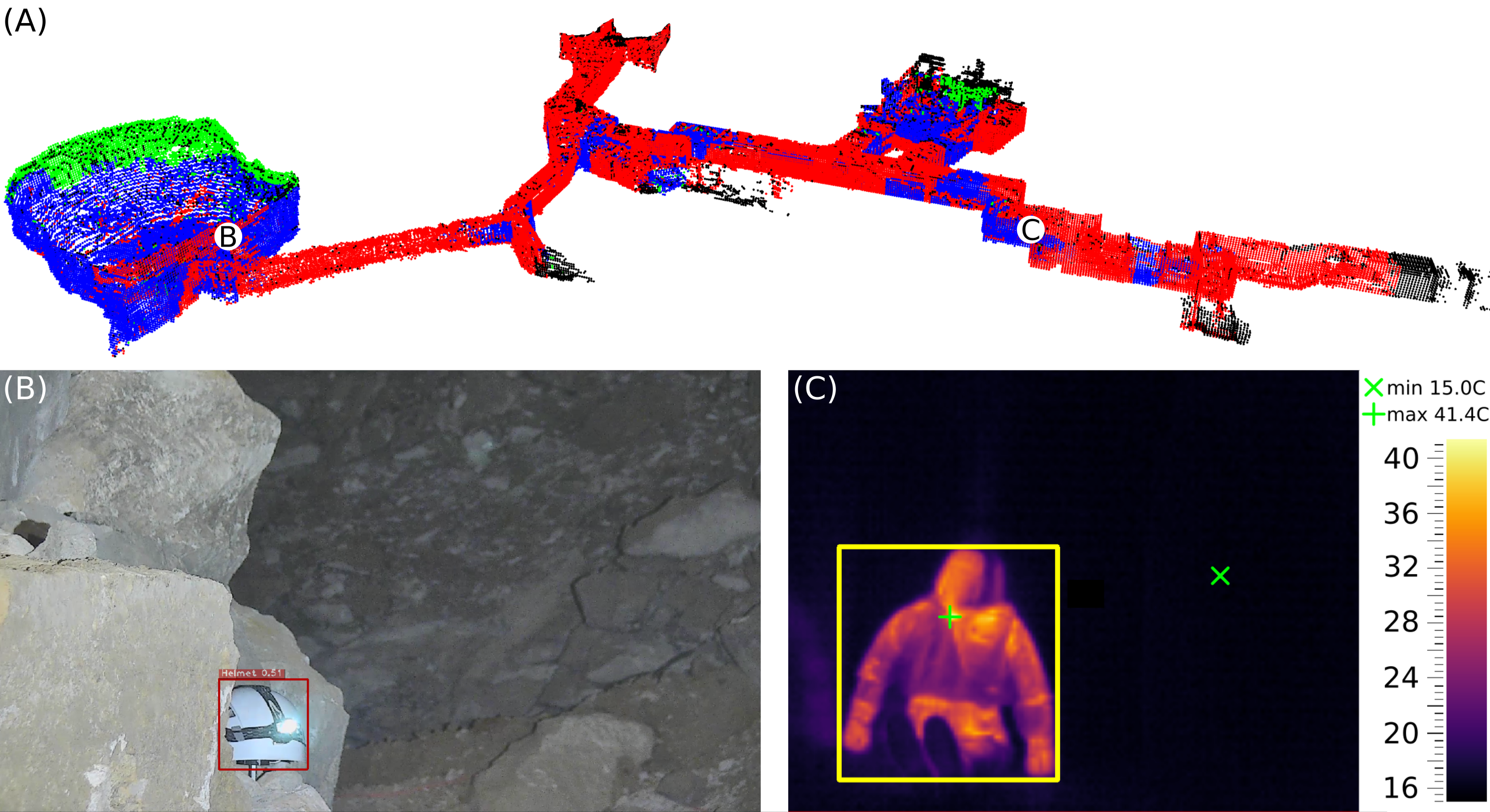}
    \caption{ANYmal's inspection payload artifact detection. (A) The scanned area after the run for ANYmal \num{2} (Explorer robot). Black points are unseen, red points were only scanned by the Alphasense Core colored cameras, and green points only by the inspection payload zoom camera. Blue areas were scanned by both sensors. Artifact detections performed with multi-sensor modalities: (B) visual detection of an helmet with the zoom camera and (C) thermal detection of a survivor. The letters B and C in image (A) display the point of the respective detections.}
    \label{fig:inspection_payload_results}
\end{figure}

\textbf{Inspection payload} \\
The main purpose of the Explorer robot's inspection payload was to scan areas classified as being far away from the Alphasense Core's color cameras. Due to the narrowness of the final course, most of the explored sections resulted to be within a close range from the color cameras. Therefore the inspection payload executed only single point scans during the Prize Round.
Overall, \num{25} scan points (cf. section~\ref{sec:inspection_payload_autonomous_scanning}) for ANYmal \num{1} and \num{156} for ANYmal \num{2} were targeted. ANYmal \num{2} was inside wider areas than ANYmal \num{1}, explaining the higher number of scans. Scans in narrow passages were either due to imperfect ray-casting in narrow and tangled sections or because the inspection payload captured the area prior to the Alphasense Core. For each scanned point, the camera images obtained with the zoom camera were processed by another instance of the same neural network used for the Alphasense Core's camera. The thermal image samples were processed by the thermal detector at \SI{3}{Hz}. Figure~\ref{fig:inspection_payload_results} shows an example of multi-modal artifact detections using the ANYmal's inspection payload module.

\begin{figure}[h]
    \centering
    \includegraphics[width=0.9\textwidth]{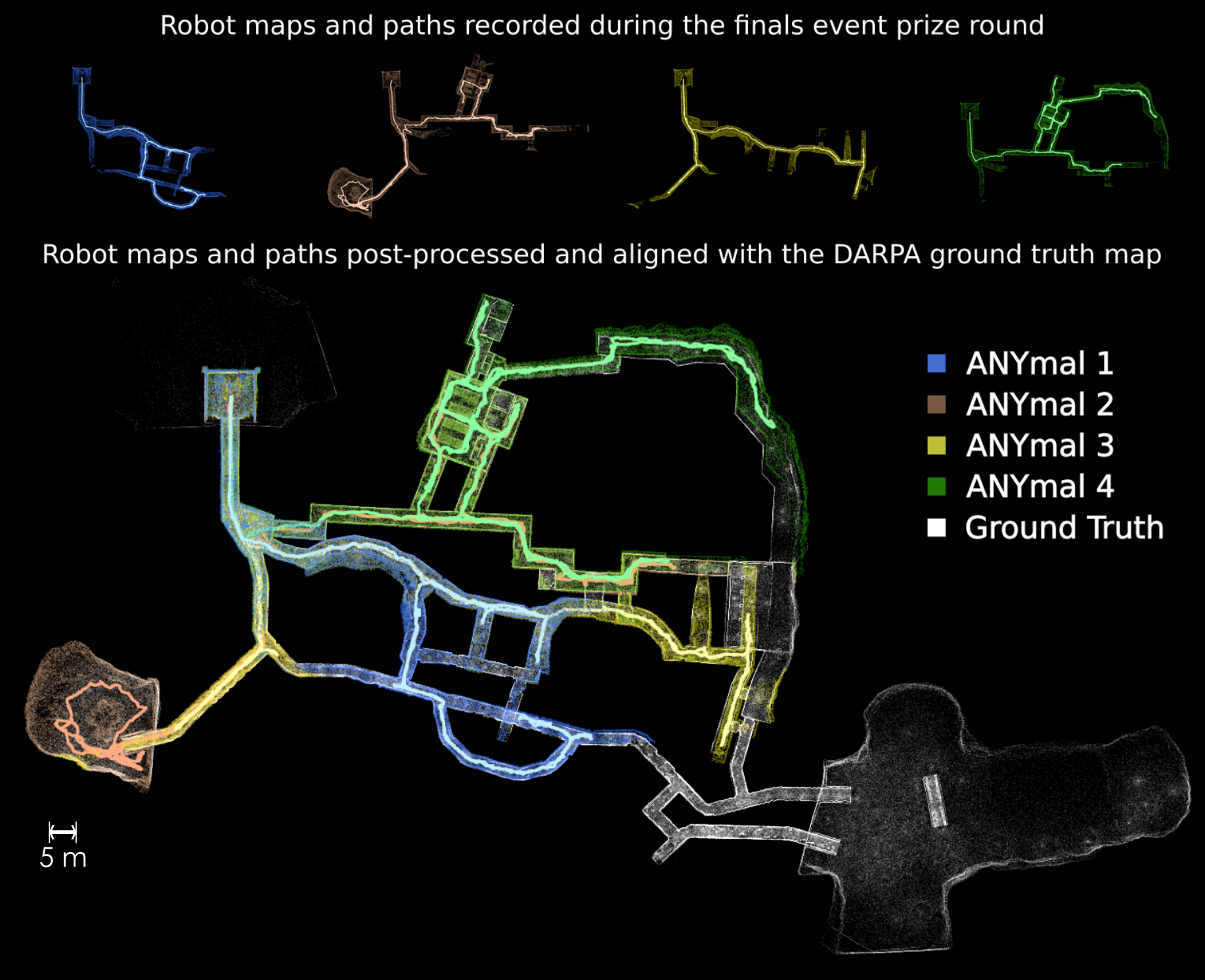}
    \caption{Robot paths and maps generated by CompSLAM and recorded during the Prize Round are shown (top row). The paths and maps were post-processed and aligned with the DARPA provided ground-truth map post competition to provide a qualitative overview of the mission performance (bottom). A video showing the Team's Prize Round is available at~\url{https://youtu.be/QON8IFc8cjE}}
    \label{fig:compslam_results_overlay}
\end{figure}

\textbf{Evaluation of the multi-modal and multi-robot perception system} \\
The onboard complementary multi-modal localization and mapping, CompSLAM, and the global multi-robot mapping approach, M3RM, were deployed on all the legged robots during the Prize Round.
Each robot individually estimated its state and the map of the environment and, when in communication range, transferred all its information, including submaps, back to the Base Station. 
However, during the Prize Round the M3RM approach had significant difficulties due to the texture-less entrance of the urban part of the competition course as the primary modality of M3RM was visual imagery. 
Consequently, for M3RM the majority of the visual constraints in this particular area were incorrect which lead to skewed maps. 
Hence, M3RM required a significantly larger amount of time to converge to the correct solution. Furthermore, the computational burden prohibited its reliable utilization after a certain point during the duration of the Prize Round. Therefore, the Human Supervisor mainly relied on the onboard CompSLAM pose estimates for the mapping and artifact reporting. Figure~\ref{fig:compslam_results_overlay} shows the maps and paths from CompSLAM for all robots as recorded during the Prize Round (top row). Furthermore, robot map and paths post-processed and aligned  with the DARPA-provided ground-truth map are also shown in the bottom portion of Figure~\ref{fig:compslam_results_overlay} to provide a qualitative measure of performance.

To give a quantitative evaluation of the onboard localization during the mission as well as the onboard and global mapping performance that could have been achieved either given different tuning parameters or more time (especially for M3RM), both CompSLAM and M3RM were post-processed and are compared with ground-truth robot trajectories.
The ground-truth robot trajectories were created by registering individual robot pointclouds to the DARPA-provided ground-truth map, as described in~\cite{ramezani2020newer}. %
Table~\ref{tab:prize_run_evaluation_rpe} presents the Relative Pose Error (RPE) metric for the CompSLAM (onboard) localization during the run and post-processed by comparing against individual ground-truth trajectories. During the competition, as the scale of environment was unknown, the down-sampling leaf-size of pointclouds was increased from \SI{0.1}{\meter} to \SI{0.2}{\meter} as a safeguard against excessive onboard computational demand, which later given the environment size, was deemed unnecessary. To measure the performance difference, the down-sampling leaf-size was restored for the post-processed results which increased the accuracy of the onboard approach, as demonstrated in Table ~\ref{tab:prize_run_evaluation_rpe}. Since the global multi-robot mapping approach, M3RM, sparsifies the robot poses by using key-frames as compared to estimating the full robot trajectory as done by CompSLAM, the RPE for M3RM cannot be calculated accurately and hence is omitted from comparison.

Similarly, Table~\ref{tab:prize_run_evaluation_ape} contains the Absolute Pose Error (APE) metric for the CompSLAM approach, during the run and post-processed as well as the M3RM (global) mapping approach post-processed.
Combining the information from all individual robots leads to obtaining more constraints and hence better constraining the solution and building a more extensive factor graph.
Despite the bigger optimization problem with more information, the M3RM solution was able to reduce the onboard drift due to intra- and inter-robot loop closures.
In particular, the upward-facing cameras of the ANYmal's sensor rig provided many valuable loop closure detections, especially in passages that were traversed in different directions.
As a result, for all final ANYmal runs, the M3RM approach improved the trajectory estimates of the onboard localization and mapping solution and significantly corrected the poses in terms of the APE.
\begin{table}[h]
\centering
\begin{tabular}{c|cc|cc}
\toprule
\multicolumn{5}{c}{\textbf{Prize Round Evaluation - mean RPE and standard deviation}} \\
\midrule
\textbf{Robot} & \multicolumn{2}{c|}{\textbf{During Run}} & \multicolumn{2}{c}{\textbf{Post-Processed}} \\
\midrule
& \multicolumn{2}{c|}{\textbf{CompSLAM}} & \multicolumn{2}{c}{\textbf{CompSLAM}} \\
& Rotation\,[$\degree$] & Translation\,[\si{\metre}] & Rotation\,[$\degree$] & Translation\,[\si{\metre}] \\
ANYmal 1 & 1.21\,(0.76) & 0.07\,(0.05) & \textbf{0.69}\,(0.36) & \textbf{0.03}\,(0.02)\\
ANYmal 2 & 0.85\,(0.54) & 0.05\,(0.04) & \textbf{0.54}\,(0.34) & \textbf{0.04}\,(0.03) \\
ANYmal 3 & 0.73\,(0.41) & 0.05\,(0.03) & \textbf{0.40}\,(0.20) & \textbf{0.04}\,(0.02) \\
ANYmal 4 & 0.89\,(0.61) & 0.06\,(0.04) & \textbf{0.56}\,(0.31) & \textbf{0.04}\,(0.03) \\
\bottomrule
\end{tabular}
\caption{Comparison of the mean Relative Pose Error (RPE) from the DARPA SubT challenge Final Event of the CompSLAM approach during the run and post-processed across all the deployed robots.}
\label{tab:prize_run_evaluation_rpe}
\end{table}

\begin{table}[h]
\centering
\setlength{\tabcolsep}{5pt}
\begin{tabular}{c|cc|cc|cc}
\toprule
\multicolumn{7}{c}{\textbf{Prize Round Evaluation - mean APE and standard deviation}} \\
\midrule
\textbf{Robot} & \multicolumn{2}{c|}{\textbf{During Run}} & \multicolumn{4}{c}{\textbf{Post-Processed}} \\
\midrule
& \multicolumn{2}{c|}{\textbf{CompSLAM}} & \multicolumn{2}{c|}{\textbf{CompSLAM}} & \multicolumn{2}{c}{\textbf{M3RM}}\\
& Rotation\,[$\degree$] & Translation\,[\si{\metre}] & Rotation\,[$\degree$] & Translation\,[\si{\metre}] & Rotation\,[$\degree$] & Translation\,[\si{\metre}]\\
ANYmal 1 & 3.71\,(2.02) & 0.60\,(0.37) & 2.45\,(0.67) & 0.72\,(0.41) & \textbf{1.59}\,(0.46) & \textbf{0.25}\,(0.13) \\
ANYmal 2 & 4.55\,(1.07) & 1.81\,(0.84) & 3.97\,(0.40) & 1.29\,(0.90) & \textbf{0.96}\,(0.30) & \textbf{0.36}\,(0.28) \\
ANYmal 3 & 1.53\,(0.68) & 0.22\,(0.29) & \textbf{0.89}\,(0.50) & 0.23\,(0.43) & 2.30\,(1.02) & \textbf{0.20}\,(0.34) \\
ANYmal 4 & 3.26\,(1.02) & 1.24\,(0.67) & 2.22\,(0.79) & 1.00\,(0.71) & \textbf{2.16}\,(0.55) & \textbf{0.24}\,(0.17) \\
\bottomrule
\end{tabular}
\caption{Comparison of the mean Absolute Pose Error (APE) from the DARPA SubT challenge Final Event between CompSLAM (during the run and post-processed) and the global M3RM (post-processed only) localization and mapping approaches.}
\label{tab:prize_run_evaluation_ape}
\end{table}

\subsubsection{Finals discussion}
In this section, we review selected highlights of individual components of our system as well as core challenges that were faced during the Final Event of the DARPA SubT Challenge.

\begin{figure}[]
    \centering
    \begin{subfigure}{.99\linewidth}
      \centering
      \includegraphics[width=0.99\textwidth]{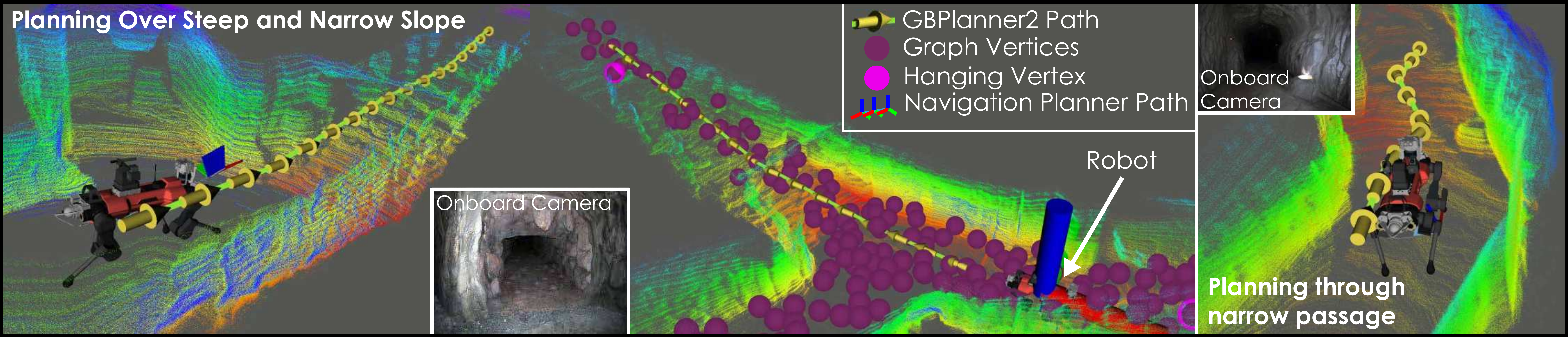}
      \caption{Cave Section}
      \label{fig:gbplanner_results_cave}
    \end{subfigure}
    \begin{subfigure}{.99\linewidth}
      \centering
      \includegraphics[width=0.99\textwidth]{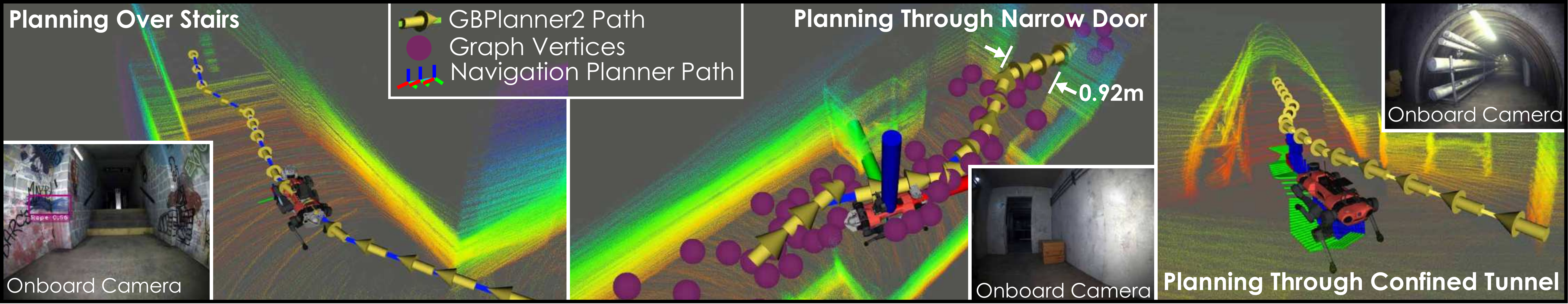}
      \caption{Urban Section}
      \label{fig:gbplanner_results_urban}
    \end{subfigure}
    \begin{subfigure}{.66\textwidth}
      \includegraphics[width=\textwidth]{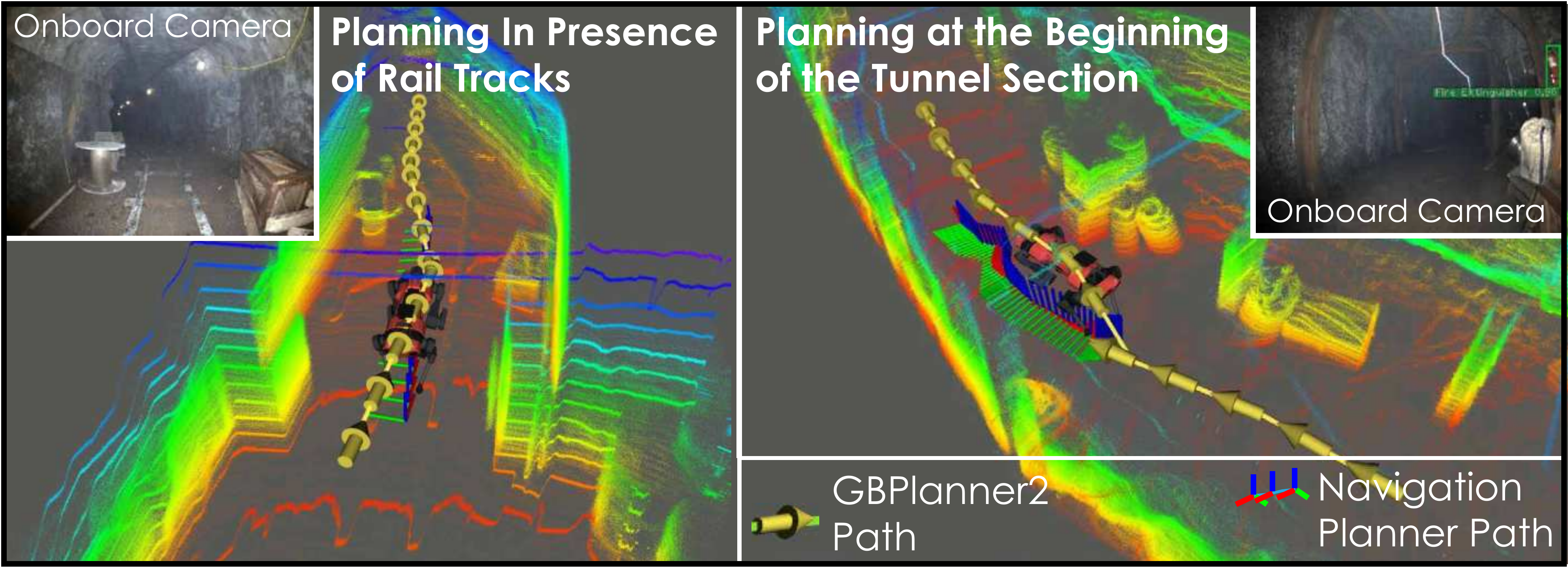}
      \caption{Tunnel Section}
      \label{fig:gbplanner_results_tunnel}
    \end{subfigure}
    \begin{subfigure}{.305\textwidth}
      \includegraphics[width=\textwidth]{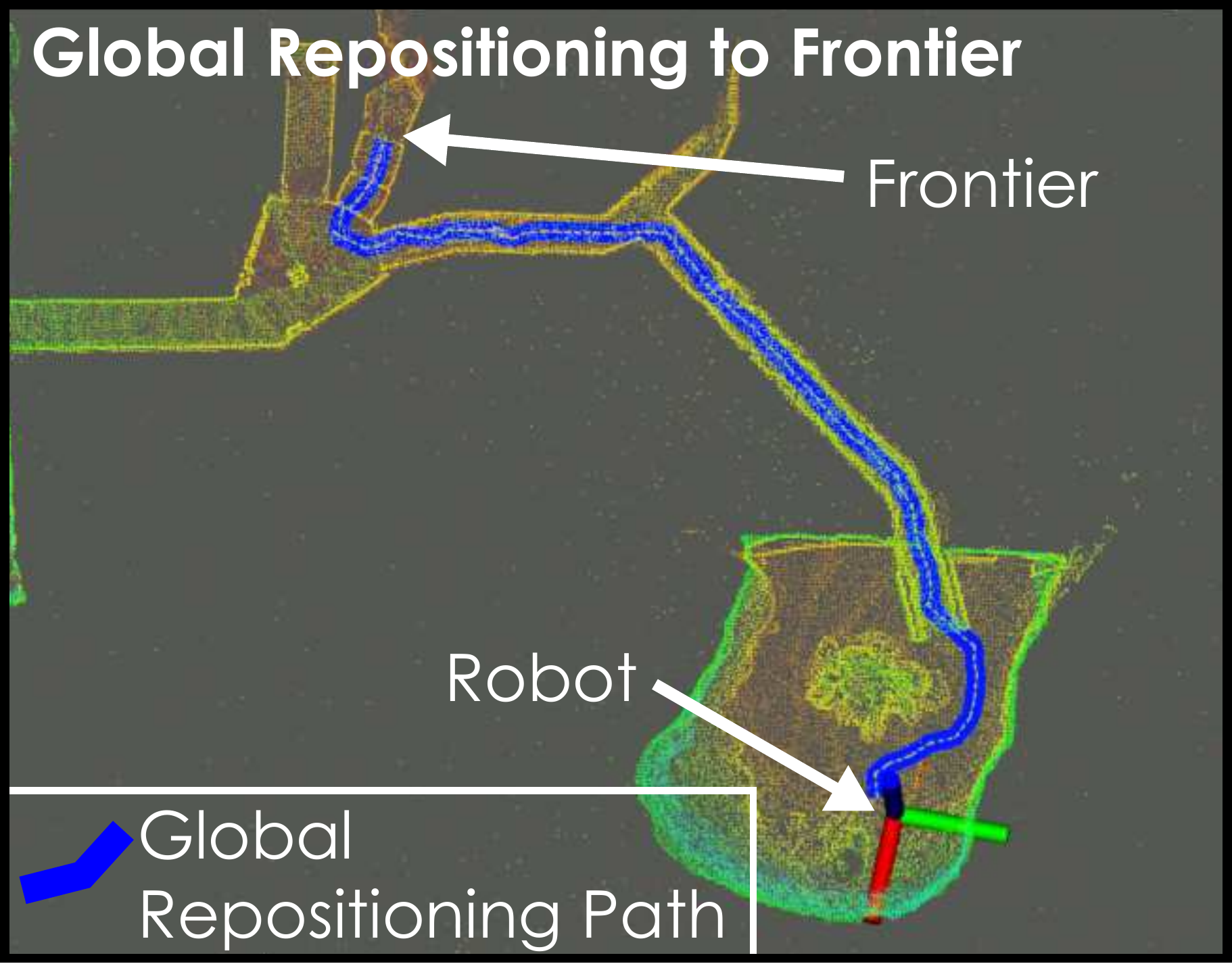}
      \caption{Global Repositioning}
      \label{fig:gbplanner_results_global}
    \end{subfigure}
    
    \caption{Highlights of the deployment of the exploration planner in the Prize Round of the Final Event. (\subref{fig:gbplanner_results_cave}), (\subref{fig:gbplanner_results_urban}), and (\subref{fig:gbplanner_results_tunnel}) show instances of local exploration in the Cave, Urban, and Tunnel sections of the course respectively. The figure shows the performance of the exploration planner in the presence of short stairs, steep slopes, narrow openings and passages. It is noted that in (\subref{fig:gbplanner_results_urban}), even though the exploration planner planned a path up the staircase, ANYmal did not walk up the stairs due to conservative tuning of the navigation planner. The path commanded by GBPlanner2, refined traversable path by the navigation planner, and the local exploration graph (edges are not shown for better visibility) can be seen along with the front camera view of the robot. (\subref{fig:gbplanner_results_global}) shows an instance of global repositioning. The robot had finished exploring one part of the Cave section and the Human Supervisor provided a goal point at the beginning of the Tunnel section. GBPlanner2 calculated a path towards the goal along the global graph.}
    \label{fig:gbplanner_results}
\end{figure}

\textbf{Exploration planner highlights}\\
The Final Event course of the SubT challenge presented settings, particularly challenging for autonomy. These include stairs, steep slopes, narrow openings, confined passages and big open rooms, and an overall large-scale of the environment. The Graph-based exploration planner was able to provide safe and efficient exploration paths in real-time. The improvements in GBPlanner2 for the ground robots enabled the planner to plan over short staircases and steep slopes. The team had done certain tuning and modification in the planner a few weeks before the competition to enhance its ability to plan through extremely narrow (\SI{1}{\meter} x \SI{1}{\meter} cross-section) openings. This paid off in the Final Event as the course involved several narrow, constrained passages which would have been difficult for the exploration planner. A few notable instances of GBPlanner2 are shown in Figure~\ref{fig:gbplanner_results}. The figure shows the paths provided by GBPlanner2 in the presence of the challenges mentioned above in the Cave, Urban, and Tunnel sections of the Final Event course during the Prize Round. The figure also shows an instance of global repositioning.

\begin{figure}[]
    \centering
    \begin{subfigure}{.99\linewidth}
      \centering
      \includegraphics[width=0.99\textwidth]{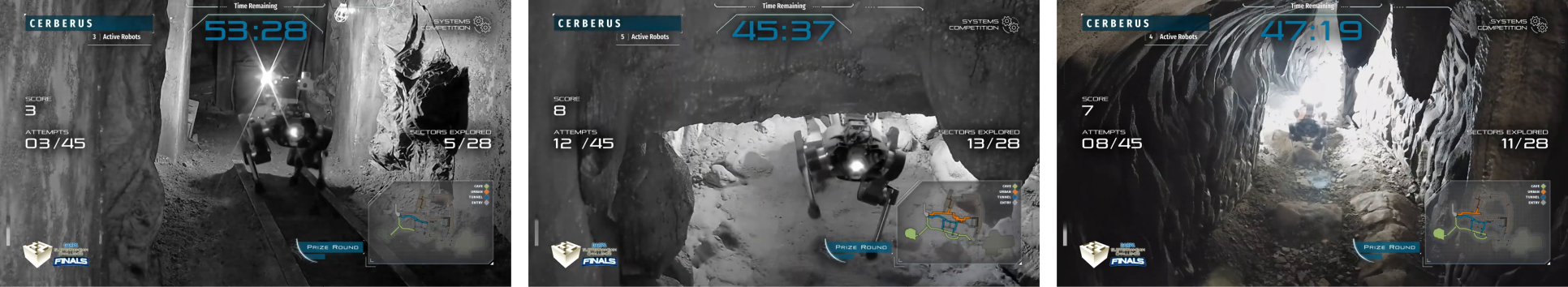}
      \caption{Locomotion challenges}
      \label{fig:anymal_mobility_locomotion}
    \end{subfigure}
    \begin{subfigure}{.99\linewidth}
      \centering
      \includegraphics[width=0.99\textwidth]{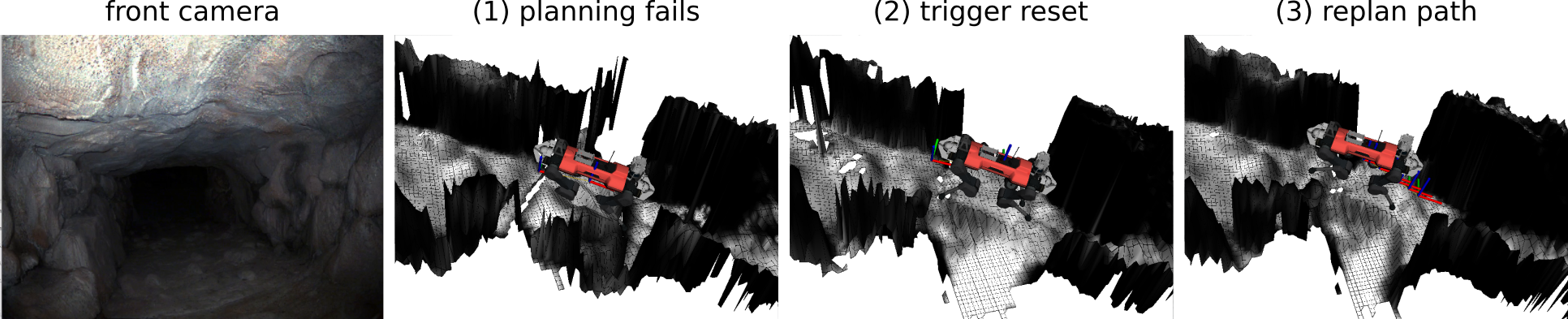}
      \caption{Automatic reset of the elevation map}
      \label{fig:anymal_mobility_reset}
    \end{subfigure}
    \begin{subfigure}{.99\textwidth}
      \centering
      \includegraphics[width=0.99\textwidth]{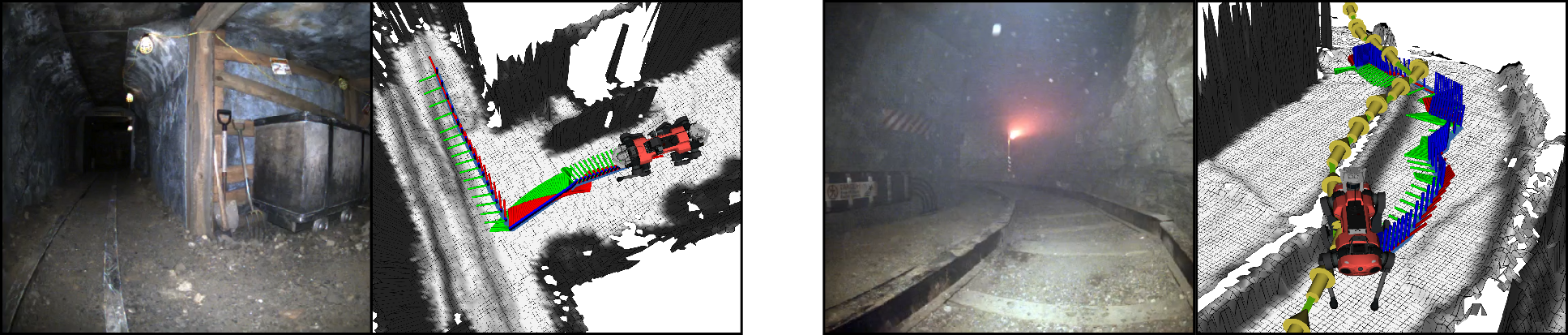}
      \caption{ANYmal navigation planner instances}
      \label{fig:anymal_mobility_navigation}
    \end{subfigure}
    \caption{Highlights of ANYmal's mobility in fully autonomous mode during the Prize Round. (\subref{fig:anymal_mobility_locomotion}) Locomotion challenges included obstacles, rough surfaces, and railway tracks (left); slopes and constrained passages (center); rough terrain (right). (\subref{fig:anymal_mobility_reset}) Automatic safety mechanisms acted during the mission to ensure the robot could deal with unexpected situations, for example, with challenging localization conditions leading to corrupted height maps, like in the case shown in part (1), where drift in the robot's odometry made the elevation map move upwards up, touching the agent's knees, thus preventing the ANYmal navigation planner from finding a traversable path due to an invalid start position. If such a condition was detected, the terrain map was automatically reset and a new planning iteration triggered. (\subref{fig:anymal_mobility_navigation}) The learned motion cost within the ANYmal navigation planner allowed us to circumnavigate obstacles like high rails, instead of crossing them. Poses represented as RGB axes were produced by the navigation planner, whereas poses displayed as yellow arrows were created by the exploration planner. This also led to safer behavior when the robot had to walk on rails, turning before stepping onto the rails to then walk straight.}
    \label{fig:anymal_mobility_results}
\end{figure}

\begin{figure}[]
    \centering
    \includegraphics[width=0.7\textwidth]{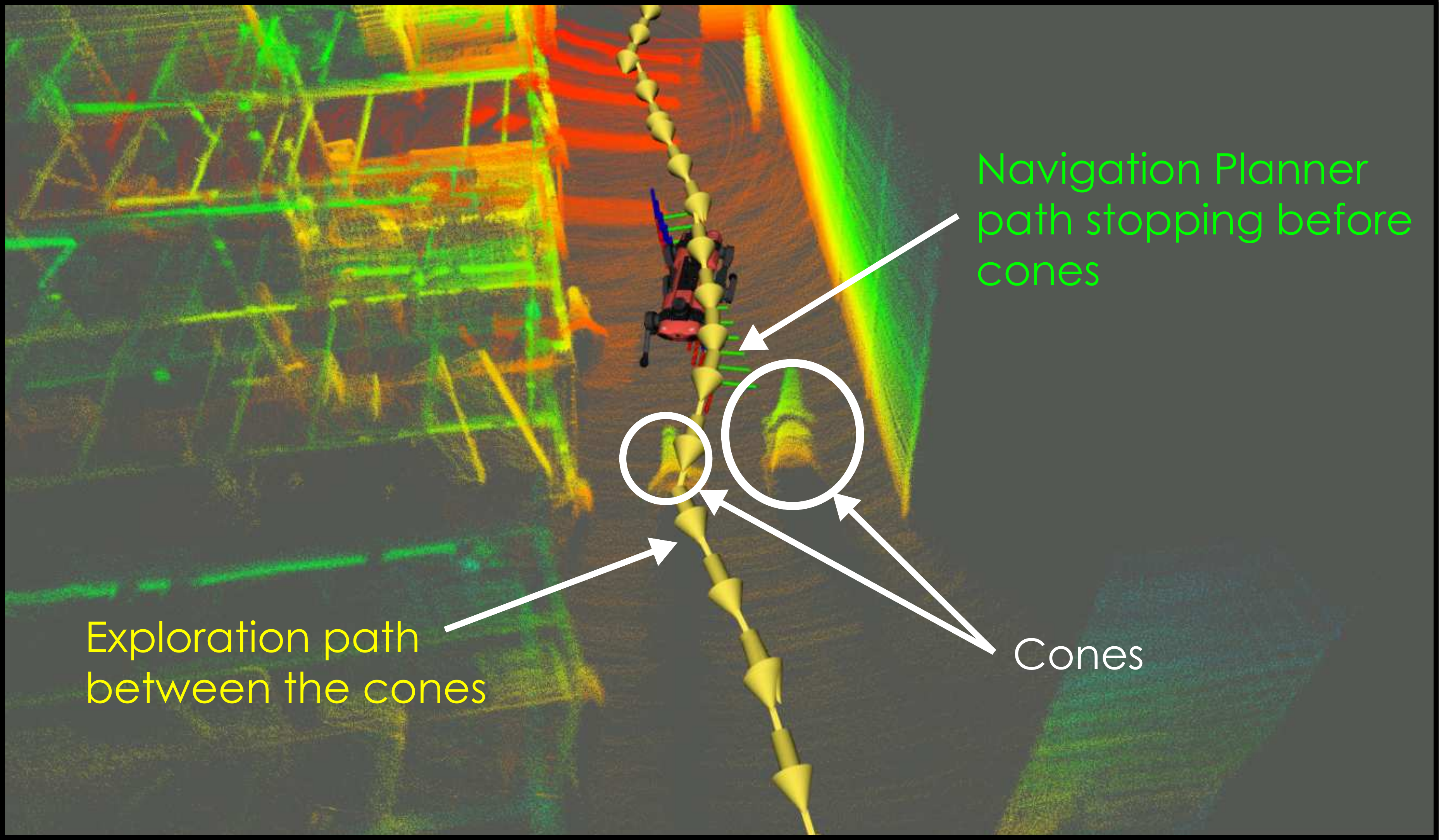}
    \caption{An instance where ANYmal \num{4} encountered a non-traversable narrow gap formed by two construction cones. The exploration planner attempted to plan through the gap, but the navigation planner prevented the robot from going through it as it was not safe for the robot to traverse through it. This shows the usefulness of the synergy between the two planners, but also highlights the lack of feedback from the navigation planner to the exploration planner.}
    \label{fig:planning_challenge_cones}
\end{figure}

\textbf{ANYmal mobility highlights}\\
The combination of our perceptive locomotion controller and the navigation planner allowed the four legged robots to safely walk in the most challenging parts of the course (see Figure~\ref{fig:anymal_mobility_locomotion}). Moreover, the navigation planner was able to stop the robot from hitting obstacles which GBPlanner2 missed, as shown in Figure~\ref{fig:planning_challenge_cones}.
In conditions challenging for the localization and mapping pipeline, the height map can become corrupted due to unstable or too imprecise pose estimates. Corrupted terrain maps could lead to a complete failure to plan for the ANYmal navigation planner.
If the robot pose was considered invalid with respect to the terrain map (which should be impossible), the ANYmal's behavior tree module triggered a reset of the height map.
This allowed us to recover from mapping failure, as shown in Figure~\ref{fig:anymal_mobility_reset}.
Furthermore, the use of learned motion costs showed to be valuable in producing safer, more efficient paths.
Figure~\ref{fig:anymal_mobility_navigation} shows how the navigation planner prefers to circumnavigate high railroad tracks instead of crossing them.
In cases where crossing rails is necessary the robot rotates before stepping onto the rails such that it can walk straight on them, with the rail between its legs.

\textbf{Operator's interface challenges}\\
The usage the \ac{GUI}s, as presented in section~\ref{sec:single_human_supervisor}, significantly improved the capabilities of the Human Supervisor to manage the overall mission and interact with the single agents, compared to the Team's setup at the previous SubT events. Despite the improvements achieved by having only two user interfaces, the context switching of selecting a different robot in the \ac{MCI} was the principal Human Supervisor's limitation. Whenever he chose a new robot in this interface, for example to task the agent with a different exploration goal, he often had first to query the robot for additional information. Specifically, every time a new robot was selected, the supervisor first had to request the robot's information, such as its local CompSLAM map or specific sensors data, wait for the agent to stream them and then decide on a new exploration goal for the agent. These data were robot specific and had to be requested for every selected robot. In the case of bandwidth-demanding data streamed over a wireless connection shared with several agents, the whole process could take up to several seconds to complete. Therefore, selecting a new robot to control included a fixed context switching, to be paid in terms of no-operation time for the supervisor, limiting how quickly he could supervise different agents. This constraint, combined with the request to check for artifacts reports and overall mission status in the \ac{MARI}, overloaded the Human Supervisor. A single \ac{GUI} to both control a selected robot as well as display an overview of the mission alongside the artifacts reports would have been beneficial and released some of the cognitive load from the Human Supervisor.   

\textbf{Communication challenges}\\
The combination of a tethered rover and deployable wireless breadcrumb modules helped the team's overall performance during the mission execution by extending the wireless network's reach. Despite the development done to create ruggedized and reliable breadcrumb nodes, we deployed only three of them at the Final Round (given the eight available modules, four per Carrier robot). We identified two main reasons for using just a few modules: the increased level and reliability of the autonomy of the agents and the cognitive overload of the Human Supervisor. The former implied that the supervisor could reliably task each agent to autonomously explore an assigned area, given a time budget, and wait for the robot to come back into communication range to then receive the artifact reports. While the robot was out of communication range, the supervisor could focus on other important matters, such as coordinating the other robots or analyzing other artifacts reports. These important matters are also the cause of the cognitive load on the Human Supervisor, that couldn't really focus on if and where to deploy the communication breadcrumbs of each Carrier robot. Deploying more breadcrumb modules could have helped the Human Supervisor stay connected with the robots even when further away from the Base Station, thus allowing to send artifacts reports sooner. This could have been possible only if an automatic mechanism to decide when to deploy a breadcrumb node had been implemented for the Carrier robot.

\begin{figure}[]
    \centering
    \includegraphics[width=0.9\textwidth]{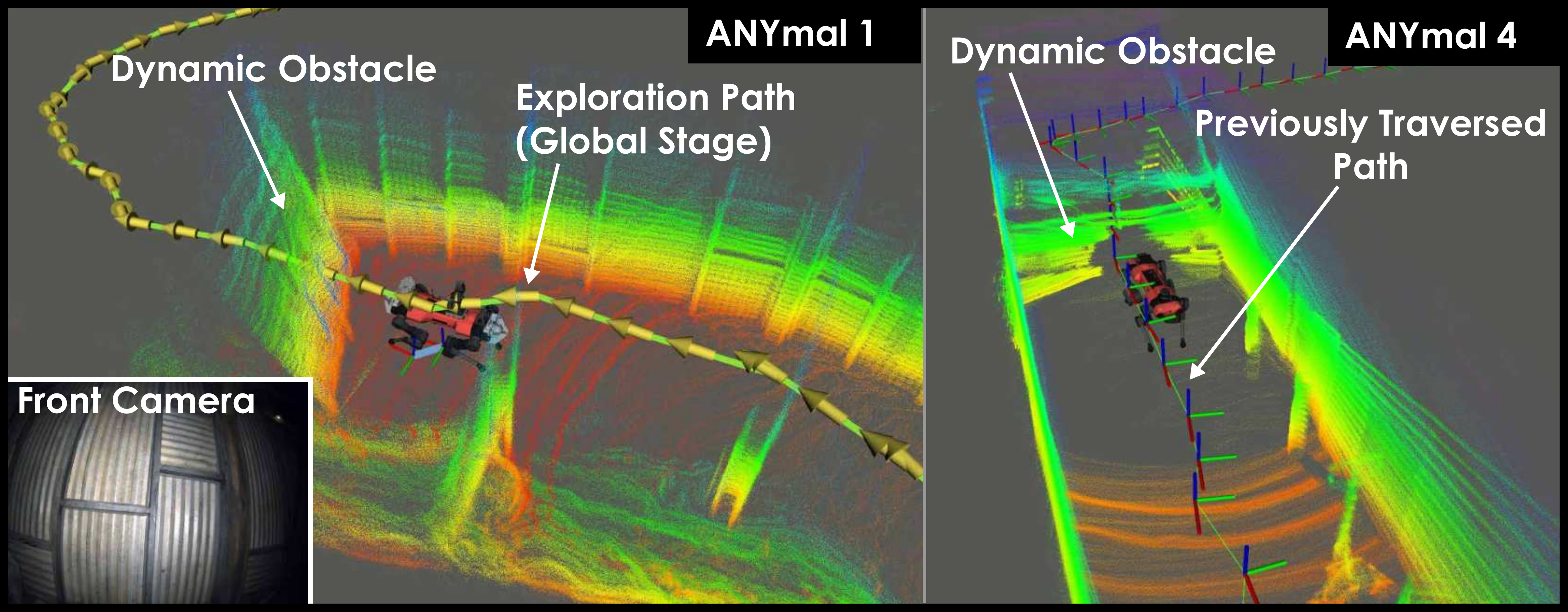}
    \caption{ANYmal \num{1} and ANYmal \num{4} encountering dynamic obstacles in the Tunnel and Urban section respectively. The dynamic obstacle blocked the path of ANYmal \num{1} while performing global repositioning. It attempted to follow the path for some time but could not move due to the presence of the obstacle. Therefore after some time, the watchdog module of ANYmal's behavior tree triggered and it started backtracking the previously traversed path. However, due to the obstacle blocking the path ANYmal \num{1} initially took, the robot was stuck for the remaining of the round. This was due to two factors happening at the same time: the backtracking of the path command can be stopped only by the Human Supervisor, but in that section of the course, there was no WiFi connection with the Base Station therefore the supervisor did not realize the robot was still functioning but stuck in an obstacle. ANYmal \num{4} was backtracking its traversed path to the home location when it encountered the dynamic obstacle. The robot was stuck for some duration after which the local layer of GBPlanner2 was manually triggered by the Human Supervisor.}
    \label{fig:planning_challenge_dobs}
\end{figure}

\textbf{Planning challenges}\\
The Final Event of the DARPA SubT Challenge included dynamic obstacles blocking certain parts of the course. The team faced challenges due to these dynamic obstacles. The global re-positioning and autonomous homing (through GBPlanner2 or backtracking of the traversed path) parts of the autonomy stack were not designed for handling dynamic obstacles. In the Prize Round two of the robots encountered the dynamic obstacles, one in the Tunnel section and the other in the Urban section. While ANYmal \num{1} was performing global repositioning, the path was blocked by a dynamic obstacle. It attempted to follow the path for some time after which the watchdog module of ANYmal's behavior tree triggered backtracking of the previously traversed path. However due to the obstacle ANYmal \num{1} was stuck for the remaining of the round. This was due to two factors happening at the same time: the backtracking of the path command can be stopped only by the Human Supervisor but in that section of the course there was no WiFi connection with the Base Station, therefore the supervisor did not realize the robot was still functioning but stuck in an obstacle. Similarly, ANYmal \num{4} encountered a dynamic obstacle while backtracking the traversed path to the home location. ANYmal \num{4} was stuck in a similar fashion as ANYmal \num{1}, but still in connection range. Therefore, the Human Supervisor could stop the recovery behavior and trigger GBPlanner2 again to explore another area. Figure~\ref{fig:planning_challenge_dobs} shows both the instances where the robots encountered the dynamic obstacles.

In the Urban section of the final course ANYmal \num{4} encountered two construction cones creating a narrow gap through which the robot could not fit through. However, due to the relaxed tuning of GBPlanner2, it planned through that gap on multiple occasions. The navigation planner correctly prevented the robot from passing through that gap. This demonstrates the usefulness of the synergy between an optimistic exploration planner and a conservative robot navigation planner. However, in our autonomy system, the two planners were loosely coupled. It meant that if the navigation planner detected that the path from the exploration planner could not be tracked, the exploration planner would be re-triggered. There was a lack of feedback from the navigation planner to the exploration planner which caused the exploration planner to plan through that gap multiple times slowing down the mission. Figure~\ref{fig:planning_challenge_cones} shows one instance of this. %

\begin{figure}[]
    \centering
    \includegraphics[width=1.00\textwidth]{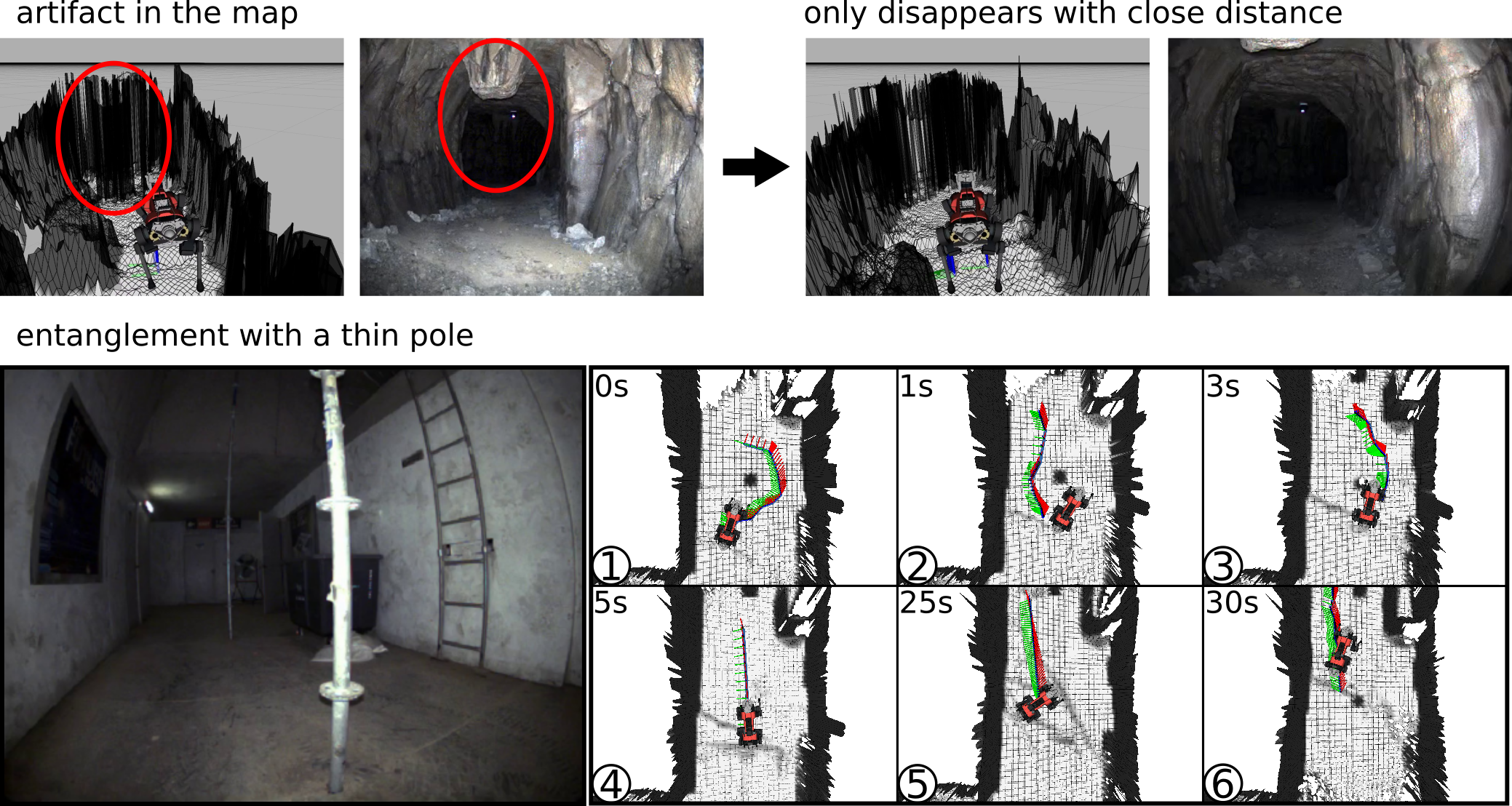}
    \caption{ANYmal navigation challenges. 
    (top) To be able to capture slopes or stairs, the upper limit of depth measurements included into the elevation map was set above robot height. This produced artifacts in the elevation map when low ceiling points are projected into the \num{2.5}D grid. These artifacts were only removed when the robot got closer, thanks to visibility cleanup and rejection based on distance, which slowed down planning.
    (bottom) A \SI{500}{\milli\second} time delay between computing a path and starting to follow it lead to imperfect path following. As a result, one robot briefly got stuck on a narrow scaffolding pole in the urban section.}
    \label{fig:anymal_mobility_challenges}
\end{figure}

In the tight underground environments encountered during the Final Event, the 2.5D height map representation used for planning became problematic.
We used a filter which allows measurements at larger height to be included into the height map, as distance increases. 
It was tuned to work on slopes and stairs, but still allowed us to pass through low openings when the robot was close to them. 
This caused artifacts in the elevation map in areas with low ceiling height, as shown in Figure~\ref{fig:anymal_mobility_challenges}(top).
While these artifacts were removed by a visibility cleanup check using ray-casting as the robot got closer, they still restricted space considered traversable by the planner and therefore slowed progress.

During data analysis after the competition, we discovered that there was a non-constant delay between the navigation planner publishing a path, and the path tracking module which we observed to be up to \SI{500}{\milli\second}.
This, in combination with height map issues and the nonsmooth nature of sampling-based planning, caused ANYmal \num{2} to get stuck on a scaffolding pole in the Urban section for about \SI{20}{\second}.
The time series of this event is shown in Figure~\ref{fig:anymal_mobility_challenges}(bottom):
When the robot approached the pole, the initial plan was to circumnavigate it on the right, which was slightly longer than on the left. 
Just as the robot started moving, a new path was published, which passed the pole on the left.
This caused the robot to move so close to the pole that it could not be perceived and disappeared from the height map.
Since the robot had followed the original path to the right, the next path went right again and got the robot stuck on the pole.
Because the pole was now missing from the height map, all new paths went straight through it.
After another \SI{20}{\second}, the robot had drifted to the left of the pole and finally got unstuck.

\begin{figure}[]
    \centering
    \includegraphics[scale=1]{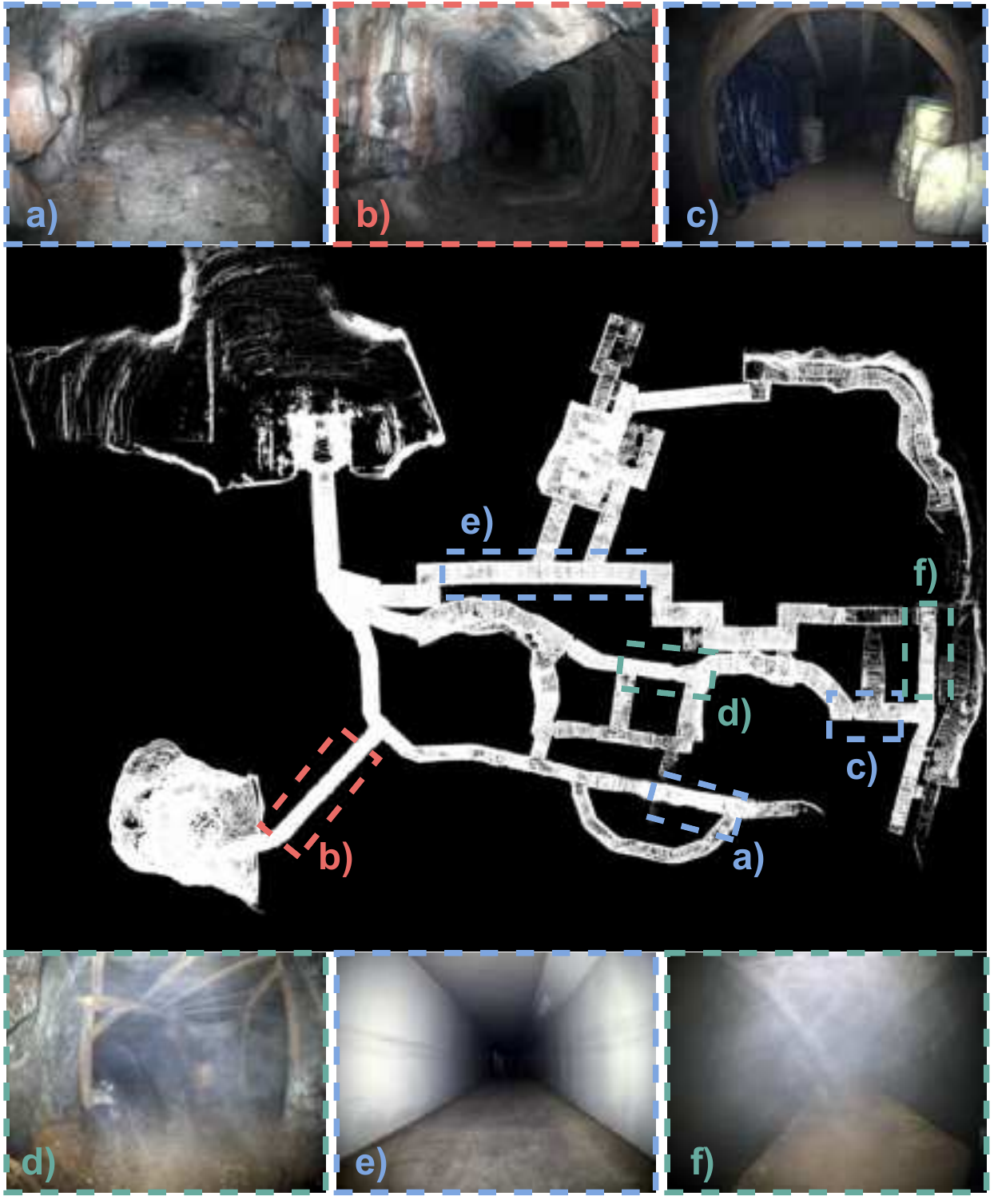}
    \caption{Illustration of a selection of different types of degradation encountered during the DARPA SubT Final Event. Blue denotes completely different kinds of environment; red denotes narrow passages; green areas denote smoke. Displayed images are taken from the onboard cameras of the deployed robots.}
    \label{fig:perception_challenges}
\end{figure}

\textbf{Perception challenges}\\
The course designers of the DARPA's organizational team introduced several challenging conditions to directly and indirectly influence the performance of robot perception. 
Figure~\ref{fig:perception_challenges} summarizes the key challenges faced during the exploration of the course environment. 
Throughout the course, several areas were completely dark and dull. 
Hence, the visual cameras had to rely on the onboard illumination.
The effective range of the onboard LEDs was approximately \SI{7}{\meter} which was enough for close passages such as tunnel and urban systems. 
However, in large open-spaces such as the cave environment it only illuminated the nearby structure and effectively missed regions further away. Such areas posed a challenge for visual odometry estimation as features were only detected tracked in the near vicinity of the camera, typically on the planar ground directly in front of the robot.

Furthermore, the course featured many narrow passages, some with smooth planar surfaces, e.g. the corridor of the urban area. Such areas posed a challenge for the LiDAR odometry, mainly due to two reasons. First, the narrow corridors significantly reduced the observation range of the already sparse VLP16 Puck LITE, making the estimation of ground and ceiling planar surfaces at longer ranger difficult. Second, the smooth planar walls created a featureless self-similar environment along the principal direction of the corridor. Accordingly, this rendered the underlying LiDAR pose estimation optimization briefly ill-conditioned, thus enforcing the use of fallback solutions for the correct robot pose and map estimation.    
Another crucial type for perceptual degradation was the introduction of obscurants such as smoke and dust, as well as dynamic elements, e.g. trap-doors, into the course. 
Different areas in the course featured motion-triggered smoke machines that produced artificial smoke visible in the visual camera and LiDAR sensors.
The artificial smoke introduced many incorrect feature detections for visual tracking and led to the loss of good visual landmarks in the scene.
Additionally, the artificial smoke might lead to incorrect measurements for the LiDAR sensors for which the configuration of only using the strongest return signal per beam is a possible remedy. Especially during the Prize Round, where our robots were exposed to multiple sections filled with smoke, our onboard CompSLAM successfully penetrated such conditions and offered robust estimates. 

Moreover, the trap doors made it particularly difficult for the registration-based approaches, e.g., scan-to-map or the LiDAR loop closures to spatially close poses, since the current pointcloud differs significantly from the previously acquired pointcloud.

As mentioned before, the M3RM performance significantly decreased due to the degraded entrance to the urban environment. 
Generally, the entrance to the urban area featured a long white and texture-less corridor with no salient regions nor structure.
Thus, putting a significant burden on the visual perception pipelines for the M3RM approach as the lack of good visual features and wrong associations lead to incorrect visual landmarks in the factor graph that resulted in erroneous estimation.

Additionally, the narrow passages not only during the entrance of the urban area but also throughout the course yielded many overexposed scenes since the onboard lighting system was not adaptively controlled. 
As a result for M3RM, in many situations, the employed detection and description of BRISK features resulted in only a handful of good landmarks since most of them were directly discarded due to having a too short feature track. 
Similar to the urban entrance, the factor graph then becomes less constrained and more sensitive to the other terms, such as the IMU and especially its biases.

\begin{figure}[]
    \centering
    \includegraphics[scale=1]{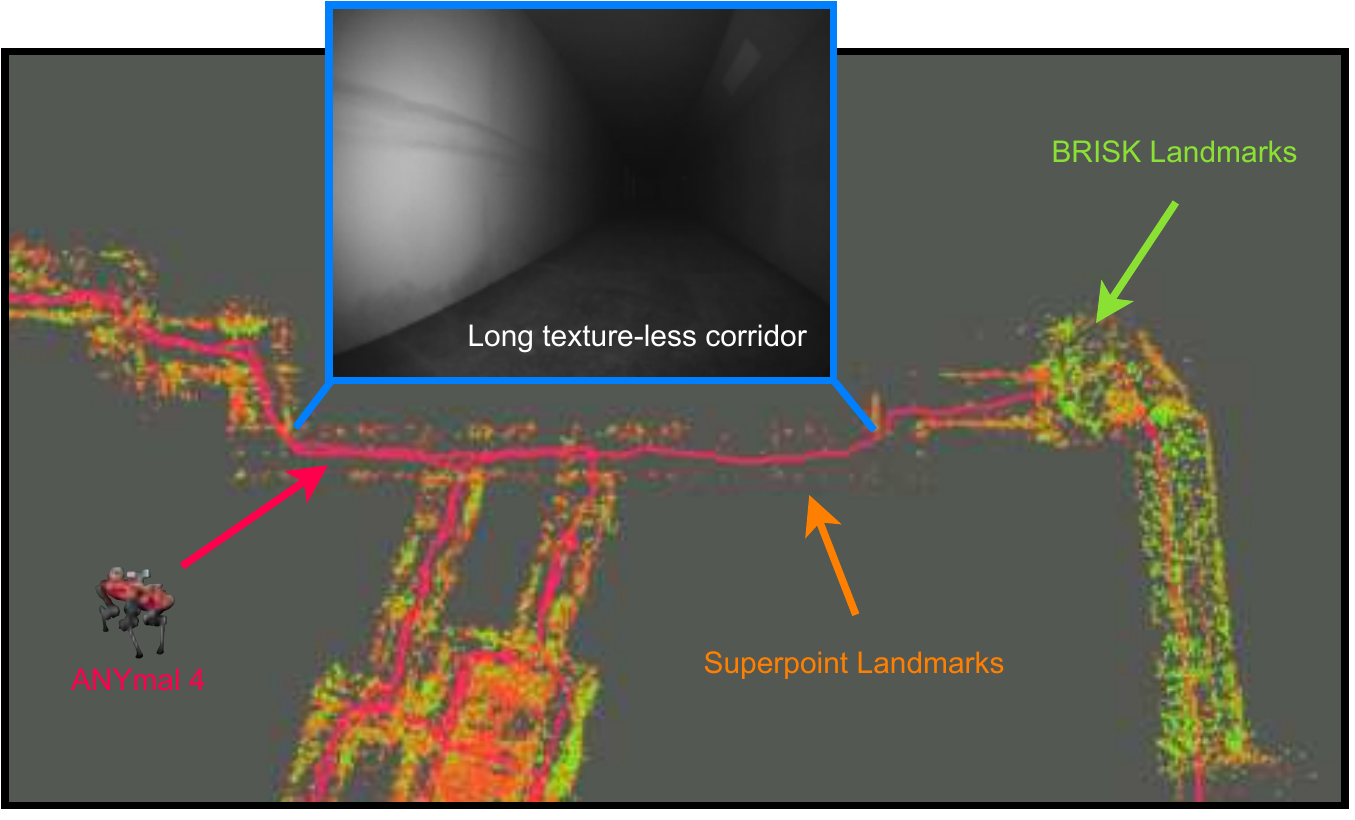}
    \caption{Visual map of the Prize Round of ANYmal \num{4}. The map only contains landmarks with a minimum of three observations. Thus, landmarks detected using BRISK are barely present in the corridor. Despite the lack of structure, Superpoint still detects several good landmarks in this challenging environment.}
    \label{fig:perception_challenge_superpoint}
\end{figure}

It is evident that a keypoint detection and description using BRISK has limited performance and efficiency in subterranean environments. 
Therefore, greatly decreasing the M3RM pipeline as the visual landmarks were primarily used for global loop closure.
An experimental processing of the Prize Round using different feature types such as Superpoint~\cite{detone2018superpoint} showed that the multi-robot map significantly improves.
Figure~\ref{fig:perception_challenge_superpoint} compares two visual maps, one built with BRISK and one with Superpoint features. 

%% file: 05_lessons_learnt.tex
During the three years of the SubT Challenge, our Team gathered a considerable amount of information and insights concerning how our research and development efforts toward efficient subterranean exploration should progress. The main point is that subterranean environments present extreme diversities and pose multiple challenges for robotic autonomy. Our approach toward the exploration of these environments involved, since the beginning of the completion, one unifying, versatile, and scalable solution across all types of underground environments. This section provides overall conclusions for lessons learned that we hope can inform other teams, research groups, technology developers, system operators, and stakeholders that share similar goals and interests.

\subsection{Lessons learned regarding the robotic technologies}

\textbf{Legged and flying robots:} Dexterous and versatile robotic mobility is the key enabling factor to exploring and searching such environments. In our experience, this can be achieved in a robotic systems-of-systems fashion involving multiple legged and flying robots that complement their skills to conquer any terrain and navigate any environment. Quadruped systems offer a versatile and unifying way to negotiate diverse terrain challenges, including rough rocky terrain, large drops/climbs, inclines, steps, stairs, mud, sand, and/or water. They can outperform similar sized tracked or wheeled vehicles in terms of mobility and represent, in our experience, the main robot category to be used when long-term and large-scale (subterranean) exploration is required. Simultaneously, flying systems proved, through our years of experience in the challenge, their significance as the only platform that can naturally be unaffected by the complexities of terrain and navigate fully vertical passages. However, aerial systems typically suffer from endurance and payload limitations, and thus should be used alongside ground robots. We believe that continuing research in legged and flying systems-of-systems, with enhanced walking dexterity, improved flying agility, and the possibility of marsupial integration of ground and flying systems, is the key to conquering the diversity of underground environments.

\textbf{Resilient multi-modal perception is the core enabling factor:} Despite historic progress in the field of SLAM, the subterranean environments presented a number of challenging conditions that significantly degraded robot perception. Consequently despite an overall resilient performance - primarily based on our onboard localization and mapping - certain corner cases and points of failure were identified. 
Systems had to be made resilient in their ability to enable operation even through the most degraded perceptual conditions. Multi-modal robotic perception fusing complementary sensor data capable of penetrating diverse conditions of perceptual-degradation, and presenting different cases of ill-conditioning (e.g., LiDAR data are not affected by lack of visual texture which challenges visible-light cameras but it can be challenged in conditions of geometric similarity which is irrelevant to visual systems), was essential in order to minimize possible breaking points in the robots' ability to localize and map. However, our developed solutions faced a set of challenges. CompSLAM, running onboard our robots, employed a hierarchical sensor fusion architecture which entails that we had to define a set of decision functions and heuristics. Despite its satisfactory performance for the purposes of the SubT Challenge, this approach may present limitations in terms of generalizability. Furthermore, our multi-robot pipeline (M3RM) was severely challenged both in terms of computational requirements and with respect to the resilience of its solution against conditions of visual degradation. 

\textbf{Resilient autonomy technologies are key to explore and search subterranean environments:} Subterranean exploration requires high-degrees of autonomy and the means to control a large team of robots through intuitive high-level commands. The team believed that every robot should be capable of resilient individual autonomy without human teleoperation or without the need to communicate with other robots or the Base Station. We have demonstrated that an exploration planning policy reflecting the key geometric challenges of underground environments could uniformly guide vastly different robotic configurations to explore and search efficiently. The ``Supervised Autonomy'' paradigm (as described in section~\ref{sec:single_human_supervisor}) proved to be effective for such an application. However, some key aspects require significant further improvements. First, for our team only single-robot autonomy was deployed despite multi-robot coordination methods being developed~\cite{GBPLANNER2COHORT_ICRA_2022}. We believe that significantly more field testing experience and accordingly algorithmic refinements are necessary in this direction. Simultaneously, the type of autonomy we deployed did not explicitly exploit semantics (e.g., explicit detection of- and navigation through-staircases or doors) in the environment and did not present a comprehensive modeling of navigation risks (e.g., the interplay between perception conditions and path planning decisions), while a further and more tight integration of traversability constraints at the exploration planning level would have possibly proved beneficial.

\textbf{Testing multi-robot mapping for resilience:}
Deployment tests of multi-robot mapping approaches, such as M3RM, is difficult as it depends heavily on the downstream tasks such as locomotion, communication, and the general state of the robot. Thus, isolated multi-robot testing for scalability and degeneracy issues would yield more information on its weaknesses without having the full system-of-systems running. 
Moreover, tuning mapping systems gets harder the more robots and environments are involved.
A better generic applicability requires less tuning-dependent solutions with fewer parameters and/or self-tuning or adapting thresholds, eventually leading to an environment agnostic real world mapping approach.
It is insufficient to handle failures locally in individual components of the whole pipeline.
Failures need to be considered and treated holistically when, e.g., computing the map or trajectories.
This obviously increases the amount of investigation but greatly increases the overall efficiency of the team. 

\textbf{Advances on legged robotics:} In the course of the SubT challenge, we could profit from rapid progress in the field of legged robots. Our team relied on the ANYbotics ``ANYmal B300'' platform during the first two years as its main ground robot. Despite this hardware solution featuring a good level of robustness and reliability, we experienced a general lack of performance consistency across multiple missions, limited payload capabilities, and insufficient mobility to handle the diverse and challenging terrain of the SubT Challenge. Switching to the ANYmal C quadruped series provided our team with the advantages of a new, stronger, larger, and more dexterous robot compared to the previous B-type platform. The ANYmal C-series was a mature, robust, and reliable hardware solution that allowed us to focus on developing specialized software modules, such as the learning-based perceptive locomotion controller and the navigation planner, that fully leveraged the capabilities of the new robot type and boosted its mobility and robustness. For example, in two years we could increase our locomotion speed from \SI{0.3}{\meter\per\second} (Tunnel Circuit) to \SI{0.45}{\meter\per\second} (Urban Circuit), up to \SI{0.7}{\meter\per\second} (Final Event, where the speed was kept at this level on purpose even though the robot could locomote faster than \SI{1}{\meter\per\second}). In the Final Event, the four ANYmal platforms could operate without a single fall, indicating this solution's maturity. Proof of the advancement in legged robotics is also visible from the overall teams’ choices during the three years of the SubT Challenge. In fact, during the first year of the competition (\num{2019}), only two teams out of eleven chose legged robots as their main ground vehicles, and both did not perform very well. Only two years later, during the Final Event, the top-6 teams utilized quadruped systems (``ANYmal C'' for our case and “Boston Dynamics Spot'' for the other teams). 
We believe that this is a tipping point in the field of UGVs, with legged robots becoming a superior and broadly applicable choice of platform. 
However, despite these advances, there is still plenty of room to increase the capabilities of the systems further. For example, the ability to climb over rough terrain is still in its infancy compared to the natural counterparts and can certainly be improved. Second, the separation of navigation, motion planning/following, and locomotion control drastically limits the possible maneuvers. Third, the robots still rely on abstract geometric terrain representations for locomotion while lacking a semantic understanding for locomotion. And lastly, they have limited to no active interaction capabilities to manipulate the environment, e.g., for terrain probing or to remove obstacles.

\textbf{Marsupial aerial-ground system-of-systems are important:} Our team developed a host of flying robots that conducted successful exploration missions during the first two circuit events, including the vertical traversal of staircases. However, as the SubT Challenge entailed that all robots start from the same starting point it essentially implied that flying robots would be much more useful if deployed deeper inside the underground environment. Indeed, the finals course contained a massive ``cave''-like chamber where if we had the marsupial integration of one of our small-sized flying robots onboard ANYmal we could have gained the benefits from the versatile navigation capabilities of aerial systems without their limited endurance being an important constraint. Unfortunately, due to limited development time, our team could complete the marsupial integration of ANYmal C with RMF-Owl~\cite{https://doi.org/10.48550/arxiv.2205.05477} only after the competition was over. 

\textbf{Communications:}
During the first two years of the competition, the reliability and performance of the employed communication solutions corresponded to a critical challenge in our experience. Therefore, before the Final Event, we developed and deployed an improved and more reliable hardware solution for subterranean communication, based on an off-the-shelf baseline radio module, alongside a new iteration of the deployable WiFi communication-extender modules. However, achieving an always-reliable communication link is particularly challenging in underground environments. To that end, in our experience a high level of autonomy in the robot agents allowed to mitigate this problem by enabling exploration when not in communication range.

\textbf{Use of simulation and synthetic data:} The artifact detection module was trained using hand-labeled data collected in underground settings. However, this approach was slow and cumbersome, resulting in the collection of a limited amount of data. Photo-realistic simulators and rendering engines could be exploited to generate data captured using various simulated camera parameters, light conditions, obscurants, and degradation, making the artifact detection module more robust.
Apart from this, simulation can help in stress testing the methods, debugging minor errors, and tuning parameters of the algorithms which is not feasible during a field test. On this front, we utilized the RotorS Gazebo Simulator~\cite{furrer2016rotors} for the simulation of aerial robots and a Gazebo-based simulator for the ANYmal C legged robots. Specifically, the path planning algorithms were tested in various simulation environments, both artificial and real models obtained or created from field data. However, the use of simulation was limited to testing individual software components for a single robot. High-fidelity simulators could be exploited to simulate more complex elements such as photo-realistic environments, obscurants, variable lighting conditions, dynamic obstacles, and behavior of communication networks for large-scale realistic simulations. Additionally, multi-robot simulation can be exploited for complete mock missions.

\subsection{On the importance of the frequency and the character of field testing}
We realized that many characteristics of the subterranean settings as well as the challenges presented by them cannot be fully understood without going into these environments. The experience from field deployments is crucial in making each component in the system robust. 

\textbf{Making the robots ``field-ready''}\\
Team CERBERUS conducted a large number of field deployments in diverse environments, across different countries, replicating the settings of the competition as much as possible including timed missions (detailed description of the nature of field deployments is provided in Section \ref{sec:field_deployment_preparation}). We conducted at least one field test every month in the year before the Final Event. The selected deployment sites spanned across all three categories of environments in the competition - tunnel, urban, and cave, each presenting different challenges as described in section~\ref{sec:qualitative_deployments}.
We have demonstrated that through testing in diverse environments, the systems were able to tackle the majority of the challenges presented across all three types of environments. Furthermore, multiple tests in the same environment helped in benchmarking the systems and keeping track of the progress.

These field tests not only helped in making the systems field ready in terms of the technology, but also served as practice missions for the Human Supervisor and the Pit Crew. Treating the deployments in the field tests as if they were in the actual competition helped in fine tuning the deployment procedures and related technologies such as the user interface, software stack launching scripts, and safety checks before deployment. 
Thoroughly planning, preparing, and testing the hardware and software components in the days before the field deployment and during the weekly shakeouts was a crucial practice that significantly improved the outcome of each field test, both in terms of the number of tested features and overall system integration in a mock-up mission scenario. The day before each field test, a functionality test of all the robots and features that were to be deployed was conducted. If issues were found in any feature, it was not tested in the field experiment. This helped in identifying issues in small yet critical parts of the systems with can easily get ignored while developing the core technology. Furthermore, this approach ensured that each implemented feature was thoroughly tested during the development process before deploying on the robot. Periodic field deployments forced us to have internal, hard deadlines that pushed the development of ``field-ready'' robotic systems.

As our team was distributed across the globe, several components of the system, including entire robots, were developed at different places. However, for the whole system-of-systems to work together good integration among all the hardware and software components across the robots is critical. Hence, we conducted several integration exercises involving members of all groups participating in the team. These exercises included integration of software and hardware components across all robots, functionality testing, and field deployment of heterogeneous robot teams. In our experience, the integration exercises were important for a) ensuring that all robots had the necessary hardware and software components correctly configured for working with each other and the Base Station, b) making the Human Supervisor familiar with the setup and user interface for all types robots, and c) conducting mock, timed missions using the entire fleet.

\textbf{Importance of field testing for resilient perception and autonomy} \\
The response of real sensors and systems in the real environments is hard to replicate synthetically, especially for different sensing modalities. Therefore such experience and data is crucial for the development and testing of the perception and autonomy algorithms. Many of the features and development directions of the CERBERUS autonomy solutions have evolved from our experiences in the field tests. Some of these features of GBPlanner2 include, but are not limited to, the bifurcated local/global planning architecture for handling large scale topology with complex geometry and terrain, optimistic ground existence check to tackle gaps in the volumetric map, introduction of the hanging vertices for handling negative obstacles. Similarly, we implemented the safety checks and recovery behaviors from unexpected events in the ANYmal behavior tree based on our experience in the field. Moreover, we introduced batched motion cost prediction for the ANYmal navigation planner to improve real-world path safety and prevent very rare planning time outliers, which only happened when executing on the robot. Field tests also motivated the introduction of a foothold safety threshold to avoid planning close to negative obstacles, which are often not adequately mapped during actual deployment. Additionally, field experiments provided a better understanding of the robot's response to the navigation commands and what can be considered as a safe action.

Not all sensing modalities can be captured in simulation accurately (e.g., thermal camera frames), and their behavior in subterranean environments can be understood only from field data. Furthermore, several hardware challenges affect the performance of perception solutions. These include handling sensor delays, time synchronization between sensors, and issues caused by other components of the robot (e.g., vibrations, occlusions, etc.). Through field evaluation, several cases of sensor degradation were identified (e.g., self-similarity, smoke, fog, darkness) which guided the development of the multi-modal localization and mapping framework.
Hence, the data collected in the field is the only way to evaluate, benchmark, and tune the localization and mapping algorithms to tackle all the challenges present in the subterranean environments.
Furthermore, the field experiments also showed the computing limitations of each robot so that the algorithms can be modified or tuned accordingly.

%% file: 06_conclusions.tex
This paper presents a comprehensive report of the technological progress made by Team CERBERUS participating in the DARPA SubT Challenge and specifically towards the Final Event. We detailed the CERBERUS' vision and motivation for the Team's overall ``system-of-systems'' methodology, followed by an extensive presentation of the employed robotic systems. The contributed research allowed a high level of autonomy and robustness in the CERBERUS robots, operating inside visually-degraded, large-scale, narrow, and broadly complex underground settings. At the core of the CERBERUS robotic system is a team of legged and flying robots, including collision-tolerant designs, further enhanced with a roving platform. The implemented autonomy components, including multi-modal perception and path-planning, navigation and control of legged robots, and the automated artifact detection and scoring systems, are detailed alongside the operator interfaces for exerting high-level control. Finally, we presented the Team's performance in the winning Prize Round of the challenge, followed by the critical lessons learned from these experiences. We hope that the presented work and released open-source code can contribute towards the overall community goal of enabling resilient robotic autonomy in extreme underground environments.

%% file: 07_open_source_list.tex
With the goal to support the overall community around subterranean robotics, our team focused on open-sourcing significant components of our technological solution. Similarly, data from field deployments are released.

\subsection{Open source code}
A selected - but not exhaustive - list of relevant open-source packages of our team are outlined below. 

\emph{GBPlanner2} - a graph-based exploration planner for subterranean environments:~\url{https://github.com/ntnu-arl/gbplanner_ros}

\emph{Art Planner} - a local navigation planner for legged robots:~\url{https://github.com/leggedrobotics/art_planner}

\emph{Elevation Mapping Cupy} - a GPU-based robot-centric elevation mapping framework for rough terrain navigation:~\url{https://github.com/leggedrobotics/elevation_mapping_cupy}

\emph{Maplab} - an open visual-inertial mapping framework, which includes the M3RM centralized multi-modal, multi-robot mapping extension:~\url{https://github.com/ethz-asl/maplab}

\emph{Darknet ROS} - Real-Time Object Detection for ROS:~\url{https://github.com/leggedrobotics/darknet_ros}

\emph{Kalibr} - a visual-inertial calibration toolbox:~\url{https://github.com/ethz-asl/kalibr}

\emph{ROVIO} - a Robust Visual Inertial Odometry framework:~\url{https://github.com/ethz-asl/rovio}

\emph{Voxblox} - a library for flexible voxel-based mapping:~\url{https://github.com/ethz-asl/voxblox}

\subsection{Simulation models}
The SubT Challenge Systems and Virtual Competitions are interconnected with the Systems' teams delivering models of their robots for the Virtual Competition, while progress achieved in the Virtual Competition can benefit the autonomy functionalities in the Systems Competition. Models of most of the robots fielded by our team in the Final Event have already been released.
The following robots related to the Final Event are available. 

\emph{ANYmal C SubT Explorer:}~\url{https://github.com/osrf/subt/tree/master/submitted_models/cerberus_anymal_c_sensor_config_1}

\emph{RMF-Owl aerial scout:}~\url{https://github.com/osrf/subt/tree/master/submitted_models/cerberus_rmf_sensor_config_1}

\emph{DJI Matrice M100-based aerial scout:}~\url{https://github.com/osrf/subt/tree/master/submitted_models/cerberus_m100_sensor_config_1}

\subsection{Open datasets}

We have released an open dataset from the Prize Round of the Final Event of the DARPA SubT Challenge. This dataset is accessible at  \url{https://www.subt-cerberus.org/code--data.html}. Through the same link, possible further releases of datasets or new software will become accessible. 